\PassOptionsToPackage{table}{xcolor}

\documentclass{article}
\usepackage{iclr2027_conference,times}

\usepackage[utf8]{inputenc} %
\usepackage[T1]{fontenc}    %

\usepackage{amsmath,amsfonts,bm}

\def\eqref#1{equation~\ref{#1}}

\def\1{\bm{1}}

\DeclareMathAlphabet{\mathsfit}{\encodingdefault}{\sfdefault}{m}{sl}
\SetMathAlphabet{\mathsfit}{bold}{\encodingdefault}{\sfdefault}{bx}{n}

\usepackage{hyperref}       %
\usepackage{url}            %
\usepackage{booktabs}       %
\usepackage{amsfonts}       %
\usepackage{nicefrac}       %

\usepackage{graphicx}

\usepackage{physics}
\usepackage{caption}
\usepackage{subcaption}
\usepackage{float}
\usepackage[nolist]{acronym}
\usepackage{xfrac}
\usepackage{wrapfig}
\usepackage{tabularx} %
\usepackage{enumitem} %
\usepackage{siunitx} %
\usepackage{xspace} %
\usepackage{tikz}
\usetikzlibrary{calc}
\usepackage{multirow}
\usepackage{dblfloatfix}
\usepackage{placeins}
\usepackage{afterpage} 
\usepackage{pdflscape}
\usepackage{rotating}
\usepackage[capitalize,noabbrev]{cleveref}
\usepackage{algorithm}
\usepackage{algorithmic}

\AtBeginDocument{\RenewCommandCopy\qty\SI}
\DeclareSIUnit{\nothing}{\relax}

\newcolumntype{Y}{>{\raggedleft\arraybackslash}X}

\graphicspath{{./resources/}{./resources/out_imgs/}}

\acrodef{CDF}[CDF]{cumulative distribution function}
\acrodef{RNN}[RNN]{recurrent neural network}
\acrodef{CNN}[CNN]{convolutional neural network}
\acrodef{MEA}[MEA]{multi-electrode array}
\acrodef{IQM}[IQM]{interquartile mean}
\acrodef{MAE}[MAE]{mean absolute error}
\acrodef{NLL}[NLL]{negative log-likelihood}
\acrodef{RGC}[RGC]{retinal ganglion cell}

\definecolor{green-line1}{HTML}{59997E}
\definecolor{green-text1}{HTML}{43735f}

\definecolor{pos-green}{HTML}{2ECC71}
\definecolor{neg-red}{HTML}{E74C3C}
\definecolor{neg-green}{HTML}{2ECC71}
\definecolor{pos-red}{HTML}{E74C3C}
\definecolor{neg-green-text}{HTML}{2c9c5a}

\definecolor{cell-neg-red}{HTML}{ffc3a6}
\definecolor{cell-pos-red}{HTML}{ffc3a6}
\definecolor{cell-pos-green}{HTML}{91cca9}
\definecolor{cell-neg-green}{HTML}{91cca9}

\newcommand{\model}[1]{\mbox{\textit{#1}}}
\newcommand{\code}[1]{\texttt{#1}}

\title{Edge Selection for the Effective use of Piecewise-Constant Distributions as Neural Network Outputs for Event Prediction}

\author{%
  Kevin Doran \\
  Sussex Neuroscience \\
  School of Engineering and Informatics \\
  University of Sussex, UK \\
  \texttt{k.a.doran@sussex.ac.uk} \\
  \And
  Tom Baden \\
  Sussex Neuroscience \\
  School of Life Sciences \\ 
  University of Sussex, UK \\
  \texttt{t.baden@sussex.ac.uk} \\
}
\hypersetup{
    hidelinks
}
\iclrfinalcopy

\begin{document}
  
\maketitle
\lhead{}

\begin{abstract}
We study the output representation of a neural network used for next event prediction. We propose partitioning the time axis into a fixed set of intervals and having a neural network output a categorical distribution over them, which we map to a (mostly) piecewise-constant probability density. We present an optimization procedure that selects interval edges in order to maximize data likelihood under the representation. The representation is well suited to processes whose inter-event distribution is a mixture of smooth and sharply peaked components—a pattern we find common in event data recorded from real-world processes.
\end{abstract}

\section{Introduction} \label{sec:introduction}
Probabilistic modelling of event sequences with neural networks is the concern of neural temporal point processes (TPPs), reviewed by \citet{shchurNeuralTemporalPoint2021}, \citet{bosserPredictiveAccuracyNeural2023} and \citet{linExtensiveSurveyEmpirical2025}. We focus on a narrow but consequential design choice: for models that take in an event history and predict the time until the next event, how should a network \textit{represent} its prediction?

A prediction, in this context, is a probability density over \( (0, \infty) \). Existing models reach such a density by parameterizing an intensity (hazard) function, or by parameterizing a distribution such as a mixture of log-normals. Our proposal takes the latter form. We have a neural network emit a categorical distribution over \( N \) intervals tiling \( (0, \infty) \), and we interpret this output as a piecewise-constant density function (with a single exponential tail for the final unbounded interval). The effectiveness of this approach depends heavily on the choice of interval edges. We present a dynamic programming optimization of the intervals that maximizes data likelihood under an empirical distribution. We will show that outputting a piecewise-continuous model over these edges beats existing output heads at next event prediction on 14 standard datasets in terms of log-likelihood.

\begin{wrapfigure}[10]{R}{0.50\textwidth}
\vspace{-0.50cm}
\includegraphics[width=0.50\textwidth]{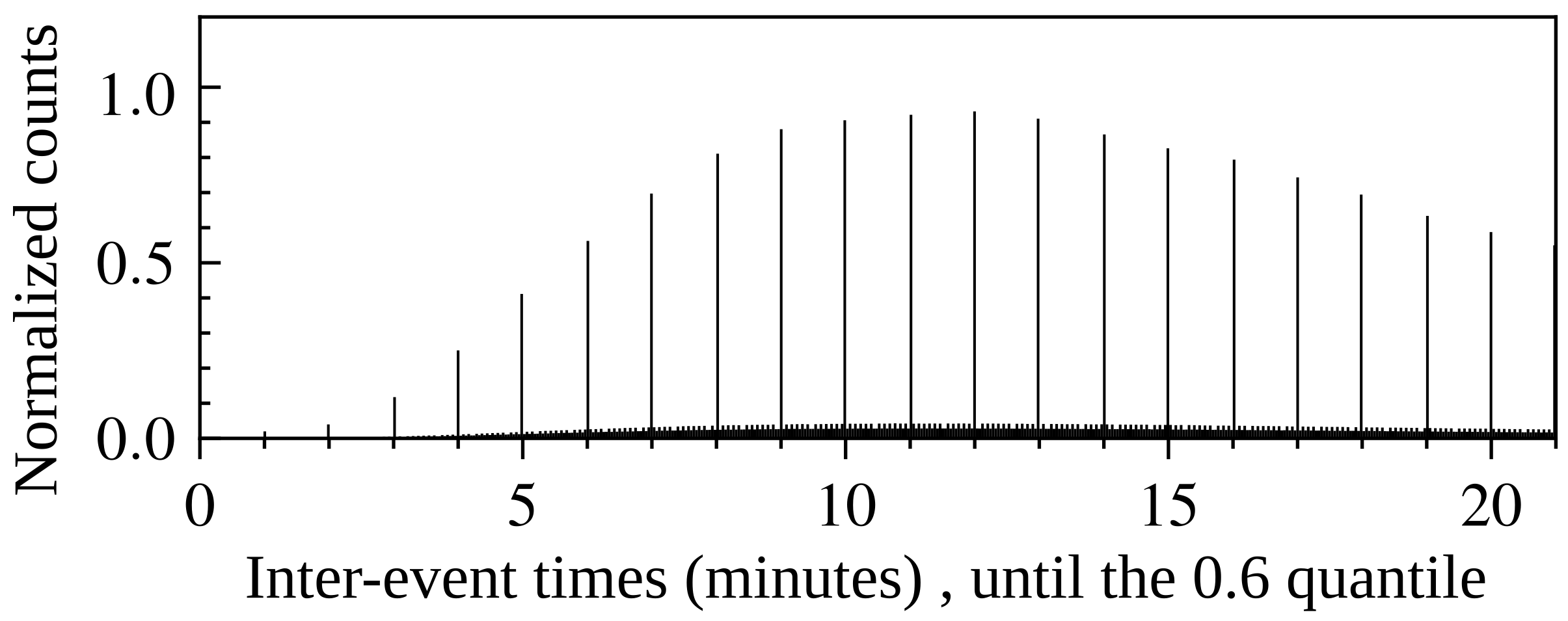}%
\caption{Histogram of inter-event times (< 0.6 quantile) for the NYC taxi dataset \citep{whongNYCsTaxiTrip2014}.}%
\label{fig:taxi_hist}
\end{wrapfigure}

This approach will be shown to be particularly effective for processes that exhibit a mixed distribution containing continuous and discrete parts. The times between taxi pickups in New York City obtained by \citet{whongNYCsTaxiTrip2014} have such a distribution—the distribution is relatively smooth but is punctuated by regular peaks (see \cref{fig:taxi_hist}). \cref{app:dataset_histograms} presents the histograms of 14 real-world datasets, demonstrating that this phenomenon is common.

The next section establishes the next event prediction problem; \cref{sec:pchead} describes the form of the piecewise-constant density head we use; \cref{sec:interval_edges} gives the edge optimization procedure; \crefrange{sec:experiments}{sec:results} cover experiment setup and results; and ablations in \cref{sec:ablations} test properties of the output representation. Beyond supporting the effectiveness of our approach, the results hint that next event prediction on datasets recorded from real-world processes should consider a resolution of interest, possibly the precision at which events are recorded. This and other points are discussed in \cref{sec:discussion} and \cref{sec:limitations}.

\begin{figure}[t]
  \centering
  \includegraphics{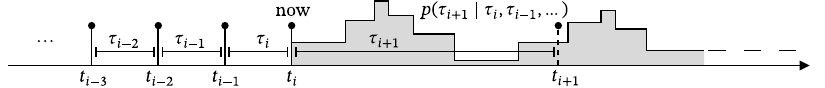}
  \caption{The next event prediction task. Given a history of inter-event times \( \tau_1, \tau_{2}, \dots, \tau_{i} \) up to the present, the task is to emit a probability distribution \( p(\tau_{i+1} \mid \tau_i, \tau_{i-1}, \dots) \) for the time until the next event, \( \tau_{i+1} \). This paper considers a density that is piecewise-constant over bins of non-uniform width, with an exponential tail over the final infinite interval.} 
  \label{fig:nexteventprediction}
\end{figure}

\section{Next event prediction} \label{sec:nextevent}
Let \( (t_i)_{i=0}^{T} \) be a sequence of non-negative strictly increasing reals representing event times. Denote inter-event times as \( \tau_i = t_i - t_{i-1} \). In this work, all models will receive a history of inter-event times, 
\( \mathcal{H}_{i} = (\tau_i, \tau_{i-1}, \ldots, \tau_{1}) \). The task is: given history \( \mathcal{H}_i \), produce a probability density function over \( (0, \infty) \) representing the time until the next event, \( \tau_{i+1} \). \cref{fig:nexteventprediction} illustrates the task and a model that outputs a piecewise-constant density function. Models are scored by the log-likelihood they assign to inter-event times of a held-out test set. For a model with parameters \( \theta \) producing a density function \( p_{\theta} \), this log-likelihood score is given by:
\begin{equation}
  \ell \ell \;=\; \sum_{i=1}^{T} \log p_\theta(\tau_{i} \mid \mathcal{H}_{i-1}),
  \label{eq:loglikelihood}
\end{equation}
summed across sequences, or if reported as a mean, it is a mean over all events.
For all models we consider, the likelihood is also a training signal (a negative log-likelihood loss).

We do not consider models of event counts within intervals, such as those from \citet{ludkeAddThinDiffusion2023} and \citet{liPointPredictionScore2024}; the probability spaces are not directly comparable as their likelihood is defined over a different observation space.

\section{Output representation: piecewise-constant density function}  \label{sec:pchead}
\begin{figure}[t]
\captionsetup[subfigure]{skip=-2pt, belowskip=2.5pt, justification=centering, margin=0cm}
\centering
\begin{subfigure}[t]{1.898in}
  \includegraphics[width=\linewidth]{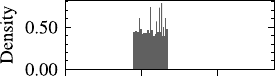}
  \caption{\model{rnn-pc}, Amazon}
\end{subfigure}\hfill
\begin{subfigure}[t]{1.75in}
  \includegraphics[width=\linewidth]{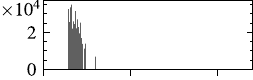}
  \caption{\model{rnn-pc}, Last.fm}
\end{subfigure}\hfill
\begin{subfigure}[t]{1.75in}
  \includegraphics[width=\linewidth]{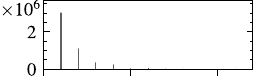}
  \caption{\model{rnn-pc}, PUBG}
\end{subfigure}\\
\captionsetup[subfigure]{skip=-2pt, belowskip=1.5pt, justification=centering, margin=0cm}
\begin{subfigure}[t]{1.90in}
  \captionsetup{justification=raggedleft, singlelinecheck=false, margin=0.22in}
  \includegraphics[width=\linewidth]{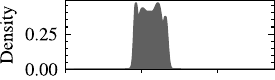}
  \caption{\model{rnn-logmix}, Amazon}
\end{subfigure}\hfill
\begin{subfigure}[t]{1.66in}
  \includegraphics[width=\linewidth]{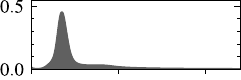}
  \caption{\model{rnn-logmix}, Last.fm}
\end{subfigure} \hfill
\begin{subfigure}[t]{1.63in}
  \includegraphics[width=\linewidth]{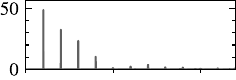}
  \caption{\model{rnn-logmix}, PUBG}
\end{subfigure} \\
\captionsetup[subfigure]{skip=3pt, belowskip=0pt}
\begin{subfigure}[t]{1.90in}
    \includegraphics[width=\linewidth]{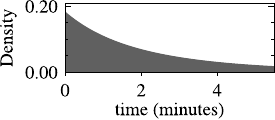}
  \captionsetup{justification=raggedleft, singlelinecheck=false, margin=0.15in}
  \caption{\model{rnn-softplus}, Amazon}
\end{subfigure}\hfill
\begin{subfigure}[t]{1.73in}
  \includegraphics[width=\linewidth]{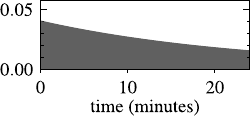}
  \captionsetup{justification=raggedleft, singlelinecheck=false, margin=0.20in}
  \caption{\model{rnn-softplus}, Last.fm}
\end{subfigure} \hfill
\begin{subfigure}[t]{1.66in}
  \includegraphics[width=\linewidth]{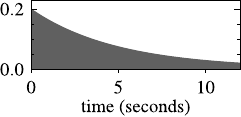}
  \caption{\model{rnn-softplus}, PUBG}
\end{subfigure}%
\caption{Probability distributions corresponding to outputs from 3 trained models with different heads: a piecewise-constant density, a mixture of lognormals density and a softplus hazard; these are labelled as \model{rnn-pc}, \model{rnn-logmix}, and \model{rnn-softplus}, respectively. See \cref{sec:experiments} for model and head details. 
  Each model is given the same input history from three datasets: Amazon, Last.fm and PUBG (datasets are introduced in \cref{sec:datasets} and histograms of inter-event times are in \cref{app:dataset_histograms}).
}
\label{fig:head_densities}
\end{figure}

We build a (mostly) piecewise-constant probability density function over \( (0, \infty) \) in two steps. A preview of outputs emitted by a model using this representation is shown in \cref{fig:head_densities} (top row).

\textbf{Step 1: categorical output.}  A network emits \( N \) logits \( \mathbf{z} = [ z_1, ..., z_N ] \) which a softmax maps to a categorical distribution, \( \mathbf{m} = \operatorname{softmax}(\mathbf{z}) \) over \( N \) intervals (see \cref{fig:pcterms} below). Write \( m_i \) for the probability mass assigned to the \( i \)-th interval. These \( N \) intervals will tile \( (0, \infty) \). Of the \( N \) intervals, \( N - 1 \) have finite length and one (the last) has infinite length. The intervals have strictly increasing edges \( a_1, a_2, \ldots, a_N \). The first interval \( (a_1, a_2) \) begins at zero; the second interval is \( [a_2, a_3) \), and so on until the final interval, \( [a_N, \infty) \). Choosing the edges is the topic of \cref{sec:interval_edges}.

\begin{figure}[h]
  \centering
  \def\svgwidth{\textwidth}
  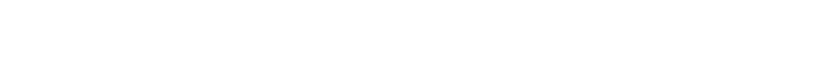
  \captionsetup{skip=5pt}
  \caption{The intervals, logits and probability masses defining the categorical output.}
  \label{fig:pcterms}
\end{figure}

\textbf{Step 2: mass to density.} %
Each finite interval gets a constant density. The final interval, being infinite in length, cannot have a constant non-zero density; instead, it gets an exponential decay.

The density function \( p(\tau) \) is then defined as:
\begin{equation}
    p(\tau) = 
    \begin{cases} 
      \frac{m_i}{w_i} & \tau \in [a_i, a_{i+1}), i < N \\
      m_{N} \lambda e^{-\lambda (\tau - a_N)} & \tau \in [a_N, \infty),
    \end{cases}
    \label{eq:piecewise_constant_density}
\end{equation} 
where \( \lambda \) is the decay of the final interval's density, and \( w_i \) is the interval width, \( w_i = a_{i+1} - a_i \). 

The decay, \( \lambda \), of the exponential distribution is fit by maximum likelihood to the tail end of the training set inter-event times; in this work, we fit \( \lambda \) using the largest \( 0.5\% \) of training set inter-event times.
The final (infinite) interval prevents the density function from being considered wholly piecewise-constant; however, as will be shown, the final interval will be designed to rarely contain events and will not significantly contribute to the likelihood score.

\subsection{Choosing interval edges} \label{sec:interval_edges}
Before placing edges, we must decide how many are needed. We use \( N = 128 \) for the main results and examine a range of values in \cref{sec:ablations}.

For placing the edges, we investigate the following approaches (acronyms are in parentheses): 
\begin{enumerate}[itemsep=0em]
  \item equal-width intervals (\model{wpc})
  \item equal-quantile intervals (\model{qpc})
  \item intervals that maximize likelihood for an unconditional model (\model{pc})
\end{enumerate}

Equal-width and equal-quantile intervals are standard techniques for histogram construction. They were also studied by \citet{kvammeContinuousDiscretetimeSurvival2021} in the context of survival analysis with static covariates. There is a long history of data-dependent construction of variable-width histograms \citep{kanazawaOptimalVariableCell1992, rozenholcCombiningRegularIrregular2010, scottHistogram2010, kontkanenMDLHistogramDensity2007, simensenRandomIrregularHistograms2026}. We use the third approach for the main results, and \cref{app:edge_selection_ablation} covers the other two.

The second and third approaches exploit narrow intervals to capture peaks in the distribution. For the main results, minimum widths are determined by the local spacing of float32 used to store edges. In \cref{sec:ablations}, we probe the effect of minimum width by introducing jitter into the datasets.

\section{Likelihood-maximizing intervals} \label{sec:ours}
The method proposed in this work is to fit a fixed piecewise-constant density function (no input history) by maximum likelihood and inherit the edges of this density as the edges used by the piecewise-constant output head. The fit involves both choosing the edges and the categorical weights, although only the edges are of interest thereafter. Such a fit will naturally place narrow intervals at concentrations of inter-event times. 

For a fixed number of intervals, dynamic programming can obtain maximum-likelihood irregular histograms \citep{rozenholcCombiningRegularIrregular2010}. This is possible as the objective decomposes additively over intervals \citep{jacksonAlgorithmOptimalPartitioning2005}. Related Bayesian construction of histograms by \citet{simensenRandomIrregularHistograms2026} and \citet{scargleSTUDIESASTRONOMICALTIME2013} also exploits this property, and it is the property we use here too.

\subsection{Fitting a piecewise-constant density function to inter-event times}

For a fixed \( N \), we wish to find the set of edges \( a_1, a_2, ..., a_{N} \) and weights \( m_i \) that maximize the data likelihood (of inter-event times) according to \eqref{eq:piecewise_constant_density}. 

Let \( \mathcal{D} = \{ \tau_k \}_{k=1}^{K} \) be the inter-event times available for fitting, split into training and validation sets: \( \mathcal{D}_{\text{train}} = \{\tau_k\}_{k=1}^{K_t} \) and \( \mathcal{D}_{\text{val}} = \{\tau_k\}_{k=1}^{K_v} \).
Denote \( s_i \) as the number of inter-event times from \( \mathcal{D}_{\text{train}} \) that fall into the \( i \)-th interval,
and \( s^*_i \) the same count for \( \mathcal{D}_{\text{val}} \).
To allow the exponential tail to be handled separately, the last edge is not fit, and instead is set to be the next representable value past the final sample, \( a_N = \operatorname{nextafter}(\max(\mathcal{D})) \), making the final interval empty of samples.
The unconditional log-likelihood of \( \mathcal{D}_{\text{val}} \) according to the N-interval piecewise density from \eqref{eq:piecewise_constant_density} is the following sum over the first \( N - 1 \) intervals:
\[ 
  \log p(\mathcal{D}_{\text{val}}) = \sum_{i=1}^{N-1} s^{*}_i (\log m_i - \log w_i), \quad\quad \text{where } w_i = a_{i+1} - a_i.
\]
For fixed edges, the log-likelihood depends on mass through the term, \( s^{*}_i \log m_i \).
For the true data generating distribution, the expected log-likelihood is maximized by setting \( m_i \) to the fraction of inter-event times that would fall into the interval. We estimate this with the empirical frequencies from \( \mathcal{D}_{\text{train}} \), setting \( m_i \) to \( s_i / K_t \). With this substitution, the log-likelihood becomes:
\[ 
  \log p(\mathcal{D}_{\text{val}}) = \sum_{i=1}^{N-1} s^{*}_i (\log s_i - \log K_t - \log w_i)
\]
To maximize this, we can ignore the constant term, \( \log K_t \), and form the following objective function \( F \) for the choice of edges \( \mathbf{a} = [a_1, a_2, \ldots, a_N] \):
\begin{equation}
  F(\mathbf{a})= \sum_{i=1}^{N-1}g(a_i, a_{i+1}), \quad \text{where } g(a_i, a_{i+1}) = s^{*}_i (\log s_i - \log w_i)
  \label{eq:F}
\end{equation}

Each term of the sum can be calculated independently by knowing the edges of an interval and the number of events that fall within it. By maximizing \( F \), we are estimating the piecewise-constant density function with the lowest cross-entropy to the data-generating distribution. 
We will maximize \( F \) over the possible choices of edges, \( a_1, \cdots, a_N \).

\subsection{Optimization}
While the objective to be maximized is simple, the search space—all possible sets of \( N \) edges—is vast. We restrict the edges to a finite set of candidates and, in doing so,  make the problem amenable to dynamic programming.

\textbf{Edge candidates.} In this work, we create a set of candidates from three sources:
\begin{enumerate}
  \item inter-event times from the data (but excluded from \( \mathcal{D} \))
\item edges produced from the equal-quantile method (see \cref{app:edge_selection_ablation})
\item \( e + \varepsilon \) for each candidate edge from 1) and 2) above, where
  \( \varepsilon \) is a small positive constant 
\end{enumerate} 

By including the edges from the equal-quantile method, at least one set of
edges can produce a likelihood similar to the equal-quantile method.
The third source of candidates is included to allow intervals to have right edges just past candidate samples.
For \( \varepsilon \), we use the float32 unit in the last place (ULP) of the candidate edge.

\begin{figure}[t]
  \centering
  \begin{subfigure}{116pt}
    \centering
    \def\svgwidth{84pt}
    \includegraphics{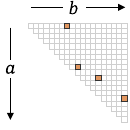}
    \caption{Score table: every candidate interval \( [e_a, e_b) \) has an entry
      \( g_{ab} = s^{*}_{ab}(\log s_{ab} - \log w_{ab}) \), defined for \( b > a \).
      Shaded cells are the chosen hops between edges.}
    \label{fig:edge_dp_table}
  \end{subfigure}
  \hfill
  \begin{subfigure}{270pt}
    \centering
    \includegraphics{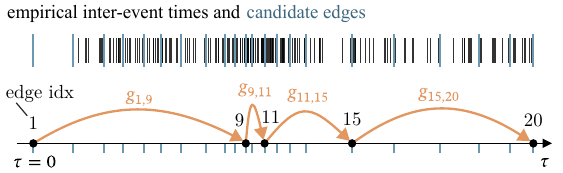}
    \caption{A path between 5 edges (the black nodes) selected from the candidate edges \( \mathcal{E} \) (blue lines). Candidates are dense where the data is dense, allowing narrow intervals to (optionally) be chosen there.}
    \label{fig:edge_dp_path}
  \end{subfigure}
  \captionsetup{skip=6pt}
  \caption{Edge selection as a dynamic program. Choosing edges
    \( \mathbf{a} = [a_1, \ldots, a_N] \) (with \( a_1 = e_1 = 0 \), \( a_N = e_E \)) to
    maximize \( F(\mathbf{a}) = \sum_{i=1}^{N-1} g(a_i, a_{i+1}) \) is equivalent to finding the
    highest-scoring path between the candidate edges \( \mathcal{E} = \{e_1, \ldots, e_E\} \),
    where the score of interval \( [e_a, e_b) \) is the table entry \( g_{ab} \) (a). 
  A small candidate set is shown; in practice \( \mathcal{E} \) is large.}
  \label{fig:edge_dp}
\end{figure}

\textbf{Dynamic programming.} Consider a finite set of candidate edges \( \mathcal{E} = \{e_1, e_2, \ldots, e_E \} \).
The lowest and highest edges are \( e_1 = 0 \) and \( e_E = a_N \). Note that typically \( E \gg N \).
Let \( V(e, r) \) be the best value of \( F \) achievable with \( r \) edges
left to place after edge \( e \). In other words:
\begin{equation}
  V(e, r) = \max_{e' \in \{x \in \mathcal{E} : x > e \}} \left[ g(e, e') + V(e', r - 1) \right].
  \label{eq:V}
\end{equation}
The optimal objective value is \( V(e_1, N - 1) \), with base case \( V(e_E, 0) = 0 \) and \( V(e, 0) = -\infty \) for all \( e < e_E \). A standard application of dynamic programming, described in \cref{alg:edge_opt}, gives the optimal edges.
\begin{algorithm}[h]
  \caption{Edge optimization. Dynamic programming optimization of edge placement to maximize data likelihood for unconditional piecewise-constant density.}
\label{alg:edge_opt}
\begin{algorithmic}[1]
  \REQUIRE sorted candidate edges $e_1 = 0 < \cdots < e_E = a_N$; training set $\mathcal{D}_{\text{train}}$, validation set $\mathcal{D}_{\text{val}}$, number of intervals $N$.
\ENSURE edges $a_1, \dots, a_N$ maximize $F$ over the candidate set
\STATE $s_{ab} \gets$ count samples in interval $[e_a, e_b)$ from $\mathcal{D}_{\text{train}}$ \COMMENT{prefix sum}
\STATE $s^{*}_{ab} \gets$ count samples in interval $[e_a, e_b)$ from $\mathcal{D}_{\text{val}}$ \COMMENT{prefix sum}
\STATE $g_{ab} \gets s^{*}_{ab}\,\big(\log s_{ab} - \log(e_b - e_a)\big)$ for all $a < b$, and $g_{ab} \gets 0$ where $s^{*}_{ab} = 0$ \COMMENT{create table of all interval scores}
\STATE $V[E, 0] \gets 0$ \quad and \quad $V[a, 0] \gets -\infty$ for all $a < E$ \COMMENT{base cases}
\FOR{$r = 1$ to $N - 1$}
    \FORALL{candidate indices $a$}
        \STATE $V[a, r] \gets \max_{b > a} \big(g_{ab} + V[b, r-1]\big)$
        \STATE $\mathrm{choice}[a, r] \gets$ the maximizing $b$ \COMMENT{back-pointer}
    \ENDFOR
\ENDFOR
\STATE recover edges $a_1, \dots, a_N$ by following $\mathrm{choice}$ from $(1, N-1)$
\STATE \textbf{return} edges $a_1, \dots, a_N$
\end{algorithmic} 
\end{algorithm} 

\textbf{Complexity.}
To choose \( N \) edges from \( E \) candidate edges, the algorithm 
uses \( O(E^2) \) time and memory to calculate and store the score matrix, \( g_{ab} \).
The bottom-up dynamic programming procedure filling in the \( V \) array uses
\( O(E^2N) \) time and \( O(EN) \) memory, making the overall complexity \(O(E^2N) \) time and \( O(E^2 + EN) \) for memory. The \( O(E^2N) \) term dominates. In this work, we use
\( E = 15000 \). In our implementation, vectorized array operations evaluate the \( E^2 \) transitions at each stage, substantially reducing the practical runtime (see \cref{app:dp_runtime} for wall-clock runtimes).

\textbf{Dataset splits.} 
In this work, at most 10\% of the dataset is sampled to source candidate edges. The remaining inter-event times are used to form \( \mathcal{D}_{\text{train}} \) and \( \mathcal{D}_{\text{val}} \), which we implement as a 5-fold split. The inter-event times come entirely from the training set designated for the subsequent training of the neural network model.

The next section describes experiments that test this output representation.

\section{Experiment setup} \label{sec:experiments}
We compare our proposed representation to existing output heads. All heads are paired with 3 model stems and trained on 14 real-world datasets. Below we describe the output heads, the model stems, the datasets and the training and evaluation procedure.

\subsection{Baseline output representations} \label{sec:heads}
Among the many diverse models designed for next event prediction, the preponderance of them use one of the following output representations. 
These output representations will be paired with the model stems from \cref{tbl:base_models}.

\textbf{Mixture of log-normals (\model{logmix})}. A network emits \( M \) tuples \( (\mu_i, \sigma_i, \pi_i) \) which parameterize a mixture of \( M \) log-normal distributions. Investigated by \citet{shchurIntensityFreeLearningTemporal2019} and \citet{panosDecomposableTransformerPoint2024}.
Unlike the approaches below, this represents a probability distribution directly, rather than as a hazard function. We use this head with \( M = 64 \) components.

\textbf{Constant hazard (\model{const)}}. A network emits a scalar which parameterizes a constant hazard function \citep{huangRecurrentPoissonProcess2019, liLearningTemporalPoint2018}. A constant hazard corresponds to an exponential probability distribution.

\textbf{Exponential hazard (\model{exp)}}. A network emits a scalar which parameterizes an exponential hazard function \citep{duRecurrentMarkedTemporal2016, upadhyayDeepReinforcementLearning2018}. An exponential hazard corresponds to a Gompertz distribution.

\textbf{Softplus hazard (\model{softplus)}}. A network emits one scalar \( x \) which parameterizes a hazard function of the form \( h(\tau) = \beta \, \operatorname{softplus}(\frac{\alpha \tau + x}{\beta}) \). A softplus hazard does not have a corresponding analytical distributional form. A softplus hazard is used in multiple works \citep{meiNeuralHawkesProcess2017, zuoTransformerHawkesProcess2020, zhangSelfAttentiveHawkesProcess2020}.

\textbf{Neural network parameterized hazard (\model{nn})}. A network takes in time as an input and evaluates the integral of a hazard function at that time. Investigated by \citet{omiFullyNeuralNetwork2019}. This approach uses automatic differentiation to compute both the hazard and the corresponding density function.

\textbf{Piecewise-continuous hazard}. A network outputs an array of values which parameterizes a piecewise-continuous hazard function. This approach was investigated by \citet{jingNeuralSurvivalRecommender2017}, using equal-width intervals. We do not test the output head from \citet{jingNeuralSurvivalRecommender2017}; however, we do test a piecewise-continuous hazard on the same edges calculated in our proposed approach (\cref{sec:ours}) as an ablation in \cref{sec:ablations}. 

Example outputs from models using the \model{logmix} and \model{softplus} heads are shown in \cref{fig:head_densities}. Limitations of existing heads are discussed in \cref{app:model_limitations}.

\subsection{Model stems} \label{sec:model_stems}
Each output head above is paired with 3 model stems listed in \cref{tbl:base_models}. The \ac{RNN} model is used so as to be similar to the recurrent networks used in the original papers introducing many of the heads. The other stems are transformer-based models in two sizes.
\begin{table*}[t]
  \centering
  \caption{Model stems used in experiments. See \cref{app:models} for implementation details. Most heads from \cref{sec:heads} were originally studied with recurrent neural network (RNN) stems.}
    \begin{tabularx}{\textwidth}{l X l l l l l}
        \toprule
        \textbf{Label} & \textbf{Stem description} & \textbf{Input} & \textbf{Layers} & \textbf{Heads} & \textbf{Embed dim.} & \textbf{Parameters} \\
        \midrule
        \model{rnn} & gated recurrent unit \citep{choLearningPhraseRepresentations2014} & 32 & 1 & NA & 64 & 13k \\
        \model{gpt-a} & GPT-2 transformer \citep{radfordLanguageModelsAre2019} & 128 & 2 & 4  & 16 & 108k \\
        \model{gpt-b} & GPT-2 transformer & 128 & 6 & 4  & 32 & 1.20M \\
        \bottomrule
    \label{tbl:base_models}
    \end{tabularx}
\end{table*}

\subsection{Datasets} \label{sec:datasets}
All datasets used in this work are sequences of event times (or lists of such sequences). The datasets used were made available alongside \citet{shchurIntensityFreeLearningTemporal2019}, \citet{ludkeAddThinDiffusion2023} and \citet{xueEasyTPPOpenBenchmarking2024} and are described in further detail in \cref{app:datasets}. 

\subsection{Training and evaluation} \label{sec:training_and_evaluation}
Models are trained to minimize NLL, and performance is evaluated in terms of NLL and \ac{MAE}. Models are trained 10 times for each dataset. The train/val/test split is randomly sampled for each repeat. Training and evaluation are described in more detail in \cref{app:training_settings}.

\section{Results}\label{sec:results}
A subset of results is shown in \cref{tbl:baseline}, comparing \model{pc} and \model{logmix} heads using the RNN stem in terms of negative log-likelihood (NLL). Focus is given to the \model{logmix} head throughout this work, as it is the most effective among the existing approaches tested. Results for all stem and head combinations are reported in \cref{tbl:app_baseline_full_nll} 
and \cref{tbl:app_baseline_full_mae} 
in \cref{app:results}. The \model{rnn-pc} model outperforms existing models on all datasets in terms of NLL, and has the lowest (best) mean rank across datasets in terms of MAE.

\begin{table*}[h]
\caption{\model{rnn-pc} and \model{rnn-logmix} evaluated on 14 real-world datasets in terms of test-set NLL and MAE. Both models share the same architecture up until the output representation. Values are means over 10 training repeats where each repeat uses a different random data split. Cells are coloured when the score difference, \( \Delta \), is statistically significant with respect to the standard errors (SE) of the means: \( |\Delta| > 1.96 \sqrt{ \mathrm{SE}_{\text{rnn-logmix}}^2 + \mathrm{SE}_{\text{rnn-pc}}^2} \); \textcolor{neg-green-text}{green} when \model{rnn-pc} scores better, \textcolor{pos-red}{red} when worse.}%
  \centering
{\setlength{\tabcolsep}{4pt}%
\begin{tabular}{llrrrr}
\toprule
 &  & \multicolumn{2}{c}{NLL $\downarrow$} & \multicolumn{2}{c}{MAE $\downarrow$} \\\cmidrule(lr){3-4} \cmidrule(lr){5-6}
Dataset & Unit & rnn-logmix & rnn-pc & rnn-logmix & rnn-pc \\
\midrule
Yelp airport & minutes & 4.71 &  4.69 & 34.68 &  35.04 \\
Yelp Mississauga & minutes & 3.92 & {\cellcolor{neg-green!20} 3.26} & 21.36 &  21.30 \\
Twitter & minutes & 4.04 & {\cellcolor{neg-green!20} 3.02} & 37.23 &  37.34 \\
Wikipedia & minutes & 4.36 & {\cellcolor{neg-green!20} 1.96} & 198.2 &  198.4 \\
Yelp Toronto & hours & 4.81 & {\cellcolor{neg-green!20} 4.60} & 62.82 &  62.75 \\
MOOC & minutes & 1.02 & {\cellcolor{neg-green!20} -6.23} & 290.5 &  290.4 \\
Amazon & days & 1.44 & {\cellcolor{neg-green!20} 1.38} & 13.58 & {\cellcolor{neg-green!20} 13.41} \\
Taobao & minutes & 3.44 & {\cellcolor{neg-green!20} -2.01} & 21.30 &  21.31 \\
PUBG & seconds & 1.64 & {\cellcolor{neg-green!20} -5.87} & 18.32 & {\cellcolor{neg-green!20} 17.51} \\
Reddit Politics & seconds & 4.68 & {\cellcolor{neg-green!20} 0.29} & 53.89 &  53.72 \\
Reddit AskScience & seconds & 5.59 & {\cellcolor{neg-green!20} 2.15} & 189.3 &  189.0 \\
Last.fm & minutes & 4.47 & {\cellcolor{neg-green!20} 2.86} & 667.0 &  674.4 \\
NYC taxi & minutes & 4.15 & {\cellcolor{neg-green!20} -1.30} & 27.48 &  27.50 \\
Stack Overflow & days & 5.09 & {\cellcolor{neg-green!20} 4.94} & 128.7 &  128.7 \\
\bottomrule
\end{tabular}}
\label{tbl:baseline}
\end{table*}

The next section describes a number of follow-up experiments.

\section{Ablations} \label{sec:ablations}
We carry out five extensions to the experiment of the previous section.

\textbf{Edge selection}. We test the equal-width and equal-quantile methods for choosing edges from \cref{sec:interval_edges}. Results are reported in \cref{tbl:app_ablation_edge_selection} in \cref{app:edge_selection_ablation}. In terms of mean rank across datasets, likelihood-maximizing edges beats equal-quantile which in turn beats equal-width, for both NLL and MAE. 
The equal-quantile model has a similar mean rank to the \model{rnn-logmix} model.

\textbf{Edge count}. We test a range of bin counts for the \model{pc} head, from 32 to 2048 and report results in \cref{tbl:app_ablation_bins_nll_both} (NLL) and \cref{tbl:app_ablation_bins_mae_both} (MAE) in \cref{app:edge_count_ablation}. For most datasets (but not all), NLL scores improve and MAE scores worsen as the number of bins increases. The degree to which NLL improves with increasing bin count correlates with the Gini coefficient 
of the datasets (see \cref{fig:app_gini_vs_bins}).

\begin{figure}
  \centering
  \begin{subfigure}[t]{2.8in}
    \vspace{0pt}
    \includegraphics[width=\linewidth]{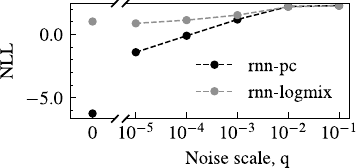}
  \end{subfigure}
  \hfill
  \begin{minipage}[t]{2.6in}
    \vspace{0pt}
  \begin{subfigure}{\linewidth}
    \includegraphics[width=\linewidth, trim=0 0.32in 0 0, clip]{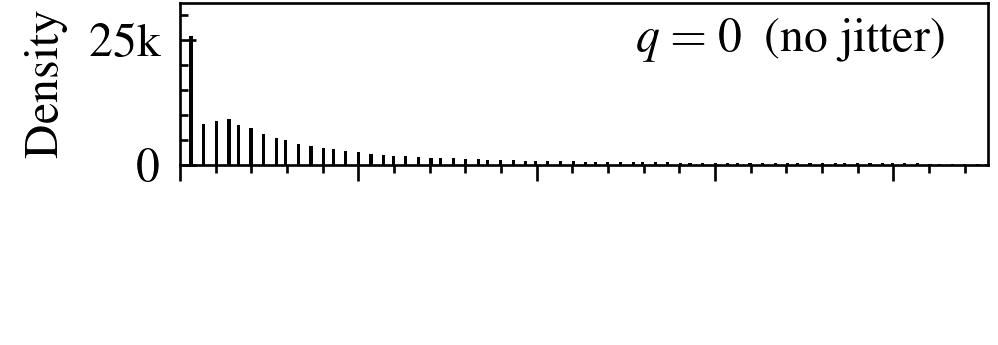}%
  \end{subfigure}
  \begin{subfigure}{\linewidth}
    \includegraphics[width=\linewidth]{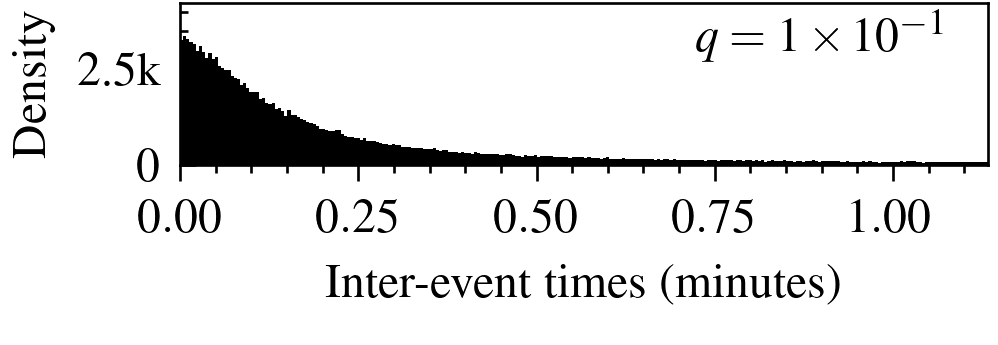}%
  \end{subfigure}
  \end{minipage}%
  \caption{\textbf{Left:} NLL scores for \model{rnn-pc} and \model{rnn-logmix} on the MOOC dataset with jittered event times. Event times are jittered by adding Gaussian noise with standard deviation proportional to the median inter-event time, \( \tilde{\tau} \). That is, \( t_{i}^* = t_{i} + \eta \) where \( \eta \sim \mathcal{N}(0, (\tilde{\tau}q)^2)  \). Five values of \( q \) are used, from \( 1\times 10^{-5} \) to \( 1 \times 10^{-1} \). \textbf{Right:} histograms of the MOOC dataset's inter-event times for the two extreme values of the noise scale, \( q = 0 \) and \( q = 1 \times 10^{-1} \).}%
  \label{fig:ds_jitter_mooc}
\end{figure}

\textbf{Temporal precision}.
We perturb the event times with jitter in order to remove the reward for concentrating density in narrow intervals. Both the \model{pc} and \model{logmix} heads can concentrate density in narrow intervals (see \cref{fig:head_densities} for an example), and it is hypothesized that this is a factor contributing to their performance advantages over the other heads (\cref{app:model_limitations} continues this argument). \cref{fig:ds_jitter_mooc} shows the effect of jittering event times on the MOOC dataset. The same experiment is carried out for the other datasets, and the results appear in \cref{fig:app_ds_jitter} in \cref{app:precision_ablation}. %

\label{sec:density_vs_hazard}
\textbf{Density vs hazard}. We test outputting a piecewise-constant hazard function with the same edges used by the \model{pc} head. A piecewise-constant hazard maps to a piecewise-exponential density. Results appear in \cref{app:density_vs_hazard_ablation}. The piecewise-constant hazard achieves broadly similar results. This suggests that the within-interval shape of the density function or whether it is formulated directly as a density or via a hazard function has less impact on performance than the selection of edge, and that our edge selection procedure can be effective for piecewise-continuous distributions with non-constant within-interval densities.

\label{sec:continuous_ds}
\textbf{Synthetic dataset}. We test the models on a sequence generated by simulating a Hawkes process. The simulation avoids the concentrations of density seen in many of the real-world datasets. Results appear in \cref{app:randproc_ablation}. Models with a \model{logmix} head are the most performant in terms of NLL and MAE; the \model{pc} head comes second, and the \model{nn} head third.

\section{Discussion} \label{sec:discussion}
The results demonstrate that the piecewise-constant output head can be effective for next event prediction. To understand why this output structure outperforms existing heads, it is illustrative to inspect the datasets being used: many of the datasets have inter-event time distributions with shapes that are difficult for a model to approximate when using the existing output structures.

\subsection{Discrete nature of events recorded from real-world processes}
Recording inter-event times from a real-world process does not imply that the distribution of inter-event times will be continuous. Consider the sequence of taxi pickup times from the  NYC taxi dataset \citep{whongNYCsTaxiTrip2014} with inter-event time distribution shown in \cref{fig:taxi_hist}. Inter-event times concentrate on minute boundaries, possibly due to the temporal resolution of the recording device or storage format. The discrete component in the distribution is not unique to the NYC taxi dataset. Inter-event time histograms for all the datasets used in this work are shown in \crefrange{fig:app_ds_histA}{fig:app_ds_hist2d_so}. Some of the histograms look smooth and some look clearly discrete. Some of the histograms that look smooth at one resolution have obvious discrete components when viewed at a finer resolution. The Stack Overflow badges dataset, which has an inter-event time distribution which appears smooth at multiple resolutions (see \cref{fig:app_so_hist}), has a strong concentration of density associated with yearly awarded badges which can be seen in the histograms of next event times in \cref{fig:app_ds_hist2d_so}.

\subsection{Representational limitations of existing output heads}
The nature of the datasets demand that models should be able to output distributions with multiple modes and with narrow peaks. All existing heads tested in this work have properties which limit them with respect to these two goals. \cref{app:model_limitations} covers the limitation of each head in detail. We summarize here: the constant, exponential and softplus hazard outputs correspond to unimodal distributions, and are also restricted in their ability to concentrate density at any non-zero time point. The mixture of lognormals head is far more flexible, but it encounters training instability in the presence of narrow peaks in the target distribution. Finally, the neural-network based head, while also capable of outputting complex distributions, is limited in its ability to modulate its output for different inputs. 

In comparison, the piecewise-constant output is flexible enough to express distributions with discrete components and multiple modes, and it can be trained reliably to do so. We hypothesize that the proposed edge selection method allows for an effective allocation of predictive resolution. We further hypothesize that the within-interval shape (constant, exponential, etc.) is of secondary importance, especially as the interval count is increased. Some evidence to support this comes from the hazard function experiment from \cref{sec:density_vs_hazard}, where the piecewise-constant hazard output, which maps to a piecewise-exponential density function, performed similarly to the piecewise-constant density output structure. 

\subsection{Implications of the disappointing results of larger models}
While increased flexibility of the output structure brought performance improvements, increased model capacity through larger parameter counts did not result in obvious improvements. All output heads were attached to three model stems: an RNN stem and two transformer models (see \cref{tbl:base_models}). The parameter counts spanned three orders of magnitude: 13k, 108k and 1.20M parameters. When scoring the stems in terms of mean rank NLL and MAE across datasets (\cref{tbl:app_baseline_stem_rank}), the largest model stem, \model{gpt-b}, is consistently the worst performing stem to be paired with any head. We draw on this result to hypothesize that larger models may perform less effectively on account of properties of the datasets. We hypothesize (and do not investigate) that qualities such as dataset size may mean that the regularization effects of model size give smaller models an advantage. If this were true, it would suggest that a benchmark formed of these commonly used datasets is insufficient to infer comparative model performance in contexts that demand larger models.

\section{Limitations of this work} \label{sec:limitations}
We argue that this work, and the field in general, is limited by the use of probability \textit{density} for evaluating next event predictions. 
The distributions of inter-event times collected from real-world processes often contain sharp peaks, and it is suitable (and desirable) that models can represent these; however, evaluating in terms of density allows narrow intervals to concentrate density beyond what would be considered useful in a practical application. For example, if a taxi company wished to predict future pickup times, there would be a temporal resolution of interest, say minutes or seconds, beyond which there is little value in making predictions. 
These considerations make a probability mass function, rather than a density function, seem like an appropriate representation for making predictions. If there is no application to inform a minimum resolution of interest, then the resolution of the recording device used to capture the events could be used. We do not pursue this direction in this work. There would be further value if real-world \textit{tasks} rather than simply event sequences from real-world processes were considered, as many of the current datasets used in this and prior work do not have an obvious connection to an application, and thus, they lack application-driven evaluation metrics.

\section{Conclusion}
For the task of next event prediction, this work proposes to have a model output a mostly piecewise-constant density function (or \cref{app:density_vs_hazard_ablation}'s piecewise-constant hazard function). We present a dynamic programming optimization procedure for selecting the edges of this piecewise-constant density function. The proposed output representation achieves competitive performance. Where the approach markedly outperforms existing output representations, we hypothesize that our approach better represents the discrete components present in datasets' inter-event distributions. The results raise the question of whether probability density is always suitable for evaluating next event prediction models operating on event sequences recorded from real-world processes.

\bibliography{main}
\bibliographystyle{iclr2027_conference}

\clearpage

\appendix

\section{Supplementary results} \label{app:results}
This appendix presents results supplementary to the main results in \cref{sec:results}. The supplementary results are:
\begin{enumerate}
  \item \cref{app:app_results_ci}: confidence intervals for the \model{rnn-logmix} vs \model{rnn-pc} table (\cref{tbl:baseline}).
  \item \cref{app:app_results_models}: an extension of \cref{tbl:baseline} that includes score differences of all stem-head combinations, with \model{rnn-logmix} as a reference. These results also include two models implemented by following the papers by \citet{zuoTransformerHawkesProcess2020} and \citet{panosDecomposableTransformerPoint2024}, \model{thp} and \model{dtpp} respectively.
\end{enumerate}
For ablations, see \cref{app:app_ablations}.

\subsection{Confidence intervals for main results (\cref{tbl:baseline})} \label{app:app_results_ci}
In \cref{sec:results}, \cref{tbl:baseline} showed results for \model{rnn-logmix} and \model{rnn-pc} evaluated on 14 real-world datasets, averaged over 10 runs. These results are extended by \cref{tbl:app_baseline_nll_ci} (NLL) and \cref{tbl:app_baseline_mae_ci} (MAE) below to include information on the variability over the 10 runs. Training runs differ by the seed used to randomly split the datasets, by random model initialization and by the stochasticity of the training process.

\begin{table*}[h]
  \caption{Confidence intervals for \cref{tbl:baseline}'s NLL results (from \cref{sec:results}). The \model{rnn-logmix} and \model{rnn-pc} models evaluated in terms of test set \ac{NLL} (lower is better) on 14 real-world datasets. 10 training repeats were carried out. Datasets were randomly split on each repeat. Point estimates are the mean over the trials. Variability is expressed as \( 95\% \) confidence intervals calculated assuming a normal distribution of the mean: \( \pm 1.96 \frac{s}{\sqrt{10}} \), where \( s \) is the sample standard deviation.} 
\label{tbl:app_baseline_nll_ci}
\centering
\vspace{0.5em}
\begin{tabular}{lrr}
\toprule
Dataset & rnn-logmix & rnn-pc \\
\midrule
Yelp airport & 4.715 $\pm$ 2.16e-02 & \textbf{4.685 $\pm$ 3.74e-02} \\
Yelp Mississauga & 3.916 $\pm$ 1.93e-02 & \textbf{3.260 $\pm$ 3.89e-02} \\
Twitter & 4.043 $\pm$ 4.44e-02 & \textbf{3.017 $\pm$ 7.35e-02} \\
Wikipedia & 4.364 $\pm$ 6.35e-02 & \textbf{1.962 $\pm$ 1.18e-01} \\
Yelp Toronto & 4.810 $\pm$ 6.27e-02 & \textbf{4.598 $\pm$ 6.98e-02} \\
MOOC & 1.023 $\pm$ 1.38e-01 & \textbf{-6.227 $\pm$ 2.90e-02} \\
Amazon & 1.439 $\pm$ 8.81e-03 & \textbf{1.380 $\pm$ 5.49e-03} \\
Taobao & 3.440 $\pm$ 1.52e-02 & \textbf{-2.010 $\pm$ 6.83e-02} \\
PUBG & 1.643 $\pm$ 1.53e-01 & \textbf{-5.872 $\pm$ 2.68e-02} \\
Reddit Politics & 4.682 $\pm$ 4.48e-02 & \textbf{0.288 $\pm$ 8.24e-02} \\
Reddit AskScience & 5.585 $\pm$ 1.16e-01 & \textbf{2.150 $\pm$ 9.73e-02} \\
Last.fm & 4.465 $\pm$ 9.78e-02 & \textbf{2.856 $\pm$ 2.26e-01} \\
NYC taxi & 4.146 $\pm$ 6.21e-03 & \textbf{-1.295 $\pm$ 4.27e-01} \\
Stack Overflow & 5.091 $\pm$ 2.79e-02 & \textbf{4.941 $\pm$ 2.55e-02} \\
\bottomrule
\end{tabular}
\end{table*}

\begin{table*}[h]
  \caption{Confidence intervals for \cref{tbl:baseline}'s MAE results 
  (from \cref{sec:results}). The \model{rnn-logmix} and \model{rnn-pc} models evaluated in terms of test set \ac{MAE} (lower is better) on 14 real-world datasets. 10 training repeats were carried out. Datasets were randomly split on each repeat. Point estimates are the mean over the trials. Variability is expressed as \( 95\% \) confidence intervals calculated assuming a normal distribution of the mean: \( \pm 1.96 \frac{s}{\sqrt{10}} \), where \( s \) is the sample standard deviation.} 
\label{tbl:app_baseline_mae_ci}
\centering
\vspace{0.5em}
\begin{tabular}{lrr}
\toprule
Dataset & rnn-logmix & rnn-pc \\
\midrule
Yelp airport & \textbf{34.68 $\pm$ 7.42e-01} & 35.04 $\pm$ 7.75e-01 \\
Yelp Mississauga & 21.36 $\pm$ 4.65e-01 & \textbf{21.30 $\pm$ 5.27e-01} \\
Twitter & \textbf{37.23 $\pm$ 6.05e-01} & 37.34 $\pm$ 6.28e-01 \\
Wikipedia & \textbf{198.2 $\pm$ 7.39e+00} & 198.4 $\pm$ 7.44e+00 \\
Yelp Toronto & 62.82 $\pm$ 2.18e+00 & \textbf{62.75 $\pm$ 2.19e+00} \\
MOOC & 290.5 $\pm$ 4.07e+00 & \textbf{290.4 $\pm$ 4.08e+00} \\
Amazon & 13.58 $\pm$ 1.13e-01 & \textbf{13.41 $\pm$ 9.73e-02} \\
Taobao & \textbf{21.30 $\pm$ 3.18e-01} & 21.31 $\pm$ 3.17e-01 \\
PUBG & 18.32 $\pm$ 1.66e-01 & \textbf{17.51 $\pm$ 6.83e-02} \\
Reddit Politics & 53.89 $\pm$ 8.39e-01 & \textbf{53.72 $\pm$ 8.30e-01} \\
Reddit AskScience & 189.3 $\pm$ 4.83e+00 & \textbf{189.0 $\pm$ 5.09e+00} \\
Last.fm & \textbf{667.0 $\pm$ 3.73e+01} & 674.4 $\pm$ 3.79e+01 \\
NYC taxi & \textbf{27.48 $\pm$ 9.12e-01} & 27.50 $\pm$ 9.10e-01 \\
Stack Overflow & \textbf{128.7 $\pm$ 1.25e+00} & 128.7 $\pm$ 1.25e+00 \\
\bottomrule
\end{tabular}
\end{table*}

\subsection{All output heads and model stems} \label{app:app_results_models}
In \cref{tbl:app_baseline_full_nll}, the same 14 datasets are used to test all model combinations of the three stems (\model{rnn}, \model{gpt-a} and \model{gpt-b}) with the six heads: \model{const}, \model{exp}, \model{softplus}, \model{nn}, \model{logmix} and \model{pc}. Also included are two full models, \model{thp}, the model from \textit{Transformer Hawkes Process} \citep{zuoTransformerHawkesProcess2020}, and \model{dtpp}, the model from \textit{Decomposable Transformer Point Processes} \citep{panosDecomposableTransformerPoint2024}. \cref{tbl:app_baseline_full_mae} shows the corresponding MAE scores.

These large tables are condensed by reporting mean ranks in \cref{tbl:app_baseline_model_rank} and \cref{tbl:app_baseline_stem_rank}.

\begin{sidewaystable*}
  \caption{Extension of \cref{tbl:baseline} in \cref{sec:results}. Test set NLL scores (lower is better) for various models on 14 real-world datasets. 6 output heads (\model{logmix}, \model{pc}, \model{nn}, \model{softplus}, \model{exp} and \model{const}) are paired with 3 model stems (\model{rnn}, \model{gpt-a} and \model{gpt-b}). Additionally, \model{thp} from \citet{zuoTransformerHawkesProcess2020} and \model{dtpp} from \citet{panosDecomposableTransformerPoint2024} are included (see \cref{app:othermodels}). Point estimates are the means over 10 repeats, where each repeat uses a random data split. The \model{rnn-logmix} model is used as a baseline, and cells are coloured when the score difference to the \model{rnn-logmix} model, \( \Delta \), is statistically significant with respect to the standard errors (SE) of the means, \( |\Delta| > 1.96 \sqrt{ \mathrm{SE}_{logmix}^2 + \mathrm{SE}_{model}^2} \); \textcolor{neg-green-text}{green} when better than \model{rnn-logmix}, \textcolor{pos-red}{red} when worse. \textbf{Bold} entries are the lowest (best) score point estimates per dataset.} 

  \centering
  \vspace{0.5em}
  {\setlength{\tabcolsep}{3pt}%
\begin{tabular}{lrrrrrrrrrrrrrr}
\toprule
Model & Yelp a. & Yelp M. & Twitter & Taobao & Wiki. & Yelp Tor. & PUBG & MOOC & Reddit Ask. & Amazon & Reddit Pol. & Last.fm & NYC taxi & Stack Overflow \\
\midrule
rnn-logmix & 4.71 & 3.92 & 4.04 & 3.44 & 4.36 & 4.81 & 1.64 & 1.02 & 5.59 & 1.439 & 4.68 & 4.47 & 4.146 & 5.09 \\
gpt-a-logmix &  4.71 &  3.91 & {\cellcolor{pos-red!20} 4.10} &  3.46 &  4.42 & {\cellcolor{neg-green!20} 4.67} & {\cellcolor{pos-red!20} 1.98} & {\cellcolor{neg-green!20} 0.39} & {\cellcolor{neg-green!20} 4.66} & {\cellcolor{pos-red!20} 1.488} & {\cellcolor{neg-green!20} 3.79} &  4.40 & {\cellcolor{neg-green!20} 3.115} &  5.06 \\
gpt-b-logmix &  4.71 &  3.91 & {\cellcolor{pos-red!20} 4.11} & {\cellcolor{pos-red!20} 3.46} &  4.42 & {\cellcolor{neg-green!20} 4.56} & {\cellcolor{pos-red!20} 1.95} & {\cellcolor{neg-green!20} 0.85} & {\cellcolor{neg-green!20} 5.00} & {\cellcolor{pos-red!20} 1.551} & {\cellcolor{neg-green!20} 3.95} &  4.43 & {\cellcolor{neg-green!20} 3.544} &  5.11 \\
rnn-pc &  4.69 & {\cellcolor{neg-green!20} \textbf{3.26}} & {\cellcolor{neg-green!20} \textbf{3.02}} & {\cellcolor{neg-green!20} \textbf{-2.01}} & {\cellcolor{neg-green!20} \textbf{1.96}} & {\cellcolor{neg-green!20} 4.60} & {\cellcolor{neg-green!20} -5.87} & {\cellcolor{neg-green!20} \textbf{-6.23}} & {\cellcolor{neg-green!20} 2.15} & {\cellcolor{neg-green!20} \textbf{1.380}} & {\cellcolor{neg-green!20} 0.29} & {\cellcolor{neg-green!20} 2.86} & {\cellcolor{neg-green!20} -1.295} & {\cellcolor{neg-green!20} 4.94} \\
gpt-a-pc & {\cellcolor{neg-green!20} \textbf{4.67}} & {\cellcolor{neg-green!20} 3.26} & {\cellcolor{neg-green!20} 3.07} & {\cellcolor{neg-green!20} -1.99} & {\cellcolor{neg-green!20} 2.02} & {\cellcolor{neg-green!20} 4.54} & {\cellcolor{neg-green!20} \textbf{-5.89}} & {\cellcolor{neg-green!20} -6.20} & {\cellcolor{neg-green!20} \textbf{2.13}} & {\cellcolor{neg-green!20} 1.385} & {\cellcolor{neg-green!20} \textbf{0.25}} & {\cellcolor{neg-green!20} \textbf{2.81}} & {\cellcolor{neg-green!20} \textbf{-1.915}} & {\cellcolor{neg-green!20} \textbf{4.94}} \\
gpt-b-pc & {\cellcolor{neg-green!20} 4.68} & {\cellcolor{neg-green!20} 3.27} & {\cellcolor{neg-green!20} 3.08} & {\cellcolor{neg-green!20} -1.98} & {\cellcolor{neg-green!20} 2.03} & {\cellcolor{neg-green!20} \textbf{4.44}} & {\cellcolor{neg-green!20} -5.89} & {\cellcolor{neg-green!20} -6.17} & {\cellcolor{neg-green!20} 2.14} & {\cellcolor{neg-green!20} 1.387} & {\cellcolor{neg-green!20} 0.25} & {\cellcolor{neg-green!20} 2.84} & {\cellcolor{neg-green!20} -1.913} & {\cellcolor{neg-green!20} 4.94} \\
rnn-nn & {\cellcolor{pos-red!20} 4.75} & {\cellcolor{pos-red!20} 3.96} &  4.06 &  3.44 &  4.40 & {\cellcolor{pos-red!20} 4.92} & {\cellcolor{pos-red!20} 3.54} & {\cellcolor{pos-red!20} 2.84} & {\cellcolor{pos-red!20} 5.80} & {\cellcolor{pos-red!20} 1.561} & {\cellcolor{pos-red!20} 5.16} & {\cellcolor{pos-red!20} 4.64} &  4.149 & {\cellcolor{pos-red!20} 5.26} \\
gpt-a-nn & {\cellcolor{pos-red!20} 4.76} &  3.94 & {\cellcolor{pos-red!20} 4.12} & {\cellcolor{pos-red!20} 3.46} &  4.44 & {\cellcolor{pos-red!20} 4.95} & {\cellcolor{pos-red!20} 3.09} & {\cellcolor{pos-red!20} 2.58} & {\cellcolor{pos-red!20} 5.80} & {\cellcolor{pos-red!20} 1.494} & {\cellcolor{pos-red!20} 5.17} &  4.51 &  4.155 & {\cellcolor{pos-red!20} 5.25} \\
gpt-b-nn & {\cellcolor{pos-red!20} 4.80} & {\cellcolor{pos-red!20} 3.98} & {\cellcolor{pos-red!20} 4.17} & {\cellcolor{pos-red!20} 3.47} & {\cellcolor{pos-red!20} 4.48} & {\cellcolor{pos-red!20} 4.91} & {\cellcolor{pos-red!20} 3.84} & {\cellcolor{pos-red!20} 2.74} & {\cellcolor{pos-red!20} 5.82} & {\cellcolor{pos-red!20} 1.554} & {\cellcolor{pos-red!20} 5.16} &  4.53 &  4.138 & {\cellcolor{pos-red!20} 5.26} \\
rnn-softplus &  4.74 & {\cellcolor{pos-red!20} 3.96} & {\cellcolor{pos-red!20} 4.36} & {\cellcolor{pos-red!20} 3.74} & {\cellcolor{pos-red!20} 5.32} & {\cellcolor{pos-red!20} 5.14} & {\cellcolor{pos-red!20} 4.00} & {\cellcolor{pos-red!20} 4.49} & {\cellcolor{pos-red!20} 5.80} & {\cellcolor{pos-red!20} 4.137} & {\cellcolor{pos-red!20} 5.17} & {\cellcolor{pos-red!20} 5.51} & {\cellcolor{pos-red!20} 4.419} & {\cellcolor{pos-red!20} 5.50} \\
gpt-a-softplus &  4.73 & {\cellcolor{pos-red!20} 3.95} & {\cellcolor{pos-red!20} 4.41} & {\cellcolor{pos-red!20} 3.76} & {\cellcolor{pos-red!20} 5.37} & {\cellcolor{pos-red!20} 5.14} & {\cellcolor{pos-red!20} 3.99} & {\cellcolor{pos-red!20} 4.51} & {\cellcolor{pos-red!20} 5.81} & {\cellcolor{pos-red!20} 4.142} & {\cellcolor{pos-red!20} 5.17} & {\cellcolor{pos-red!20} 5.47} & {\cellcolor{pos-red!20} 4.401} & {\cellcolor{pos-red!20} 5.56} \\
gpt-b-softplus &  4.74 & {\cellcolor{pos-red!20} 3.96} & {\cellcolor{pos-red!20} 4.41} & {\cellcolor{pos-red!20} 3.76} & {\cellcolor{pos-red!20} 5.39} & {\cellcolor{pos-red!20} 5.14} & {\cellcolor{pos-red!20} 3.99} & {\cellcolor{pos-red!20} 4.53} & {\cellcolor{pos-red!20} 5.81} & {\cellcolor{pos-red!20} 4.144} & {\cellcolor{pos-red!20} 5.17} & {\cellcolor{pos-red!20} 5.48} & {\cellcolor{pos-red!20} 4.399} & {\cellcolor{pos-red!20} 5.55} \\
rnn-exp &  4.73 & {\cellcolor{pos-red!20} 3.96} & {\cellcolor{pos-red!20} 4.35} & {\cellcolor{pos-red!20} 3.74} & {\cellcolor{pos-red!20} 5.34} & {\cellcolor{pos-red!20} 5.14} & {\cellcolor{pos-red!20} 4.00} & {\cellcolor{pos-red!20} 5.12} & {\cellcolor{pos-red!20} 5.80} & {\cellcolor{pos-red!20} 4.133} & {\cellcolor{pos-red!20} 5.17} & {\cellcolor{pos-red!20} 5.70} & {\cellcolor{pos-red!20} 4.421} & {\cellcolor{pos-red!20} 5.50} \\
gpt-a-exp &  4.73 & {\cellcolor{pos-red!20} 3.95} & {\cellcolor{pos-red!20} 4.40} & {\cellcolor{pos-red!20} 3.75} & {\cellcolor{pos-red!20} 5.39} & {\cellcolor{pos-red!20} 5.15} & {\cellcolor{pos-red!20} 3.99} & {\cellcolor{pos-red!20} 5.13} & {\cellcolor{pos-red!20} 5.81} & {\cellcolor{pos-red!20} 4.140} & {\cellcolor{pos-red!20} 5.18} & {\cellcolor{pos-red!20} 5.67} & {\cellcolor{pos-red!20} 4.417} & {\cellcolor{pos-red!20} 5.56} \\
gpt-b-exp &  4.74 & {\cellcolor{pos-red!20} 3.96} & {\cellcolor{pos-red!20} 4.40} & {\cellcolor{pos-red!20} 3.76} & {\cellcolor{pos-red!20} 5.41} & {\cellcolor{pos-red!20} 5.14} & {\cellcolor{pos-red!20} 3.99} & {\cellcolor{pos-red!20} 5.14} & {\cellcolor{pos-red!20} 5.81} & {\cellcolor{pos-red!20} 4.137} & {\cellcolor{pos-red!20} 5.17} & {\cellcolor{pos-red!20} 5.65} & {\cellcolor{pos-red!20} 4.401} & {\cellcolor{pos-red!20} 5.55} \\
rnn-const & {\cellcolor{pos-red!20} 4.75} & {\cellcolor{pos-red!20} 3.99} & {\cellcolor{pos-red!20} 4.58} & {\cellcolor{pos-red!20} 4.09} & {\cellcolor{pos-red!20} 5.90} & {\cellcolor{pos-red!20} 5.15} & {\cellcolor{pos-red!20} 4.01} & {\cellcolor{pos-red!20} 6.61} & {\cellcolor{pos-red!20} 5.82} & {\cellcolor{pos-red!20} 4.364} & {\cellcolor{pos-red!20} 5.17} & {\cellcolor{pos-red!20} 6.05} & {\cellcolor{pos-red!20} 4.573} & {\cellcolor{pos-red!20} 5.56} \\
gpt-a-const & {\cellcolor{pos-red!20} 4.75} & {\cellcolor{pos-red!20} 3.98} & {\cellcolor{pos-red!20} 4.62} & {\cellcolor{pos-red!20} 4.12} & {\cellcolor{pos-red!20} 6.00} & {\cellcolor{pos-red!20} 5.16} & {\cellcolor{pos-red!20} 4.00} & {\cellcolor{pos-red!20} 6.53} & {\cellcolor{pos-red!20} 5.82} & {\cellcolor{pos-red!20} 4.366} & {\cellcolor{pos-red!20} 5.17} & {\cellcolor{pos-red!20} 6.01} & {\cellcolor{pos-red!20} 4.524} & {\cellcolor{pos-red!20} 5.56} \\
gpt-b-const & {\cellcolor{pos-red!20} 4.75} & {\cellcolor{pos-red!20} 3.98} & {\cellcolor{pos-red!20} 4.62} & {\cellcolor{pos-red!20} 4.12} & {\cellcolor{pos-red!20} 6.02} & {\cellcolor{pos-red!20} 5.16} & {\cellcolor{pos-red!20} 4.00} & {\cellcolor{pos-red!20} 6.53} & {\cellcolor{pos-red!20} 5.82} & {\cellcolor{pos-red!20} 4.365} & {\cellcolor{pos-red!20} 5.17} & {\cellcolor{pos-red!20} 6.04} & {\cellcolor{pos-red!20} 4.523} & {\cellcolor{pos-red!20} 5.56} \\
thp-0 &  4.74 & {\cellcolor{pos-red!20} 3.95} & {\cellcolor{pos-red!20} 4.40} & {\cellcolor{pos-red!20} 3.76} & {\cellcolor{pos-red!20} 5.46} & {\cellcolor{pos-red!20} 5.17} & {\cellcolor{pos-red!20} 4.00} & {\cellcolor{pos-red!20} 5.19} & {\cellcolor{pos-red!20} 5.83} & {\cellcolor{pos-red!20} 4.142} & {\cellcolor{pos-red!20} 5.18} & {\cellcolor{pos-red!20} inf} & {\cellcolor{pos-red!20} inf} & {\cellcolor{pos-red!20} 5.53} \\
dtpp &  4.74 &  3.92 &  4.09 & {\cellcolor{pos-red!20} 3.50} &  4.39 & {\cellcolor{pos-red!20} 4.94} & {\cellcolor{pos-red!20} 3.57} & {\cellcolor{pos-red!20} 1.88} & {\cellcolor{pos-red!20} 5.80} & {\cellcolor{pos-red!20} 1.557} & {\cellcolor{pos-red!20} 5.13} &  4.51 &  4.139 & {\cellcolor{pos-red!20} 5.16} \\
\bottomrule
\end{tabular}
}
  \label{tbl:app_baseline_full_nll}
\end{sidewaystable*}

\begin{sidewaystable*}
\caption{Extension of \cref{tbl:baseline} in \cref{sec:results}. Test set MAE scores (lower is better) for various models on 14 real-world datasets. 6 output heads (\model{logmix}, \model{pc}, \model{nn}, \model{softplus}, \model{exp} and \model{const}) are paired with 3 model stems (\model{rnn}, \model{gpt-a} and \model{gpt-b}). Additionally, \model{thp} from \citet{zuoTransformerHawkesProcess2020} and \model{dtpp} from \citet{panosDecomposableTransformerPoint2024} are included (see \cref{app:othermodels}). Point estimates are the means over 10 repeats, where each repeat uses a random data split. The \model{rnn-logmix} model is used as a baseline, and cells are coloured when the score difference to the \model{rnn-logmix} model, \( \Delta \), is statistically significant with respect to the standard errors (SE) of the means, \( |\Delta| > 1.96 \sqrt{ \mathrm{SE}_{\text{logmix}}^2 + \mathrm{SE}_{\text{model}}^2} \); \textcolor{neg-green-text}{green} when better than \model{rnn-logmix}, \textcolor{pos-red}{red} when worse. \textbf{Bold} entries are the lowest (best) score point estimates per dataset.} 

  \centering
  \vspace{0.5em}
  {\setlength{\tabcolsep}{3pt}%
\begin{tabular}{lrrrrrrrrrrrrrr}
\toprule
Model & Yelp a. & Yelp M. & Twitter & Taobao & Wiki. & Yelp Tor. & PUBG & MOOC & Reddit Ask. & Amazon & Reddit Pol. & Last.fm & NYC taxi & Stack Overflow \\
\midrule
rnn-logmix & 34.68 & 21.36 & \textbf{37.23} & 21.30 & \textbf{198.2} & 62.82 & 18.32 & 290.5 & 189.3 & 13.58 & 53.89 & \textbf{667.0} & 27.48 & 128.7 \\
gpt-a-logmix &  34.55 &  21.42 &  37.96 &  21.41 &  199.8 &  62.94 & {\cellcolor{neg-green!20} 17.49} &  290.5 &  191.1 &  13.46 &  54.04 &  670.2 &  27.57 &  \textbf{128.6} \\
gpt-b-logmix &  \textbf{34.44} &  21.43 &  37.98 &  21.43 &  200.1 &  62.86 & {\cellcolor{neg-green!20} 17.49} &  290.5 &  189.3 &  13.47 &  53.77 &  672.0 &  27.48 &  128.9 \\
rnn-pc &  35.04 &  \textbf{21.30} &  37.34 &  21.31 &  198.4 &  \textbf{62.75} & {\cellcolor{neg-green!20} 17.51} &  \textbf{290.4} &  189.0 & {\cellcolor{neg-green!20} \textbf{13.41}} &  53.72 &  674.4 &  27.50 &  128.7 \\
gpt-a-pc &  34.70 &  21.55 & {\cellcolor{pos-red!20} 38.07} &  21.46 &  199.9 &  63.02 & {\cellcolor{neg-green!20} 17.46} &  290.5 &  190.0 &  13.47 &  53.63 &  673.6 &  \textbf{27.40} &  128.8 \\
gpt-b-pc &  34.59 &  21.43 & {\cellcolor{pos-red!20} 38.05} &  21.46 &  200.2 &  62.85 & {\cellcolor{neg-green!20} 17.44} &  290.5 &  190.4 &  13.46 &  53.64 &  673.7 &  27.43 &  129.0 \\
rnn-nn &  35.06 &  21.60 &  37.32 &  \textbf{21.29} &  198.5 &  63.12 & {\cellcolor{neg-green!20} 17.59} &  291.3 &  \textbf{188.9} & {\cellcolor{pos-red!20} 13.99} &  53.75 &  674.1 &  27.49 &  129.1 \\
gpt-a-nn &  34.70 &  21.44 &  37.88 &  21.38 &  200.0 &  63.58 & {\cellcolor{neg-green!20} 17.49} &  290.8 &  191.2 & {\cellcolor{pos-red!20} 13.98} &  53.92 &  680.6 &  27.62 &  129.8 \\
gpt-b-nn &  34.99 &  21.63 &  37.79 &  21.39 &  200.3 &  63.10 & {\cellcolor{neg-green!20} 17.50} &  290.9 &  193.1 & {\cellcolor{pos-red!20} 13.99} &  53.70 &  680.6 &  27.51 &  129.2 \\
rnn-softplus &  35.00 &  21.51 & {\cellcolor{pos-red!20} 38.28} & {\cellcolor{pos-red!20} 21.74} &  204.1 &  62.90 & {\cellcolor{neg-green!20} 17.54} &  291.9 &  189.1 & {\cellcolor{pos-red!20} 16.58} &  53.69 & {\cellcolor{pos-red!20} 722.8} &  27.71 & {\cellcolor{pos-red!20} 131.9} \\
gpt-a-softplus &  34.72 &  21.45 & {\cellcolor{pos-red!20} 38.77} & {\cellcolor{pos-red!20} 21.88} &  205.2 &  63.02 & {\cellcolor{neg-green!20} 17.46} &  291.1 &  190.3 & {\cellcolor{pos-red!20} 16.49} &  \textbf{53.55} &  695.0 &  27.86 & {\cellcolor{pos-red!20} 132.3} \\
gpt-b-softplus &  34.75 &  21.53 & {\cellcolor{pos-red!20} 38.65} & {\cellcolor{pos-red!20} 21.93} &  205.4 &  63.05 & {\cellcolor{neg-green!20} 17.46} &  291.1 &  190.3 & {\cellcolor{pos-red!20} 16.55} &  53.58 &  694.5 &  27.91 & {\cellcolor{pos-red!20} 132.2} \\
rnn-exp &  34.76 &  21.48 & {\cellcolor{pos-red!20} 38.17} & {\cellcolor{pos-red!20} 21.70} &  205.3 &  62.90 & {\cellcolor{neg-green!20} 17.53} & {\cellcolor{pos-red!20} 302.4} &  189.2 & {\cellcolor{pos-red!20} 16.54} &  53.70 &  701.3 &  27.72 & {\cellcolor{pos-red!20} 131.9} \\
gpt-a-exp &  34.66 &  21.45 & {\cellcolor{pos-red!20} 38.55} & {\cellcolor{pos-red!20} 21.87} &  205.6 &  63.45 & {\cellcolor{neg-green!20} \textbf{17.44}} & {\cellcolor{pos-red!20} 301.4} &  190.0 & {\cellcolor{pos-red!20} 16.60} &  54.03 &  699.2 &  27.88 & {\cellcolor{pos-red!20} 133.0} \\
gpt-b-exp &  34.77 &  21.44 & {\cellcolor{pos-red!20} 38.51} & {\cellcolor{pos-red!20} 21.90} &  207.0 &  63.06 & {\cellcolor{neg-green!20} 17.45} & {\cellcolor{pos-red!20} 301.9} &  190.2 & {\cellcolor{pos-red!20} 16.34} &  53.58 &  698.5 &  27.90 & {\cellcolor{pos-red!20} 132.1} \\
rnn-const &  35.39 & {\cellcolor{pos-red!20} 21.94} & {\cellcolor{pos-red!20} 42.27} & {\cellcolor{pos-red!20} 24.71} & {\cellcolor{pos-red!20} 248.3} &  63.03 & {\cellcolor{neg-green!20} 17.64} & {\cellcolor{pos-red!20} 448.3} &  192.1 & {\cellcolor{pos-red!20} 20.28} &  53.72 & {\cellcolor{pos-red!20} 781.3} & {\cellcolor{pos-red!20} 30.32} & {\cellcolor{pos-red!20} 136.2} \\
gpt-a-const &  35.02 & {\cellcolor{pos-red!20} 21.94} & {\cellcolor{pos-red!20} 43.38} & {\cellcolor{pos-red!20} 24.99} & {\cellcolor{pos-red!20} 254.2} &  63.32 & {\cellcolor{neg-green!20} 17.56} & {\cellcolor{pos-red!20} 450.9} &  192.9 & {\cellcolor{pos-red!20} 20.39} &  53.60 & {\cellcolor{pos-red!20} 785.7} & {\cellcolor{pos-red!20} 30.81} & {\cellcolor{pos-red!20} 136.7} \\
gpt-b-const &  34.99 & {\cellcolor{pos-red!20} 22.00} & {\cellcolor{pos-red!20} 43.12} & {\cellcolor{pos-red!20} 25.23} & {\cellcolor{pos-red!20} 250.6} &  63.32 & {\cellcolor{neg-green!20} 17.56} & {\cellcolor{pos-red!20} 458.5} &  193.4 & {\cellcolor{pos-red!20} 20.44} &  53.62 & {\cellcolor{pos-red!20} 810.1} & {\cellcolor{pos-red!20} 30.91} & {\cellcolor{pos-red!20} 137.0} \\
thp-0 & {\cellcolor{pos-red!20} 45.56} & {\cellcolor{pos-red!20} 26.28} & {\cellcolor{pos-red!20} 41.07} & {\cellcolor{pos-red!20} 23.10} &  202.6 & {\cellcolor{pos-red!20} 79.93} & {\cellcolor{pos-red!20} 22.94} &  290.6 & {\cellcolor{pos-red!20} 245.7} & {\cellcolor{pos-red!20} 28.45} & {\cellcolor{pos-red!20} 75.91} &  708.0 & {\cellcolor{pos-red!20} 38.60} & {\cellcolor{pos-red!20} 153.4} \\
dtpp &  34.73 &  21.36 &  37.77 &  21.35 &  199.3 &  63.00 & {\cellcolor{neg-green!20} 17.52} &  290.5 &  189.6 & {\cellcolor{neg-green!20} 13.44} &  53.68 &  671.9 &  27.47 &  129.1 \\
\bottomrule
\end{tabular}
}
  \label{tbl:app_baseline_full_mae}
\end{sidewaystable*}

\begin{table*}[h]
  \caption{Mean rank of models across all 14 datasets, in terms of NLL and MAE. The scores are taken from \cref{tbl:app_baseline_full_nll} and \cref{tbl:app_baseline_full_mae}. \textbf{Bold} entries are the lowest (best) mean rank for each metric (i.e. per column).}
\label{tbl:app_baseline_model_rank}
\centering
\vspace{0.5em}
{\setlength{\tabcolsep}{4pt}%
\begin{tabular}{lrrrr}
\toprule
 & \multicolumn{2}{c}{NLL rank $\downarrow$} & \multicolumn{2}{c}{MAE rank $\downarrow$} \\\cmidrule(lr){2-3} \cmidrule(lr){4-5}
Model & \# & mean & \# & mean \\
\midrule
rnn-logmix & 5 & 5.36 & 2 & 5.29 \\
gpt-a-logmix & 4 & 4.86 & 7 & 6.71 \\
gpt-b-logmix & 6 & 5.71 & 5 & 6.29 \\
\midrule
rnn-pc & 2 & 2.14 & 1 & \textbf{5.14} \\
gpt-a-pc & 1 & \textbf{1.50} & 6 & 6.64 \\
gpt-b-pc & 3 & 2.43 & 3 & 5.79 \\
\midrule
rnn-nn & 8 & 9.07 & 8 & 9.21 \\
gpt-a-nn & 9 & 9.14 & 9 & 9.64 \\
gpt-b-nn & 10 & 10.64 & 10 & 10.64 \\
\midrule
rnn-softplus & 11 & 12.79 & 14 & 11.71 \\
gpt-a-softplus & 13 & 12.93 & 11 & 10.79 \\
gpt-b-softplus & 14 & 13.86 & 15 & 11.86 \\
\midrule
rnn-exp & 11 & 12.79 & 12 & 11.29 \\
gpt-a-exp & 16 & 14.21 & 16 & 12.64 \\
gpt-b-exp & 14 & 13.86 & 13 & 11.57 \\
\midrule
rnn-const & 18 & 18.07 & 17 & 16.86 \\
gpt-a-const & 18 & 18.07 & 18 & 16.93 \\
gpt-b-const & 20 & 18.43 & 19 & 17.43 \\
\midrule
thp-0 & 17 & 16.29 & 20 & 17.79 \\
dtpp & 7 & 7.86 & 3 & 5.79 \\
\bottomrule
\end{tabular}}
\end{table*}

\begin{table*}[h]
  \caption{Mean rank of three model stems—\model{rnn}, \model{gpt-a} and \model{gpt-b}—in terms of NLL and MAE across the 14 real-world datasets. In addition to this rank (which appears in the final row), the mean rank is also calculated (for the same datasets) by first aggregating by output head. Each row has an NLL and a MAE ranking. The scores are taken from \cref{tbl:app_baseline_full_nll} and \cref{tbl:app_baseline_full_mae}. \textbf{Bold} entries are the lowest (best) mean rank across the three model stems (i.e. per row, but separate for the two metrics).} 
\label{tbl:app_baseline_stem_rank}
\centering
\vspace{0.5em}
{\setlength{\tabcolsep}{4pt}%
\begin{tabular}{lrrrrrr}
\toprule
 & \multicolumn{3}{c}{NLL rank $\downarrow$} & \multicolumn{3}{c}{MAE rank $\downarrow$} \\\cmidrule(lr){2-4} \cmidrule(lr){5-7}
Head & rnn & gpt-a & gpt-b & rnn & gpt-a & gpt-b \\
\midrule
logmix & 2.21 & \textbf{1.50} & 2.29 & \textbf{1.71} & 2.07 & 2.21 \\
pc & 2.07 & \textbf{1.50} & 2.43 & \textbf{1.71} & 2.07 & 2.21 \\
nn & 1.93 & \textbf{1.79} & 2.29 & \textbf{1.79} & 2.00 & 2.21 \\
softplus & \textbf{1.79} & \textbf{1.79} & 2.43 & 1.93 & \textbf{1.86} & 2.21 \\
exp & \textbf{1.64} & 2.21 & 2.14 & \textbf{1.79} & 2.14 & 2.07 \\
const & \textbf{1.93} & \textbf{1.93} & 2.14 & \textbf{1.50} & 2.00 & 2.50 \\
\midrule
all heads & 1.93 & \textbf{1.79} & 2.29 & \textbf{1.74} & 2.02 & 2.24 \\
\bottomrule
\end{tabular}}
\end{table*}

\clearpage

\section{Ablations} \label{app:app_ablations}
Five ablations were carried out: 
\begin{enumerate}
  \item \cref{app:edge_selection_ablation}: varying the method used to select edges.
  \item \cref{app:edge_count_ablation}: varying the number of edges.
  \item \cref{app:density_vs_hazard_ablation}: switching from a piecewise-constant density function to a piecewise-constant hazard function.
  \item \cref{app:precision_ablation}: perturbing event times to investigate the effect of narrow intervals.
  \item \cref{app:randproc_ablation}: using event sequences from a synthetic process (rather than recorded from real-world processes).
\end{enumerate}

\subsection{Edge selection method} \label{app:edge_selection_ablation} 
In addition to the main approach of this work, \cref{sec:interval_edges} also mentioned two other ways to select edges to form a piecewise distribution: equal-width and equal-quantile. In this section, we compare all three edge selection approaches. 

\begin{figure*}
  \captionof{table}{With \model{rnn-logmix} as a reference, the three edge selection methods are compared: \model{rnn-wpc} (equal-width), \model{rnn-qpc} (equal-quantile) and \model{rnn-pc} (maximum likelihood). All models share the same architecture up until the output representation. They are evaluated on 14 datasets in terms of test-set NLL and MAE. Values are means over 10 training repeats, where each repeat uses a random data split. Cells are coloured when the score difference to the \model{rnn-logmix} model, \( \Delta \), is statistically significant with respect to the standard errors (SE) of the means, \( |\Delta| > 1.96 \sqrt{ \mathrm{SE}_{\text{rnn-logmix}}^2 + \mathrm{SE}_{\text{model}}^2} \); 
\textcolor{neg-green-text}{green} when a model is better than \model{rnn-logmix}, \textcolor{pos-red}{red} when worse.}%
  \centering
{\setlength{\tabcolsep}{3pt}%
\begin{tabular}{lrrrrrrrr}
\toprule
 & \multicolumn{4}{c}{NLL $\downarrow$} & \multicolumn{4}{c}{MAE $\downarrow$} \\\cmidrule(lr){2-5} \cmidrule(lr){6-9}
Dataset & rnn-logmix & rnn-pc & rnn-qpc & rnn-wpc & rnn-logmix & rnn-pc & rnn-qpc & rnn-wpc \\
\midrule
\makebox[1.9cm][l]{Yelp airport} & 4.715 &  4.685 & {\cellcolor{pos-red!20} 4.779} & {\cellcolor{pos-red!20} 4.772} & 34.68 &  35.04 &  35.11 &  35.10 \\
\makebox[1.9cm][l]{Yelp Miss.} & 3.916 & {\cellcolor{neg-green!20} 3.260} & {\cellcolor{pos-red!20} 3.949} & {\cellcolor{pos-red!20} 3.968} & 21.36 &  21.30 &  21.64 &  21.52 \\
\makebox[1.9cm][l]{Twitter} & 4.043 & {\cellcolor{neg-green!20} 3.017} &  4.089 & {\cellcolor{pos-red!20} 4.173} & 37.23 &  37.34 &  37.54 &  37.50 \\
\makebox[1.9cm][l]{Wiki.} & 4.364 & {\cellcolor{neg-green!20} 1.962} &  4.385 & {\cellcolor{pos-red!20} 5.417} & 198.2 &  198.4 &  198.5 & {\cellcolor{pos-red!20} 211.6} \\
\makebox[1.9cm][l]{Yelp Toronto} & 4.810 & {\cellcolor{neg-green!20} 4.598} &  4.808 & {\cellcolor{pos-red!20} 5.005} & 62.82 &  62.75 &  62.75 &  62.79 \\
\makebox[1.9cm][l]{MOOC} & 1.023 & {\cellcolor{neg-green!20} -6.227} & {\cellcolor{neg-green!20} -4.189} & {\cellcolor{pos-red!20} 5.470} & 290.5 &  290.4 &  290.4 & {\cellcolor{pos-red!20} 341.9} \\
\makebox[1.9cm][l]{Amazon} & 1.439 & {\cellcolor{neg-green!20} 1.380} &  1.447 & {\cellcolor{pos-red!20} 1.464} & 13.58 & {\cellcolor{neg-green!20} 13.41} & {\cellcolor{neg-green!20} 13.41} & {\cellcolor{neg-green!20} 13.38} \\
\makebox[1.9cm][l]{Taobao} & 3.440 & {\cellcolor{neg-green!20} -2.010} & {\cellcolor{neg-green!20} 2.493} & {\cellcolor{pos-red!20} 3.494} & 21.30 &  21.31 &  21.33 &  21.34 \\
\makebox[1.9cm][l]{PUBG} & 1.643 & {\cellcolor{neg-green!20} -5.872} & {\cellcolor{neg-green!20} -3.021} & {\cellcolor{pos-red!20} 3.922} & 18.32 & {\cellcolor{neg-green!20} 17.51} & {\cellcolor{neg-green!20} 17.51} & {\cellcolor{neg-green!20} 17.52} \\
\makebox[1.9cm][l]{Reddit Politics} & 4.682 & {\cellcolor{neg-green!20} 0.288} & {\cellcolor{neg-green!20} 3.806} & {\cellcolor{pos-red!20} 5.155} & 53.89 &  53.72 &  53.70 &  53.71 \\
\makebox[1.9cm][l]{Reddit AskSci.} & 5.585 & {\cellcolor{neg-green!20} 2.150} &  5.486 & {\cellcolor{pos-red!20} 5.851} & 189.3 &  189.0 &  188.6 &  189.9 \\
\makebox[1.9cm][l]{Last.fm} & 4.465 & {\cellcolor{neg-green!20} 2.856} &  4.484 & {\cellcolor{pos-red!20} 6.571} & 667.0 &  674.4 &  667.7 & {\cellcolor{pos-red!20} 793.8} \\
\makebox[1.9cm][l]{NYC taxi} & 4.146 & {\cellcolor{neg-green!20} -1.295} & {\cellcolor{neg-green!20} 0.433} & {\cellcolor{pos-red!20} 4.199} & 27.48 &  27.50 &  27.50 &  27.49 \\
\makebox[1.9cm][l]{Stack Overflow} & 5.091 & {\cellcolor{neg-green!20} 4.941} &  5.119 & {\cellcolor{pos-red!20} 5.405} & 128.7 &  128.7 &  128.7 &  129.0 \\
\bottomrule
\end{tabular}}
\label{tbl:app_ablation_edge_selection}
\end{figure*}

\subsubsection{Equal-width and equal-quantile intervals. Details.} 
We implement the equal-width and equal-quantile edge selection methods as follows. For equal-width edge selection, the \( N - 2 \) central intervals split the region between  \( a_2 \) and \( a_N \) evenly by width. For equal-quantile edge selection, those same central intervals are split evenly by quantile. Quantiles are estimated from the training set following the method recommended by \citet{hyndmanSampleQuantilesStatistical1996}. In both equal-width and equal-quantile approaches, the first and last intervals are assigned length by quantile: they are given small probability masses \( m_1 \) and \( m_N \) (we use \( m_1 = m_N = \frac{1}{4N} \)) so they can cover the potentially long tails on either side of a distribution without being responsible for as many events as the central intervals.

\subsubsection{Results}
\cref{tbl:app_ablation_edge_selection} reports NLL and MAE results for the three different edge selection methods: 
\begin{enumerate}[itemsep=0em]
  \item equal-width intervals (\model{wpc})
  \item equal-quantile intervals (\model{qpc})
  \item intervals which maximize likelihood for an unconditional model (\model{pc}) (ours)
\end{enumerate}
All models are tested with the \model{rnn} stem and using 128 output intervals. The results for \model{rnn-logmix} are included for comparison. It is noteworthy that the piecewise-continuous density with equal-quantile edges also outperforms the mixture of lognormals (in terms of mean rank across datasets for both NLL and MAE). 

\subsection{Edge count} \label{app:edge_count_ablation}
The \model{rnn-pc} model is tested over a range of output bin counts from 32 to 2048, with results appearing in \cref{tbl:app_ablation_bins_nll} (NLL) and \cref{tbl:app_ablation_bins_mae} (MAE). The results are reported as means over 10 runs, where a run involves training and evaluating a model with a random data split. There is no bin count that is best across all datasets. A possible explanation for the heterogeneous results is that expressive power and susceptibility to overfitting are both increased with increasing bin count. In \cref{tbl:app_ablation_bins_nll_ci} and \cref{tbl:app_ablation_bins_mae_ci}, the same results are shown with confidence intervals to indicate variability over the 10 runs.
\begin{sidewaystable*}

\caption{The \model{rnn-pc} model with various output bin counts, evaluated in terms of test set NLL (lower is better) on 14 real-world datasets. Values are means over 10 repeats. Table (a) reports point estimates with the 128-bin model as a reference, and Table (b) reports point estimates with confidence intervals.}%
\label{tbl:app_ablation_bins_nll_both}

  \begin{subtable}{\textwidth}
    \caption{With the 128-bin model (\model{rnn-pc}) as a reference, the other bin count models are compared in terms of test set NLL. Values are means over 10 training repeats, where each repeat uses a random data split. Cells are coloured when the score difference to the \model{rnn-pc} model, \( \Delta \), is statistically significant with respect to the standard errors (SE) of the means, 
\( |\Delta| > 1.96 \sqrt{ \mathrm{SE}_{\text{rnn-pc}}^2 + \mathrm{SE}_{\text{model}}^2} \);
\textcolor{neg-green-text}{green} when a model is better than \model{rnn-pc}, \textcolor{pos-red}{red} when worse.}%
{\setlength{\tabcolsep}{3pt}%
\begin{tabular}{lrrrrrrrrrrrrrr}
\toprule
Model & Yelp a. & Yelp M. & Twitter & Taobao & Wiki. & Yelp Tor. & PUBG & MOOC & Reddit Ask. & Amazon & Reddit Pol. & Last.fm & NYC taxi & Stack Overflow \\
\midrule
rnn-pc & 4.69 & 3.26 & 3.02 & -2.01 & 1.96 & 4.60 & -5.87 & -6.23 & 2.15 & 1.380 & 0.29 & 2.86 & -1.295 & 4.94 \\
rnn-pc-32 &  4.71 & {\cellcolor{pos-red!20} 3.72} & {\cellcolor{pos-red!20} 3.84} & {\cellcolor{pos-red!20} 1.52} & {\cellcolor{pos-red!20} 3.90} & {\cellcolor{pos-red!20} 4.82} & {\cellcolor{pos-red!20} -1.33} & {\cellcolor{pos-red!20} -3.94} & {\cellcolor{pos-red!20} 4.59} &  \textbf{1.378} & {\cellcolor{pos-red!20} 3.41} & {\cellcolor{pos-red!20} 4.09} & {\cellcolor{pos-red!20} 1.168} & {\cellcolor{pos-red!20} 5.05} \\
rnn-pc-64 &  \textbf{4.67} & {\cellcolor{pos-red!20} 3.53} & {\cellcolor{pos-red!20} 3.50} & {\cellcolor{pos-red!20} -0.12} & {\cellcolor{pos-red!20} 3.15} & {\cellcolor{pos-red!20} 4.72} & {\cellcolor{pos-red!20} -3.77} & {\cellcolor{pos-red!20} -5.35} & {\cellcolor{pos-red!20} 3.55} &  1.378 & {\cellcolor{pos-red!20} 2.06} & {\cellcolor{pos-red!20} 3.61} & {\cellcolor{pos-red!20} -0.548} & {\cellcolor{pos-red!20} 4.98} \\
rnn-pc-256 &  4.70 & {\cellcolor{neg-green!20} 2.92} & {\cellcolor{neg-green!20} 2.36} & {\cellcolor{neg-green!20} -3.66} & {\cellcolor{neg-green!20} 0.48} & {\cellcolor{neg-green!20} 4.51} & {\cellcolor{neg-green!20} -7.59} & {\cellcolor{neg-green!20} -6.86} & {\cellcolor{neg-green!20} 0.45} &  1.382 & {\cellcolor{neg-green!20} -1.70} & {\cellcolor{neg-green!20} 1.74} &  -1.593 &  4.92 \\
rnn-pc-512 & {\cellcolor{pos-red!20} 4.73} & {\cellcolor{neg-green!20} 2.50} & {\cellcolor{neg-green!20} 1.63} & {\cellcolor{neg-green!20} -4.84} & {\cellcolor{neg-green!20} -0.96} & {\cellcolor{neg-green!20} 4.45} & {\cellcolor{neg-green!20} -8.61} & {\cellcolor{neg-green!20} -7.45} & {\cellcolor{neg-green!20} -1.30} &  1.385 & {\cellcolor{neg-green!20} -3.44} & {\cellcolor{neg-green!20} 0.51} & {\cellcolor{neg-green!20} -2.046} & {\cellcolor{neg-green!20} \textbf{4.90}} \\
rnn-pc-1024 & {\cellcolor{pos-red!20} 4.80} & {\cellcolor{neg-green!20} 2.32} & {\cellcolor{neg-green!20} 1.11} & {\cellcolor{neg-green!20} -5.42} & {\cellcolor{neg-green!20} -2.15} & {\cellcolor{neg-green!20} \textbf{4.41}} & {\cellcolor{neg-green!20} -8.95} & {\cellcolor{neg-green!20} -8.15} & {\cellcolor{neg-green!20} -2.77} & {\cellcolor{pos-red!20} 1.390} & {\cellcolor{neg-green!20} -4.68} & {\cellcolor{neg-green!20} -0.92} & {\cellcolor{neg-green!20} -2.918} & {\cellcolor{neg-green!20} 4.90} \\
rnn-pc-2048 & {\cellcolor{pos-red!20} 4.86} & {\cellcolor{neg-green!20} \textbf{2.26}} & {\cellcolor{neg-green!20} \textbf{1.03}} & {\cellcolor{neg-green!20} \textbf{-5.47}} & {\cellcolor{neg-green!20} \textbf{-2.91}} & {\cellcolor{neg-green!20} 4.42} & {\cellcolor{neg-green!20} \textbf{-8.97}} & {\cellcolor{neg-green!20} \textbf{-8.92}} & {\cellcolor{neg-green!20} \textbf{-3.78}} & {\cellcolor{pos-red!20} 1.394} & {\cellcolor{neg-green!20} \textbf{-5.35}} & {\cellcolor{neg-green!20} \textbf{-2.10}} & {\cellcolor{neg-green!20} \textbf{-4.227}} &  4.92 \\
\bottomrule
\end{tabular}
}
\label{tbl:app_ablation_bins_nll}
\vspace{2em}
\end{subtable}

\begin{subtable}{\textwidth}
  \caption{Confidence intervals for Table (a) above. 10 training repeats were carried out. Point estimates are the means over the 10 repeats. Variability is expressed as \( 95\% \) confidence intervals calculated assuming a normal distribution of the mean: \( \pm 1.96 \frac{s}{\sqrt{10}} \), where \( s \) is the sample standard deviation.} 
\centering
\vspace{0.5em}
{\setlength{\tabcolsep}{3pt}%
\begin{tabular}{lrrrrrrr}
\toprule
Dataset & rnn-pc & rnn-pc-32 & rnn-pc-64 & rnn-pc-256 & rnn-pc-512 & rnn-pc-1024 & rnn-pc-2048 \\
\midrule
Yelp a. & 4.685 $\pm$ 3.74e-02 & 4.712 $\pm$ 2.88e-02 & \textbf{4.674 $\pm$ 3.12e-02} & 4.703 $\pm$ 3.11e-02 & 4.731 $\pm$ 2.97e-02 & 4.796 $\pm$ 3.33e-02 & 4.865 $\pm$ 4.63e-02 \\
Yelp M. & 3.260 $\pm$ 3.89e-02 & 3.722 $\pm$ 2.91e-02 & 3.528 $\pm$ 4.38e-02 & 2.918 $\pm$ 6.16e-02 & 2.503 $\pm$ 7.14e-02 & 2.318 $\pm$ 8.99e-02 & \textbf{2.257 $\pm$ 8.04e-02} \\
Twitter & 3.017 $\pm$ 7.35e-02 & 3.836 $\pm$ 5.04e-02 & 3.497 $\pm$ 6.90e-02 & 2.363 $\pm$ 1.03e-01 & 1.631 $\pm$ 1.13e-01 & 1.105 $\pm$ 1.14e-01 & \textbf{1.031 $\pm$ 1.16e-01} \\
Wiki. & 1.962 $\pm$ 1.18e-01 & 3.898 $\pm$ 7.33e-02 & 3.148 $\pm$ 9.33e-02 & 0.485 $\pm$ 1.48e-01 & -0.961 $\pm$ 1.71e-01 & -2.149 $\pm$ 1.79e-01 & \textbf{-2.908 $\pm$ 1.82e-01} \\
Yelp Tor. & 4.598 $\pm$ 6.98e-02 & 4.819 $\pm$ 6.44e-02 & 4.715 $\pm$ 6.93e-02 & 4.509 $\pm$ 7.18e-02 & 4.446 $\pm$ 7.70e-02 & \textbf{4.408 $\pm$ 8.29e-02} & 4.421 $\pm$ 7.71e-02 \\
MOOC & -6.227 $\pm$ 2.90e-02 & -3.940 $\pm$ 3.23e-02 & -5.346 $\pm$ 2.90e-02 & -6.862 $\pm$ 2.59e-02 & -7.448 $\pm$ 2.44e-02 & -8.152 $\pm$ 2.28e-02 & \textbf{-8.915 $\pm$ 1.97e-02} \\
Amazon & 1.380 $\pm$ 5.49e-03 & \textbf{1.378 $\pm$ 5.55e-03} & 1.378 $\pm$ 5.29e-03 & 1.382 $\pm$ 5.60e-03 & 1.385 $\pm$ 5.54e-03 & 1.390 $\pm$ 5.80e-03 & 1.394 $\pm$ 5.69e-03 \\
Taobao & -2.010 $\pm$ 6.83e-02 & 1.519 $\pm$ 3.59e-02 & -0.124 $\pm$ 5.36e-02 & -3.662 $\pm$ 8.74e-02 & -4.839 $\pm$ 8.68e-02 & -5.425 $\pm$ 8.79e-02 & \textbf{-5.472 $\pm$ 8.80e-02} \\
PUBG & -5.872 $\pm$ 2.68e-02 & -1.333 $\pm$ 2.25e-02 & -3.771 $\pm$ 2.57e-02 & -7.585 $\pm$ 2.83e-02 & -8.609 $\pm$ 3.17e-02 & -8.947 $\pm$ 3.67e-02 & \textbf{-8.969 $\pm$ 3.47e-02} \\
Reddit Pol. & 0.288 $\pm$ 8.24e-02 & 3.413 $\pm$ 5.12e-02 & 2.063 $\pm$ 6.91e-02 & -1.697 $\pm$ 8.47e-02 & -3.441 $\pm$ 7.70e-02 & -4.680 $\pm$ 6.84e-02 & \textbf{-5.348 $\pm$ 6.64e-02} \\
Reddit Ask. & 2.150 $\pm$ 9.73e-02 & 4.586 $\pm$ 5.79e-02 & 3.549 $\pm$ 8.10e-02 & 0.445 $\pm$ 1.01e-01 & -1.296 $\pm$ 1.04e-01 & -2.770 $\pm$ 9.58e-02 & \textbf{-3.781 $\pm$ 8.39e-02} \\
Last.fm & 2.856 $\pm$ 2.26e-01 & 4.093 $\pm$ 1.73e-01 & 3.605 $\pm$ 2.09e-01 & 1.744 $\pm$ 2.49e-01 & 0.507 $\pm$ 2.60e-01 & -0.921 $\pm$ 2.62e-01 & \textbf{-2.097 $\pm$ 2.47e-01} \\
NYC taxi & -1.295 $\pm$ 4.27e-01 & 1.168 $\pm$ 2.49e-01 & -0.548 $\pm$ 3.69e-01 & -1.593 $\pm$ 4.35e-01 & -2.046 $\pm$ 4.04e-01 & -2.918 $\pm$ 3.46e-01 & \textbf{-4.227 $\pm$ 2.60e-01} \\
Stack Overflow & 4.941 $\pm$ 2.55e-02 & 5.050 $\pm$ 2.74e-02 & 4.983 $\pm$ 2.97e-02 & 4.917 $\pm$ 3.00e-02 & \textbf{4.903 $\pm$ 3.20e-02} & 4.904 $\pm$ 2.86e-02 & 4.923 $\pm$ 2.84e-02 \\
\bottomrule
\end{tabular}
}
\label{tbl:app_ablation_bins_nll_ci}
\end{subtable}

\end{sidewaystable*}

\begin{sidewaystable*}

\caption{The \model{rnn-pc} model with various output bin counts, evaluated in terms of test set MAE (lower is better) on 14 real-world datasets. Values are means over 10 repeats. Table (a) reports point estimates with the 128-bin model as a reference, and Table (b) reports point estimates with confidence intervals.}%
\label{tbl:app_ablation_bins_mae_both}

  \begin{subtable}{\textwidth}
    \caption{With the 128-bin model (\model{rnn-pc}) as a reference, the other bin count models are compared in terms of test set MAE. Values are means over 10 training repeats, where each repeat uses a random data split. Cells are coloured when the score difference to the \model{rnn-pc} model, \( \Delta \), is statistically significant with respect to the standard errors (SE) of the means, 
\( |\Delta| > 1.96 \sqrt{ \mathrm{SE}_{\text{rnn-pc}}^2 + \mathrm{SE}_{\text{model}}^2} \);
\textcolor{neg-green-text}{green} when a model is better than \model{rnn-pc}, \textcolor{pos-red}{red} when worse.}%
{\setlength{\tabcolsep}{3pt}%
\begin{tabular}{lrrrrrrrrrrrrrr}
\toprule
Model & Yelp a. & Yelp M. & Twitter & Taobao & Wiki. & Yelp Tor. & PUBG & MOOC & Reddit Ask. & Amazon & Reddit Pol. & Last.fm & NYC taxi & Stack Overflow \\
\midrule
rnn-pc & 35.04 & \textbf{21.30} & 37.34 & \textbf{21.31} & 198.4 & 62.75 & 17.51 & 290.4 & 189.0 & 13.41 & 53.72 & 674.4 & 27.50 & 128.7 \\
rnn-pc-32 &  \textbf{34.70} &  21.36 &  \textbf{37.31} &  21.42 &  198.5 &  \textbf{62.75} & {\cellcolor{pos-red!20} 17.64} &  290.4 &  190.2 &  \textbf{13.38} &  53.78 &  681.1 &  27.55 &  \textbf{128.7} \\
rnn-pc-64 &  34.79 &  21.36 &  37.32 &  21.33 &  \textbf{198.3} &  62.76 &  \textbf{17.51} &  \textbf{290.4} &  189.2 &  13.39 &  53.75 &  686.1 &  \textbf{27.50} &  128.7 \\
rnn-pc-256 &  35.38 &  21.39 &  37.37 &  21.32 &  198.4 &  62.75 &  17.51 &  290.4 &  188.7 &  13.43 &  53.71 &  668.8 &  27.50 &  128.7 \\
rnn-pc-512 &  35.38 &  21.57 &  37.39 &  21.34 &  198.4 &  62.76 &  17.52 &  290.4 &  \textbf{188.7} &  13.45 &  \textbf{53.71} &  668.4 &  27.50 &  128.7 \\
rnn-pc-1024 &  35.38 & {\cellcolor{pos-red!20} 22.05} &  37.48 &  21.36 &  198.5 &  62.82 &  17.52 &  290.5 &  188.8 &  13.47 &  53.72 &  \textbf{668.2} &  27.51 &  128.8 \\
rnn-pc-2048 &  35.38 & {\cellcolor{pos-red!20} 22.76} & {\cellcolor{pos-red!20} 38.92} &  21.53 &  198.6 &  62.92 &  17.52 &  290.5 &  188.9 & {\cellcolor{pos-red!20} 13.54} &  53.77 &  668.4 &  27.52 &  129.0 \\
\bottomrule
\end{tabular}
}
\label{tbl:app_ablation_bins_mae}
\vspace{2em}
\end{subtable}

\begin{subtable}{\textwidth}
\caption{Confidence intervals for Table (a) above. 10 training repeats were carried out. Point estimates are the means over the 10 repeats. Variability is expressed as \( 95\% \) confidence intervals calculated assuming a normal distribution of the mean: \( \pm 1.96 \frac{s}{\sqrt{10}} \), where \( s \) is the sample standard deviation.} 
\centering
\vspace{0.5em}
{\setlength{\tabcolsep}{3pt}%
\begin{tabular}{lrrrrrrr}
\toprule
Dataset & rnn-pc & rnn-pc-32 & rnn-pc-64 & rnn-pc-256 & rnn-pc-512 & rnn-pc-1024 & rnn-pc-2048 \\
\midrule
Yelp a. & 35.04 $\pm$ 7.75e-01 & \textbf{34.70 $\pm$ 8.23e-01} & 34.79 $\pm$ 8.00e-01 & 35.38 $\pm$ 7.42e-01 & 35.38 $\pm$ 7.43e-01 & 35.38 $\pm$ 7.40e-01 & 35.38 $\pm$ 7.49e-01 \\
Yelp M. & \textbf{21.30 $\pm$ 5.27e-01} & 21.36 $\pm$ 5.21e-01 & 21.36 $\pm$ 5.71e-01 & 21.39 $\pm$ 5.99e-01 & 21.57 $\pm$ 5.76e-01 & 22.05 $\pm$ 5.69e-01 & 22.76 $\pm$ 5.85e-01 \\
Twitter & 37.34 $\pm$ 6.28e-01 & \textbf{37.31 $\pm$ 6.21e-01} & 37.32 $\pm$ 6.19e-01 & 37.37 $\pm$ 6.29e-01 & 37.39 $\pm$ 6.32e-01 & 37.48 $\pm$ 6.33e-01 & 38.92 $\pm$ 6.28e-01 \\
Wiki. & 198.4 $\pm$ 7.44e+00 & 198.5 $\pm$ 7.37e+00 & \textbf{198.3 $\pm$ 7.39e+00} & 198.4 $\pm$ 7.42e+00 & 198.4 $\pm$ 7.39e+00 & 198.5 $\pm$ 7.42e+00 & 198.6 $\pm$ 7.41e+00 \\
Yelp Tor. & 62.75 $\pm$ 2.19e+00 & \textbf{62.75 $\pm$ 2.19e+00} & 62.76 $\pm$ 2.19e+00 & 62.75 $\pm$ 2.17e+00 & 62.76 $\pm$ 2.17e+00 & 62.82 $\pm$ 2.17e+00 & 62.92 $\pm$ 2.16e+00 \\
MOOC & 290.4 $\pm$ 4.08e+00 & 290.4 $\pm$ 4.07e+00 & \textbf{290.4 $\pm$ 4.09e+00} & 290.4 $\pm$ 4.08e+00 & 290.4 $\pm$ 4.08e+00 & 290.5 $\pm$ 4.07e+00 & 290.5 $\pm$ 4.07e+00 \\
Amazon & 13.41 $\pm$ 9.73e-02 & \textbf{13.38 $\pm$ 9.60e-02} & 13.39 $\pm$ 1.03e-01 & 13.43 $\pm$ 1.08e-01 & 13.45 $\pm$ 1.08e-01 & 13.47 $\pm$ 1.05e-01 & 13.54 $\pm$ 1.14e-01 \\
Taobao & \textbf{21.31 $\pm$ 3.17e-01} & 21.42 $\pm$ 3.17e-01 & 21.33 $\pm$ 3.14e-01 & 21.32 $\pm$ 3.17e-01 & 21.34 $\pm$ 3.14e-01 & 21.36 $\pm$ 3.20e-01 & 21.53 $\pm$ 3.19e-01 \\
PUBG & 17.51 $\pm$ 6.83e-02 & 17.64 $\pm$ 9.68e-02 & \textbf{17.51 $\pm$ 6.77e-02} & 17.51 $\pm$ 7.02e-02 & 17.52 $\pm$ 6.97e-02 & 17.52 $\pm$ 6.84e-02 & 17.52 $\pm$ 7.06e-02 \\
Reddit Pol. & 53.72 $\pm$ 8.30e-01 & 53.78 $\pm$ 8.30e-01 & 53.75 $\pm$ 8.34e-01 & 53.71 $\pm$ 8.36e-01 & \textbf{53.71 $\pm$ 8.36e-01} & 53.72 $\pm$ 8.37e-01 & 53.77 $\pm$ 8.37e-01 \\
Reddit Ask. & 189.0 $\pm$ 5.09e+00 & 190.2 $\pm$ 4.84e+00 & 189.2 $\pm$ 5.24e+00 & 188.7 $\pm$ 4.92e+00 & \textbf{188.7 $\pm$ 4.88e+00} & 188.8 $\pm$ 4.89e+00 & 188.9 $\pm$ 4.90e+00 \\
Last.fm & 674.4 $\pm$ 3.79e+01 & 681.1 $\pm$ 4.09e+01 & 686.1 $\pm$ 4.03e+01 & 668.8 $\pm$ 3.61e+01 & 668.4 $\pm$ 3.71e+01 & \textbf{668.2 $\pm$ 3.67e+01} & 668.4 $\pm$ 3.73e+01 \\
NYC taxi & 27.50 $\pm$ 9.10e-01 & 27.55 $\pm$ 9.11e-01 & \textbf{27.50 $\pm$ 9.11e-01} & 27.50 $\pm$ 9.12e-01 & 27.50 $\pm$ 9.10e-01 & 27.51 $\pm$ 9.09e-01 & 27.52 $\pm$ 9.10e-01 \\
Stack Overflow & 128.7 $\pm$ 1.25e+00 & \textbf{128.7 $\pm$ 1.25e+00} & 128.7 $\pm$ 1.26e+00 & 128.7 $\pm$ 1.25e+00 & 128.7 $\pm$ 1.26e+00 & 128.8 $\pm$ 1.25e+00 & 129.0 $\pm$ 1.28e+00 \\
\bottomrule
\end{tabular}
}
\label{tbl:app_ablation_bins_mae_ci}
\end{subtable}

\end{sidewaystable*}

\begin{figure}
  \centering
  \includegraphics[width=2.625in]{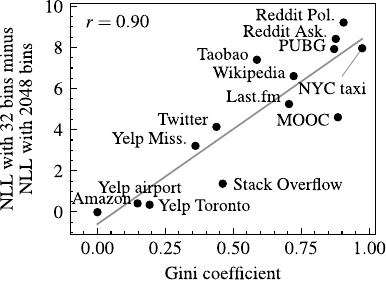}
  \caption{Gini coefficient for each dataset plotted against the NLL score difference obtained by two \model{rnn-pc} model variants: using 32 output bins and using 2048 output bins. Datasets with a higher Gini coefficient tend to benefit from the \model{rnn-pc} model using more output bins (higher NLL difference). This trend suggests that the Gini coefficient (and possibly other measures of distribution inequality) can be used to guide the choice of bin count for the \model{rnn-pc} model.}
\label{fig:app_gini_vs_bins}
\end{figure}

\subsection{Temporal precision} \label{app:precision_ablation}

\begin{figure*}
\centering
\captionsetup[subfigure]{skip=0pt, belowskip=3pt, justification=centering, margin=0cm}

\begin{subfigure}[t]{2.4in}
  \centering
  \includegraphics[width=\linewidth]{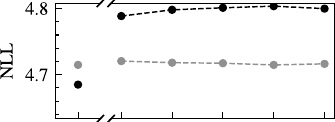}\hfill%
  \caption{\textbf{Yelp airport}}
  \label{fig:app_yelp_airport_jitter}
\end{subfigure}\hspace{0.33in}%
\begin{subfigure}[t]{2.31in}
  \centering
  \includegraphics[width=\linewidth]{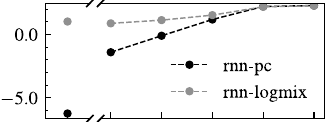}\hfill%
  \caption{\textbf{MOOC}}
  \label{fig:app_mooc_jitter}
\end{subfigure}\\
\begin{subfigure}[t]{2.4in}
  \centering
  \includegraphics[width=\linewidth]{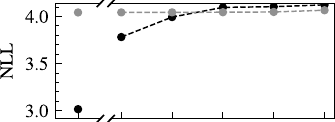}\hfill%
  \caption{\textbf{Twitter}}
  \label{fig:app_twitter_jitter}
\end{subfigure}\hspace{0.40in}%
\begin{subfigure}[t]{2.23in}
  \centering
  \includegraphics[width=\linewidth]{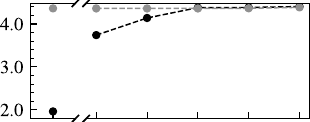}\hfill%
  \caption{\textbf{Wikipedia}}
  \label{fig:app_wikipedia_jitter}
\end{subfigure}\\
\begin{subfigure}[t]{2.4in}
  \centering
  \includegraphics[width=\linewidth]{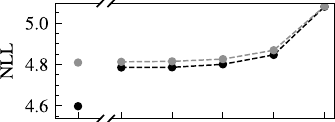}\hfill%
  \caption{\textbf{Yelp Toronto}}
  \label{fig:app_yelp_toronto_jitter}
\end{subfigure}\hspace{0.40in}%
\begin{subfigure}[t]{2.23in}
  \centering
  \includegraphics[width=\linewidth]{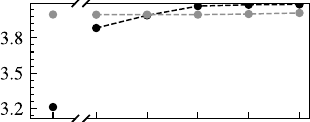}\hfill%
  \caption{\textbf{Yelp Mississauga}}
  \label{fig:app_yelp_mississauga_jitter}
\end{subfigure}\\
\begin{subfigure}[t]{2.4in}
  \centering
  \includegraphics[width=\linewidth]{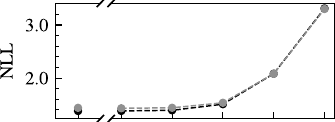}\hfill%
  \caption{\textbf{Amazon}}
  \label{fig:app_taobao_amazon}
\end{subfigure}\hspace{0.40in}%
\begin{subfigure}[t]{2.23in}
  \centering
  \includegraphics[width=\linewidth]{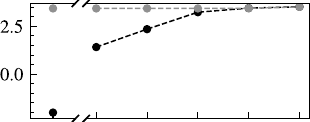}\hfill%
  \caption{\textbf{Taobao}}
  \label{fig:app_taobao_jitter}
\end{subfigure}\\
\begin{subfigure}[t]{2.4in}
  \centering
  \includegraphics[width=\linewidth]{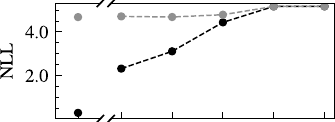}\hfill%
  \caption{\textbf{Reddit Politics}}
  \label{fig:app_redditpol_jitter}
\end{subfigure}\hspace{0.33in}%
\begin{subfigure}[t]{2.31in}
  \centering
  \includegraphics[width=\linewidth]{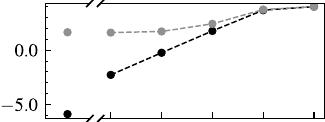}\hfill%
  \caption{\textbf{PUBG}}
  \label{fig:app_pubg_jitter}
\end{subfigure}\\
\begin{subfigure}[t]{2.4in}
  \centering
  \includegraphics[width=\linewidth]{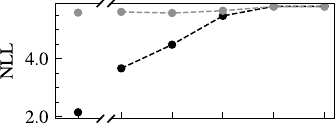}\hfill%
  \caption{\textbf{Reddit AskScience}}
  \label{fig:app_bubg_jitter}
\end{subfigure}\hspace{0.40in}%
\begin{subfigure}[t]{2.23in}
  \centering
  \includegraphics[width=\linewidth]{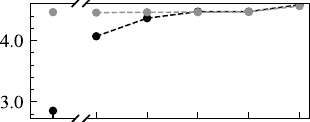}\hfill%
  \caption{\textbf{Last.fm}}
  \label{fig:app_lastfm_jitter}
\end{subfigure}\\
\captionsetup[subfigure]{skip=3pt, belowskip=0pt, justification=centering, margin=0cm}
\hspace{0.07in}\begin{subfigure}[t]{2.45in}
  \centering
  \includegraphics[width=\linewidth]{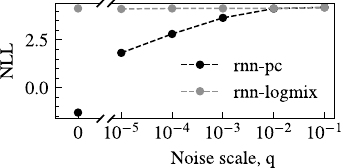}\hfill%
  \caption{\textbf{NYC taxi}}
  \label{fig:app_nyctaxi_jitter}
\end{subfigure}\hspace{0.35in}%
\begin{subfigure}[t]{2.29in}
  \centering
  \includegraphics[width=\linewidth]{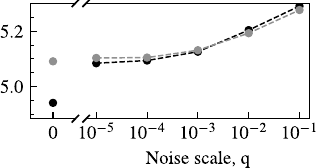}\hfill%
  \caption{\textbf{Stack Overflow badges}}
  \label{fig:app_sobadges_jitter}
\end{subfigure}
\caption{NLL scores for \model{rnn-pc} and \model{rnn-logmix} on 14 datasets with jittered event times. Event times are jittered by adding Gaussian noise with standard deviation proportional to the median inter-event time, \( \tilde{\tau} \). That is, \( t_{i}^* = t_{i} + \eta \) where \( \eta \sim \mathcal{N}(0, (\tilde{\tau}q)^2)  \). Five values of \( q \) are used, from \( 1\times 10^{-5} \) to \( 1 \times 10^{-1} \). Values are means over 10 repeats, where each repeat involves a new random data split and noise seed. Both models' scores improve under lower noise, but more so for \model{rnn-pc}.}
\label{fig:app_ds_jitter}
\end{figure*}

In this section, we perturb the event times with jitter in order to remove the reward for concentrating density in narrow intervals. \cref{fig:app_ds_jitter} shows the effect of jittering event times across all 14 datasets on the NLL scores for \model{rnn-pc} and \model{rnn-logmix}. Event times are jittered by adding Gaussian noise with standard deviation proportional to the median inter-event time, \( \tilde{\tau} \). That is, \( t_{i}^* = t_{i} + \eta \) where \( \eta \sim \mathcal{N}(0, (\tilde{\tau}q)^2)  \). Five values of \( q \) are used, from \( 1\times 10^{-5} \) to \( 1 \times 10^{-1} \). Both models' scores worsen under increased perturbation. 

The \model{rnn-pc} model's score continues to improve even when removing the lowest level of noise. This is evidence to support the hypothesis that the \model{rnn-pc} model is rewarded for concentrating density in narrow intervals. For multiple datasets, such as the NYC taxi dataset (\cref{fig:app_nyctaxi_jitter}), the \model{rnn-pc} begins to outperform the \model{rnn-logmix} even when moderate ($q=10^{-3}$) noise is present. This suggests that the \model{rnn-pc} does not rely on arbitrarily narrow intervals (down to float32 ULP) to outperform the \model{rnn-logmix} model. For most datasets, the \model{rnn-logmix} model's score improves to a plateau as noise is reduced.

\subsection{Density versus hazard} \label{app:density_vs_hazard_ablation}
\begin{table*}
  \caption{Density versus hazard. A piecewise-constant hazard output paired with the \model{rnn} model is compared to the \model{rnn-pc} model. Two piecewise-constant hazard outputs are considered: \model{rnn-qpch} and \model{rnn-pch}. They differ in the edge choice method—\model{rnn-qpch} uses the equal-quantile method and \model{rnn-pch} uses the maximum likelihood method. The models are evaluated on the 14 real-world datasets in terms of test-set NLL and MAE. All models share the same architecture up until the output representation. Values are means over 10 training repeats, where each repeat uses a random data split. Cells are coloured when the score difference to the \model{rnn-pc} model,
  \( \Delta \), is statistically significant with respect to the standard errors (SE) of the means, \( |\Delta| > 1.96 \sqrt{ \mathrm{SE}_{\text{rnn-pc}}^2 + \mathrm{SE}_{\text{rnn-}x}^2} \); 
\textcolor{neg-green-text}{green} when a model is better than \model{rnn-pc}, \textcolor{pos-red}{red} when worse.}%
\centering
{\setlength{\tabcolsep}{3pt}%
\begin{tabular}{lrrrrrr}
\toprule
 & \multicolumn{3}{c}{NLL $\downarrow$} & \multicolumn{3}{c}{MAE $\downarrow$} \\\cmidrule(lr){2-4} \cmidrule(lr){5-7}
Dataset & rnn-pc & rnn-qpch & rnn-pch & rnn-pc & rnn-qpch & rnn-pch \\
\midrule
Yelp airport & 4.685 & {\cellcolor{pos-red!20} 4.781} &  4.682 & 35.04 &  35.22 &  35.16 \\
Yelp Mississauga & 3.260 & {\cellcolor{pos-red!20} 3.947} &  3.234 & 21.30 &  21.78 &  21.61 \\
Twitter & 3.017 & {\cellcolor{pos-red!20} 4.089} &  2.998 & 37.34 &  37.55 &  37.37 \\
Wikipedia & 1.962 & {\cellcolor{pos-red!20} 4.376} &  1.927 & 198.4 &  198.7 &  198.4 \\
Yelp Toronto & 4.598 & {\cellcolor{pos-red!20} 4.805} &  4.646 & 62.75 &  62.76 &  62.77 \\
MOOC & -6.227 & {\cellcolor{pos-red!20} -4.189} & {\cellcolor{neg-green!20} -6.271} & 290.4 &  290.5 &  290.4 \\
Amazon & 1.380 & {\cellcolor{pos-red!20} 1.447} &  1.386 & 13.41 &  13.42 &  13.40 \\
Taobao & -2.010 & {\cellcolor{pos-red!20} 2.489} &  -2.041 & 21.31 &  21.33 &  21.32 \\
PUBG & -5.872 & {\cellcolor{pos-red!20} -3.040} &  -5.905 & 17.51 &  17.51 &  17.51 \\
Reddit Politics & 0.288 & {\cellcolor{pos-red!20} 3.800} &  0.245 & 53.72 &  53.70 &  53.71 \\
Reddit AskScience & 2.150 & {\cellcolor{pos-red!20} 5.477} &  2.204 & 189.0 &  188.5 &  188.6 \\
Last.fm & 2.856 & {\cellcolor{pos-red!20} 4.539} &  3.069 & 674.4 &  671.3 &  671.5 \\
NYC taxi & -1.295 & {\cellcolor{pos-red!20} 0.428} &  -1.298 & 27.50 &  27.50 &  27.50 \\
Stack Overflow & 4.941 & {\cellcolor{pos-red!20} 5.116} &  4.935 & 128.7 &  128.7 &  128.7 \\
\bottomrule
\end{tabular}}
\label{tbl:app_ablation_hazard}
\end{table*}

Outputting a density function or a hazard function over a set of intervals are similar in a very specific sense: a piecewise-constant hazard function maps deterministically to a piecewise-\textit{exponential} density function with the same partition edges. In terms of the density function, only the within-interval shape differs. A further difference is that loss through one logit of a piecewise density function propagates to all other logits through the normalization of the softmax operation, while only outputs preceding the target time receive a loss signal in the case of the piecewise hazard function. Both approaches share the benefit of high expressive power without training instability (see \cref{app:model_limitations}). 
A minor detraction from the piecewise hazard function in terms of elegance is that it must be mapped to a density function for loss calculation and for the NLL and MAE evaluations. On the other hand, the piecewise hazard is more elegant in terms of its final interval, which does not need a separate form, as a constant hazard over an infinite tail interval maps to an exponential density. 

In \cref{tbl:app_ablation_hazard}, we compare the \model{pc} head to two hazard-based heads; one head uses edges from the equal-quantile based edge selection method (labeled as \model{qpch}) and the other head uses the likelihood-maximizing edge selection method (labeled as \model{pch}). The performance scores suggest that these two approaches are broadly similar in terms of effectiveness. 

For both edge selection methods, switching to a piecewise hazard function yields mixed results: improved NLL in aggregate (mean rank across datasets) but consistently worse-or-equal MAE. The differences, however, are small in comparison to the differences seen when switching to another head (e.g. \model{logmix} or \model{nn}). For the equal-quantile edges, the differences are especially small. This is some evidence in support of the claim that using a piecewise hazard function leads to NLL and MAE scores broadly similar to those of a piecewise density function with the same edges.

\subsection{Event sequence from a simulated Hawkes process} \label{app:randproc_ablation}
The datasets considered in the main text are all recorded from real-world processes. In this section we test the models on a synthetic continuous dataset—inter-event times sampled from a Hawkes process. \cref{tbl:app_randproc_ci} shows the results for the next event prediction task on the Hawkes process dataset. Models with a \model{logmix} are the most performant in terms of NLL and MAE, with the \model{pc} head second, and the \model{nn} head third. These results show that the \model{pc} head performance does not become deleterious on truly continuous datasets. 

\subsubsection{Dataset generation}
The Hawkes process is defined by the following conditional hazard function:

\begin{equation}
\lambda(t|H_t) = \mu + \sum_{t_i<t}\sum_{j=1}^{M} \alpha_j\beta_je^{-\beta_j(t-t_i)}
\end{equation}

We follow \citet{omiFullyNeuralNetwork2019} and set parameters as follows: $M=2, \mu = 0.2, \alpha_1 = 0.4, \beta_1 = 1, \alpha_2 = 0.4,$ and $\beta_2 = 20$. Being synthetic, we can generate sequences of any length. We generate a sequence of $\approx 2^25$ events, but use just $2^{20}$ events for training, 100k for evaluation and 1M for test. The training events are taken from the beginning of the sequence, and the events for the validation and test sets and taken from the end. 10 sequences were generated (one per training repeat) each with a different random seed.

\begin{table*}[h]
  \caption{Hawkes process dataset results. NLL and MAE results for all models. 10 training repeats were carried out, where variability is in the random dataset split, random model initialization and training stochasticity. Point estimates are means over the repeats, and variability is expressed as $95\%$ confidence intervals calculated assuming a normal distribution of the mean: $\pm 1.96 \frac{s}{\sqrt{10}}$, where $s$ is the sample standard deviation. The Hawkes process is defined by a conditional hazard function: $\lambda(t|H_t) = \mu + \sum_{t_i<t}\sum_{j=1}^{M} \alpha_j\beta_j\exp{-\beta_j(t-t_i)}$, with the parameters set as in \citet{omiFullyNeuralNetwork2019} to $M=2, \mu = 0.2, \alpha_1 = 0.4, \beta_1 = 1, \alpha_2 = 0.4,$ and $\beta_2 = 20$. \textbf{Bold} highlights the lowest (best) point estimate.
}
\label{tbl:app_randproc_ci}
\centering
\vspace{0.5em}
\begin{tabular}{lrr}
\toprule
model & NLL $\downarrow$ & MAE $\downarrow$ \\
\midrule
rnn-logmix & -0.036 $\pm$ 2.32e-03 & \textbf{0.954 $\pm$ 1.56e-03} \\
gpt-a-logmix & \textbf{-0.036 $\pm$ 3.75e-03} & 0.954 $\pm$ 2.44e-03 \\
gpt-b-logmix & -0.035 $\pm$ 3.60e-03 & 0.954 $\pm$ 2.42e-03 \\
rnn-pc & -0.029 $\pm$ 3.74e-03 & 0.955 $\pm$ 2.43e-03 \\
gpt-a-pc & -0.034 $\pm$ 3.75e-03 & 0.954 $\pm$ 2.45e-03 \\
gpt-b-pc & -0.031 $\pm$ 3.70e-03 & 0.954 $\pm$ 2.47e-03 \\
rnn-nn & -0.015 $\pm$ 3.71e-03 & 0.955 $\pm$ 2.46e-03 \\
gpt-a-nn & 0.018 $\pm$ 1.01e-02 & 0.964 $\pm$ 2.45e-03 \\
gpt-b-nn & -0.026 $\pm$ 4.54e-03 & 0.954 $\pm$ 2.78e-03 \\
rnn-softplus & 0.370 $\pm$ 3.82e-03 & 0.980 $\pm$ 2.50e-03 \\
gpt-a-softplus & 0.359 $\pm$ 3.87e-03 & 0.976 $\pm$ 2.25e-03 \\
gpt-b-softplus & 0.368 $\pm$ 4.27e-03 & 0.974 $\pm$ 2.96e-03 \\
rnn-exp & 0.358 $\pm$ 3.86e-03 & 0.976 $\pm$ 2.74e-03 \\
gpt-a-exp & 0.355 $\pm$ 4.30e-03 & 0.975 $\pm$ 2.13e-03 \\
gpt-b-exp & 0.358 $\pm$ 4.42e-03 & 0.974 $\pm$ 2.74e-03 \\
rnn-const & 0.611 $\pm$ 4.21e-03 & 1.084 $\pm$ 3.84e-03 \\
gpt-a-const & 0.609 $\pm$ 4.57e-03 & 1.085 $\pm$ 5.98e-03 \\
gpt-b-const & 0.624 $\pm$ 4.37e-03 & 1.070 $\pm$ 5.29e-03 \\
thp-0 & 0.672 $\pm$ 2.48e-02 & 1.016 $\pm$ 3.24e-03 \\
dtpp & -0.026 $\pm$ 3.85e-03 & 0.955 $\pm$ 2.48e-03 \\
\bottomrule
\end{tabular}
\end{table*}

\clearpage
\section{Datasets} \label{app:datasets}
This appendix describes the 14 datasets used throughout this work. \cref{tbl:app_dataset_sources} lists the types of events and the data sources for each dataset.

\begin{table}[htbp]
  \centering
\caption{Sources of the event sequences. The unit column reports the units used in this work, which may differ from the units used in the original source (see \cref{app:dataset_units}).}
  \label{tbl:app_dataset_sources}
  \begin{tabularx}{\linewidth}{l X l p{1.2in}}
    \toprule
    Dataset & Event type & Unit & Source \\
    \midrule
    Yelp airport & Time between check-ins at McCarran International Airport. & minutes  & \citep{ludkeAddThinDiffusion2023} \\
    Yelp Mississauga & Time between check-ins at businesses in Mississauga. & minutes  & \citep{ludkeAddThinDiffusion2023} \\
    Twitter & Time between tweets of a single user. & minutes &  \citep{ludkeAddThinDiffusion2023} \\
    Wikipedia & Time between edits of the site's top edited pages. & minutes  & \citep{shchurIntensityFreeLearningTemporal2019} \\
    Yelp Toronto & Time between check-ins at restaurants in Toronto. & hours  & \citep{shchurIntensityFreeLearningTemporal2019} \\
    MOOC & Time between user interactions on an online course system. & minutes  & \citep{shchurIntensityFreeLearningTemporal2019, kumarPredictingDynamicEmbedding2019} \\
    Amazon & Time between user interactions on Amazon. & days \footnote{}  & \citep{xueEasyTPPOpenBenchmarking2024} \\
    Taobao & Time between user interactions on Taobao. & minutes  & \citep{xueEasyTPPOpenBenchmarking2024, xueHYPROHybridlyNormalized2022} \\
    PUBG & Time between player deaths in games of Player Unknown's Battleground (PUBG). & seconds  & \citep{ludkeAddThinDiffusion2023} \\
    Reddit Politics & Time between posts on the politics subreddit. & seconds  & \citep{ludkeAddThinDiffusion2023} \\
    Reddit AskScience & Time between comments on the askscience subreddit. & seconds  & \citep{ludkeAddThinDiffusion2023} \\
    Last.fm & Time between tracks played, per user. & minutes  & \citep{shchurIntensityFreeLearningTemporal2019} \\
    NYC taxi & Time between taxi pickup times in New York City, recorded per medallion. & minutes  & \citep{whongNYCsTaxiTrip2014} \\
    Stack Overflow & Time between the awarding of badges on Stack Overflow, per user. & days  & \citep{soDataDump2024} \\
    \bottomrule
  \end{tabularx}
\end{table}
\footnotetext{Converted from what was estimated, by inspecting the event times, to be units of \textasciitilde 60 days.}

\subsection{Dataset histograms} \label{app:dataset_histograms}
The histograms of event times for all of the datasets are shown in \cref{fig:app_ds_histA} and \cref{fig:app_ds_histB}. Two x-axis ranges are used: up to the 95th percentile and up to the 60th percentile of the inter-event times. When zooming in to the 60th percentile, it is easier to see discrete components present in the distributions.

\begin{figure*}
\centering
\includegraphics[width=0.9\linewidth]{"out_imgs/theoretical_density_legend.pdf"}%
\par\vspace{0.10in}%
\begin{subfigure}[t]{2.881in}
    \includegraphics[width=\linewidth, trim=20pt 0 0 0, clip]{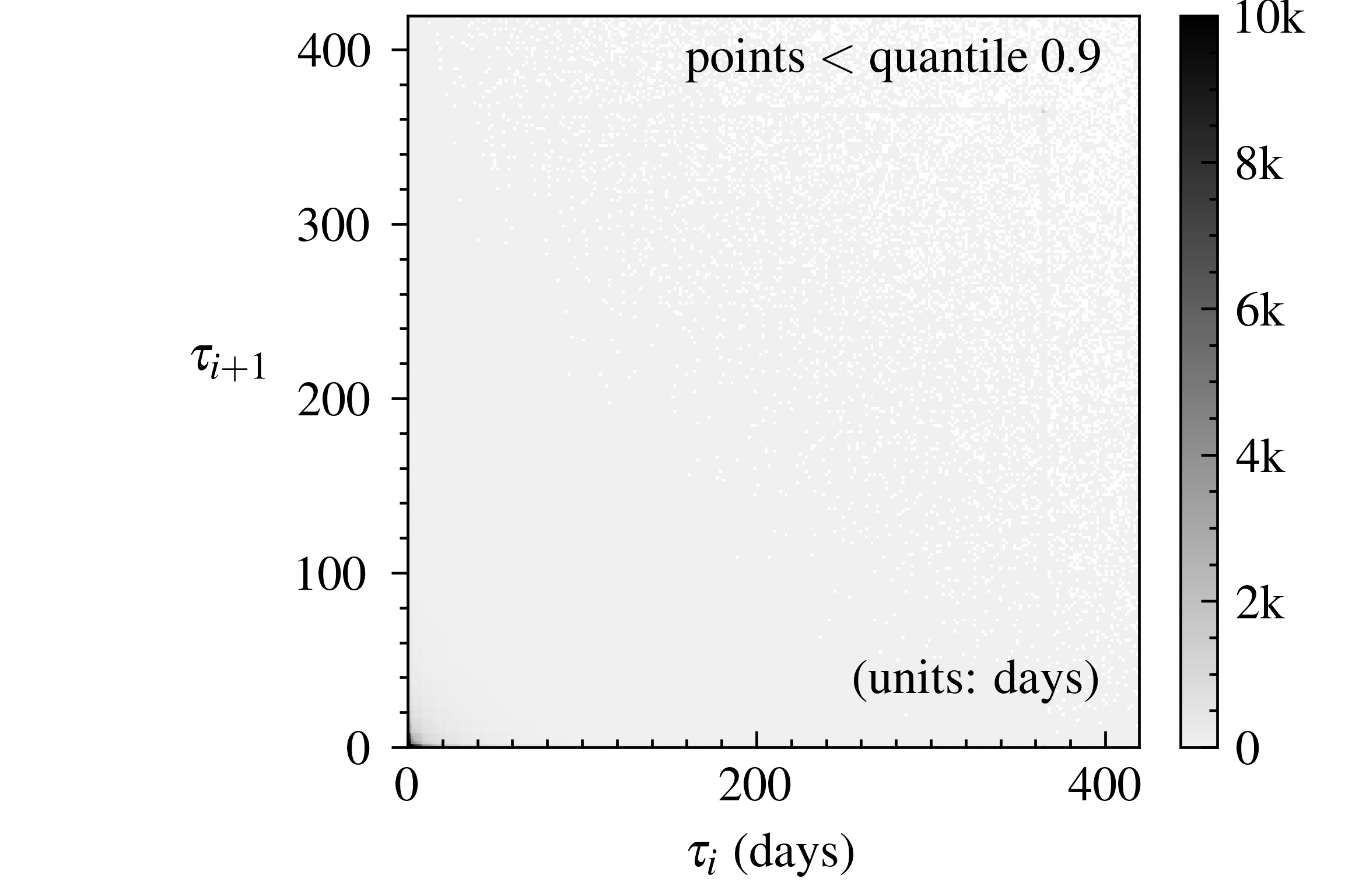}
  \end{subfigure}\hfill%
\begin{subfigure}[t]{2.619in}
    \includegraphics[width=\linewidth, trim=40pt 0 0 0, clip]{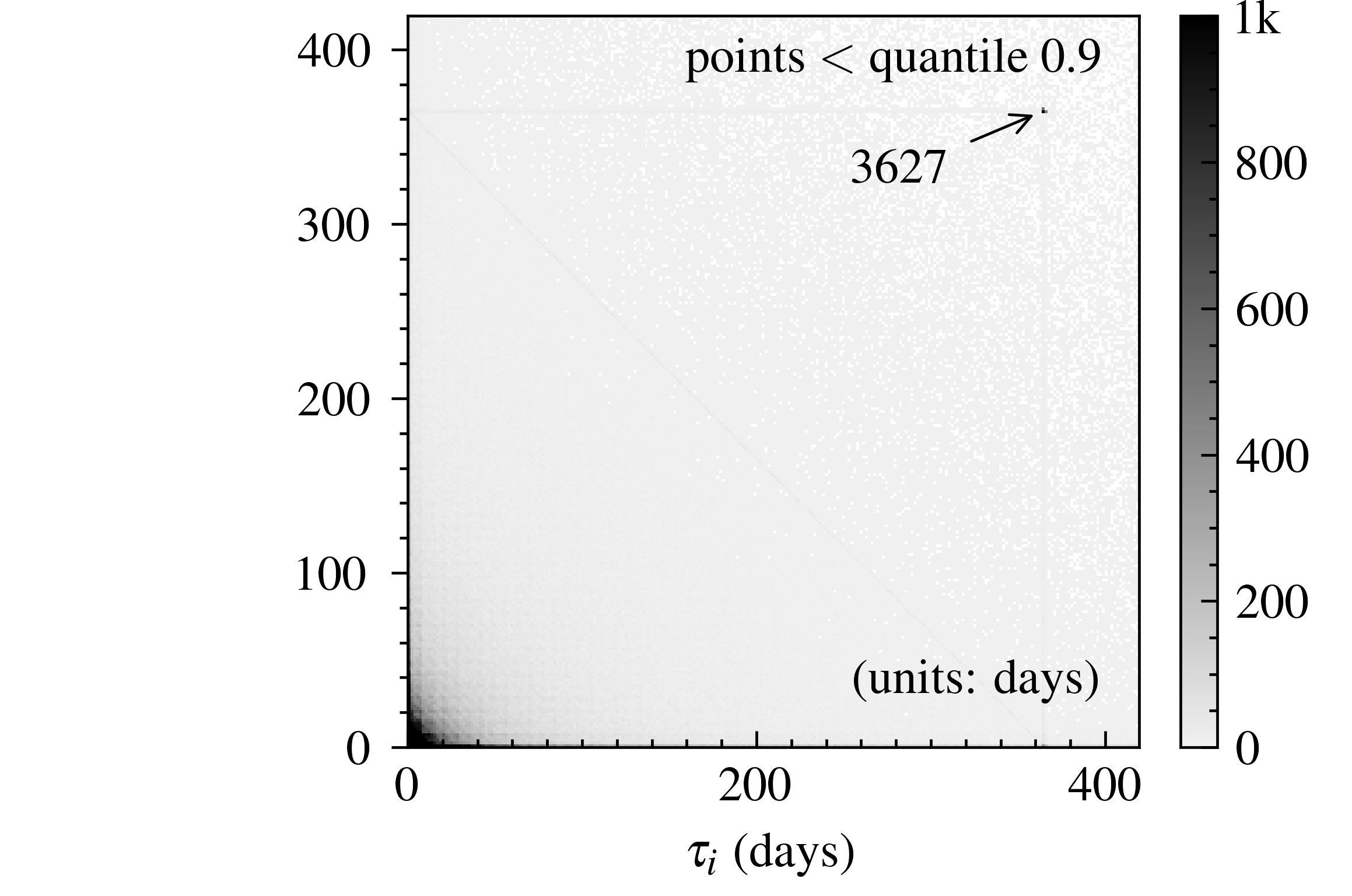}
\end{subfigure}%
\caption{Distribution of pairs of inter-event times, \( \tau_i \) and \( \tau_{i+1} \) for the \textbf{Stack Overflow} dataset. The \textbf{right} figure is a clipped version of the \textbf{left} figure. 
  The peak at \textasciitilde 365 days is small relative to the overall density, but is visible in the \textbf{right} figure when the maximum colormap value is reduced. This peak presumably corresponds to a badge typically awarded yearly.}
\label{fig:app_ds_hist2d_so}
\end{figure*}

Conditional histograms in \cref{fig:app_ds_hist2d_discrete} show the grid nature of some of the datasets. Conditional histograms in \cref{fig:app_ds_hist2d_so} show how even the Stack Overflow badges dataset, which appears to have a relatively smooth distribution, also has a narrow point density associated with yearly badges.

\begin{table}[h!]
  \caption{Gini coefficients for the distribution of inter-event times of the real-world datasets.} 
\label{tbl:app_gini}
\centering
\vspace{0.5em}
\begin{tabular}{lr}
\toprule
Dataset & Gini coefficient \\
\midrule
Amazon & 0.000 \\
Stack Overflow & 0.141 \\
Yelp airport & 0.148 \\
Yelp Toronto & 0.192 \\
Yelp Mississauga & 0.361 \\
Twitter & 0.438 \\
Taobao & 0.581 \\
Last.fm & 0.705 \\
Wikipedia & 0.722 \\
PUBG & 0.871 \\
Reddit AskScience & 0.877 \\
MOOC & 0.885 \\
Reddit Politics & 0.906 \\
NYC taxi & 0.926 \\
\bottomrule
\end{tabular}
\end{table}

\subsection{Gini coefficient}
\begin{figure*}
\centering
\captionsetup[subfigure]{skip=2pt, belowskip=3pt, justification=centering, margin=0cm}

\begin{subfigure}[t]{5.5in}
  \centering
  \includegraphics[width=2.85in, trim=0      0.18in 0.3in 0, clip]{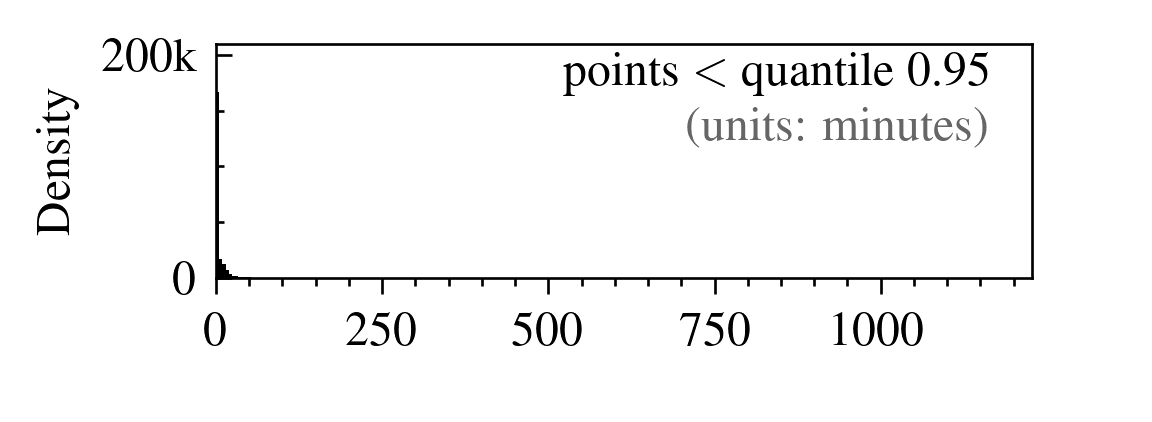}\hfill%
  \includegraphics[width=2.64in, trim=0.20in 0.18in 0.3in 0, clip]{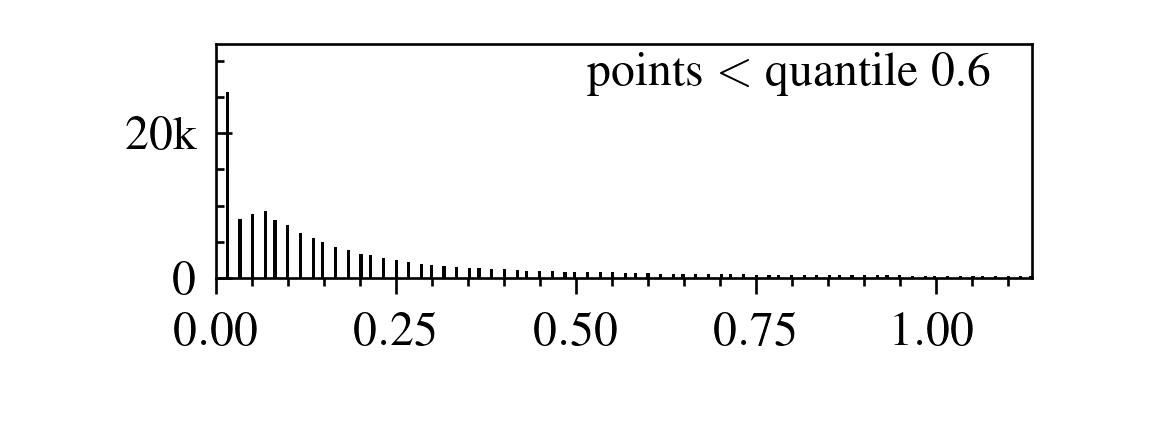}
  \caption{\textbf{MOOC} (time between user interactions)}
\end{subfigure}%

\begin{subfigure}[t]{5.5in}
  \centering
  \includegraphics[width=2.85in, trim=0      0.18in 0.3in 0, clip]{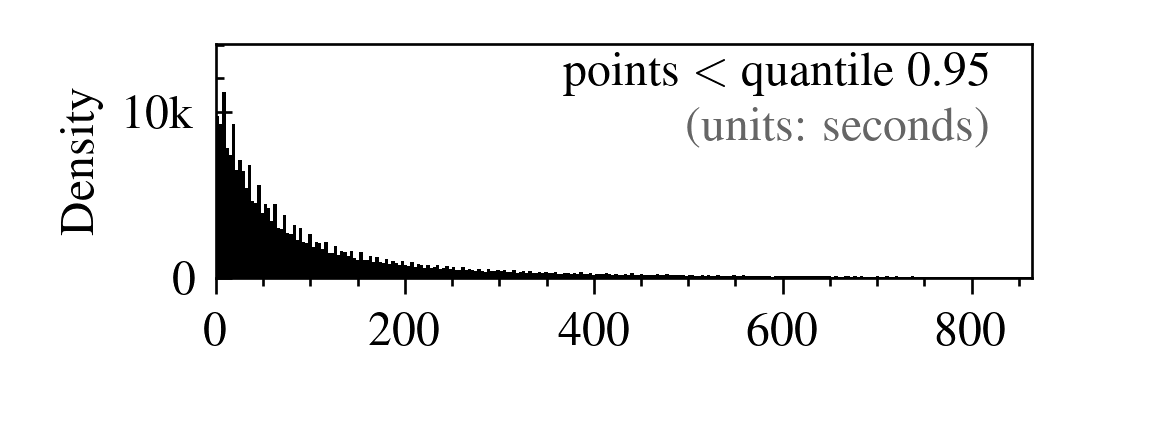}\hfill%
  \includegraphics[width=2.64in, trim=0.20in 0.18in 0.3in 0, clip]{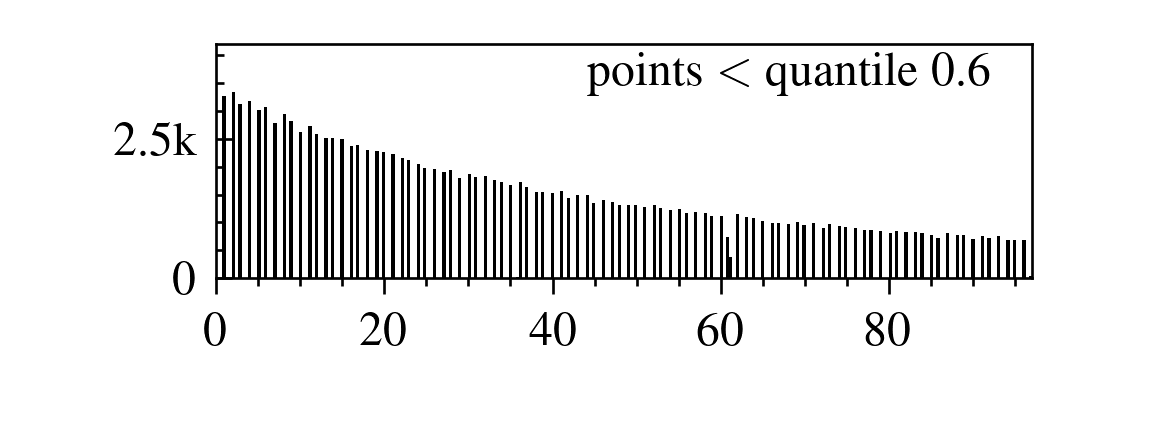}
  \caption{\textbf{Reddit AskScience} (time between comments)}
\end{subfigure}%

\begin{subfigure}[t]{5.5in}
  \centering
  \includegraphics[width=2.85in, trim=0      0.18in 0.3in 0, clip]{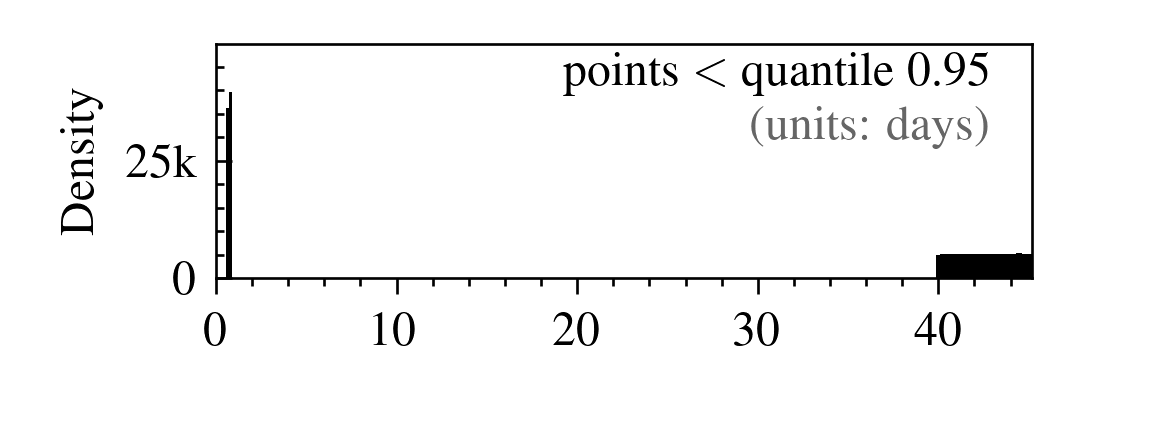}\hfill%
  \includegraphics[width=2.64in, trim=0.20in 0.18in 0.3in 0, clip]{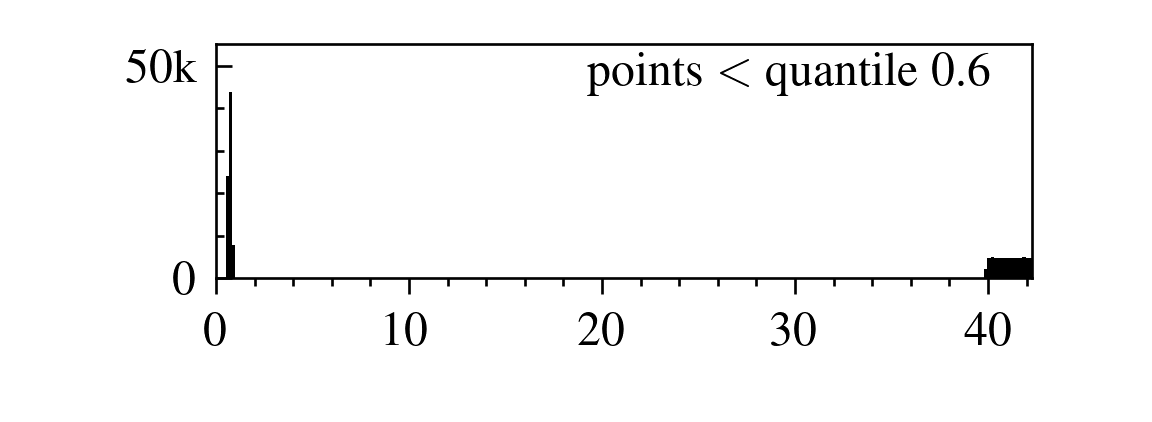}
  \caption{\textbf{Amazon user} (time between user interactions)}
\end{subfigure}%

\begin{subfigure}[t]{5.5in}
  \centering
  \includegraphics[width=2.85in, trim=0      0.18in 0.3in 0, clip]{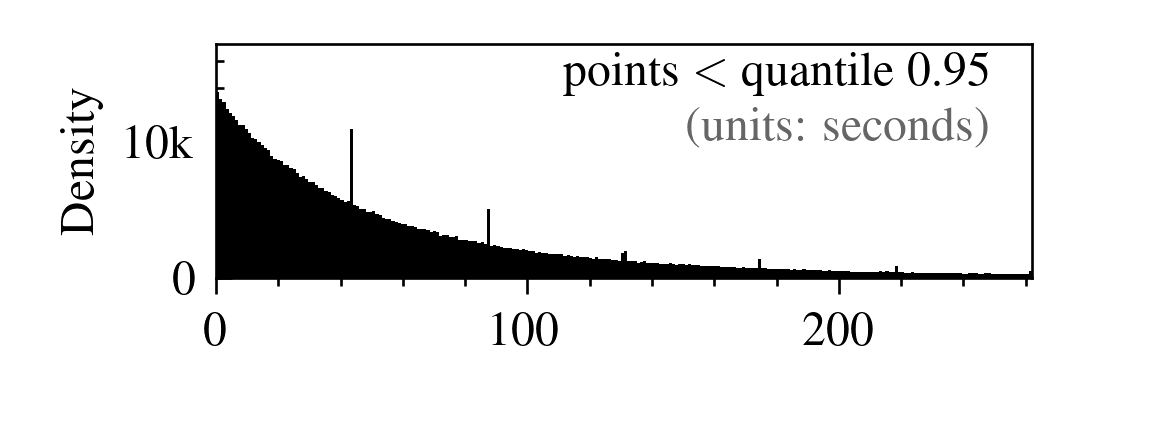}\hfill%
  \includegraphics[width=2.64in, trim=0.20in 0.18in 0.3in 0, clip]{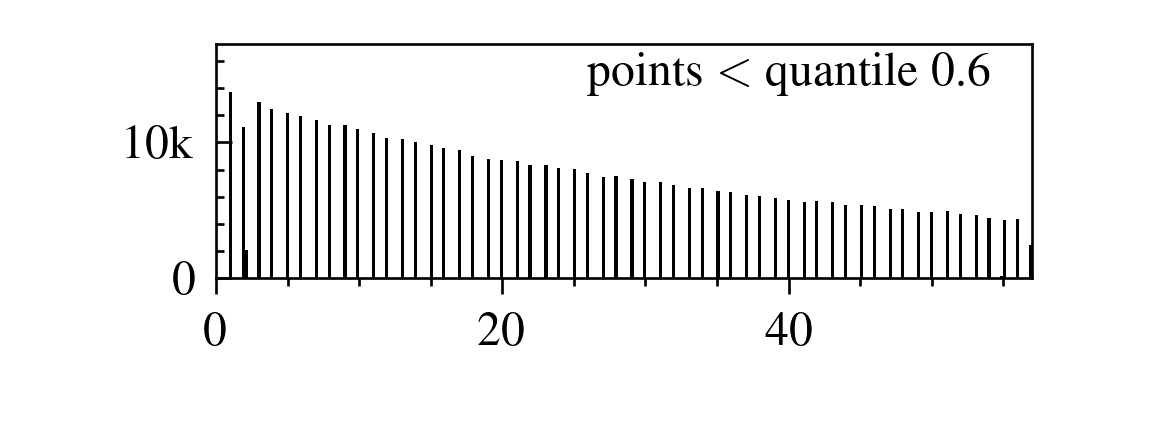}
  \caption{\textbf{Reddit Politics} (time between posts)}
\end{subfigure}%

\begin{subfigure}[t]{5.5in}
  \centering
  \includegraphics[width=2.85in, trim=0      0.18in 0.3in 0, clip]{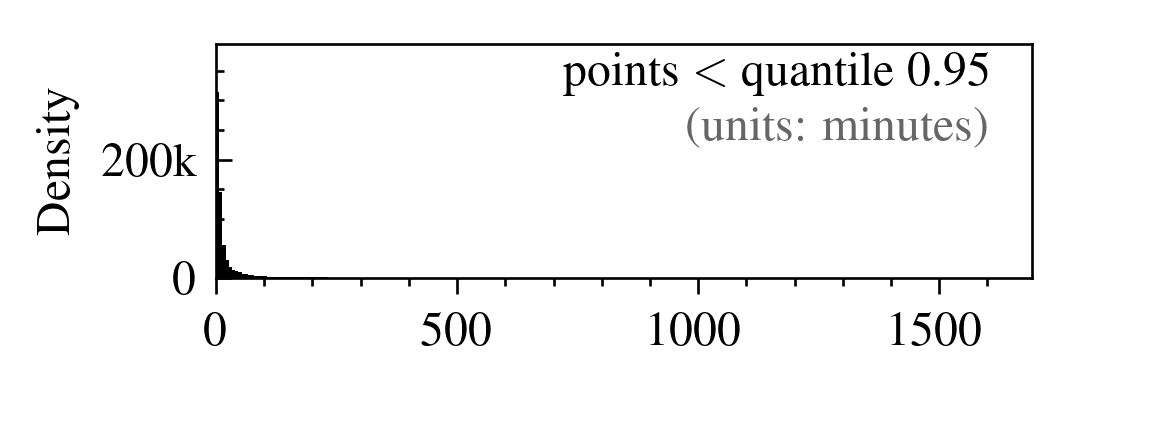}\hfill%
  \includegraphics[width=2.64in, trim=0.20in 0.18in 0.3in 0, clip]{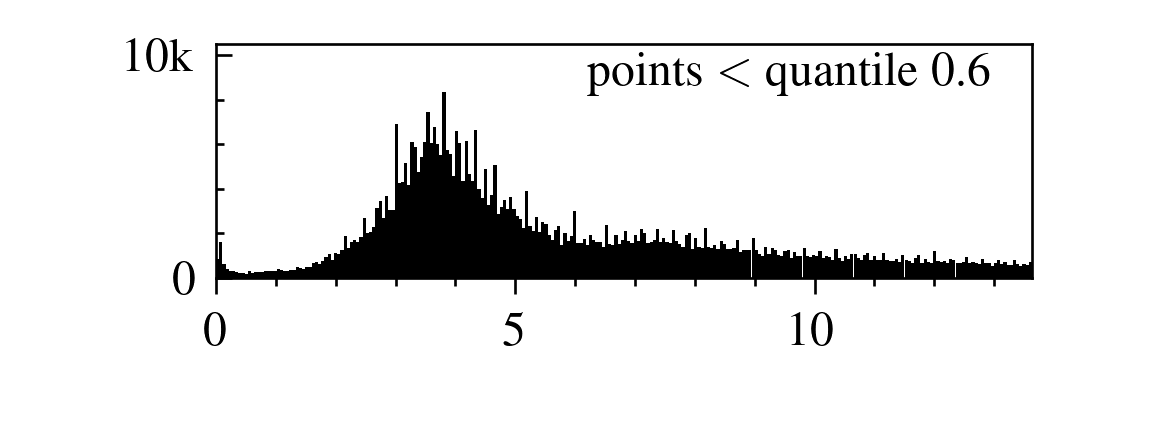}
  \caption{\textbf{Last.fm} (time between track plays)}
\end{subfigure}%

\begin{subfigure}[t]{5.5in}
  \centering
  \includegraphics[width=2.85in, trim=0      0.18in 0.3in 0, clip]{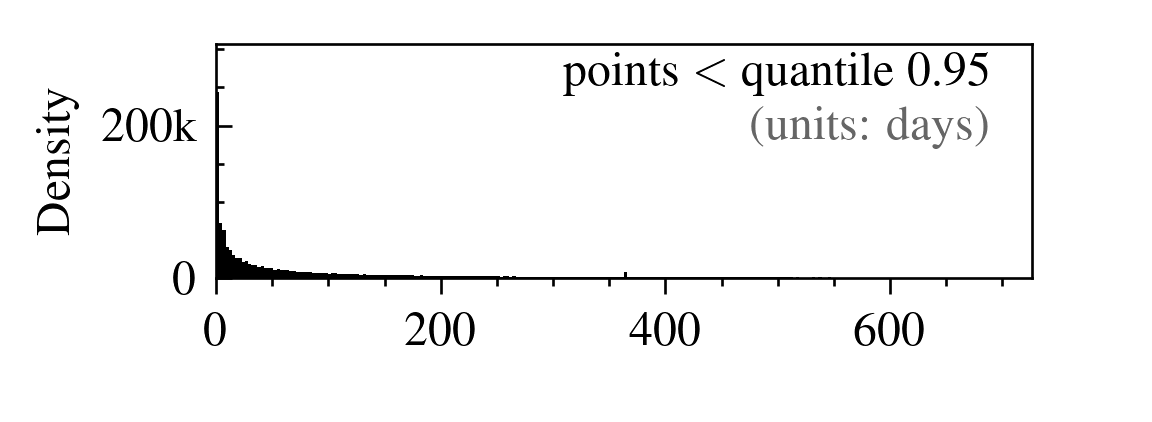}\hfill%
  \includegraphics[width=2.64in, trim=0.20in 0.18in 0.3in 0, clip]{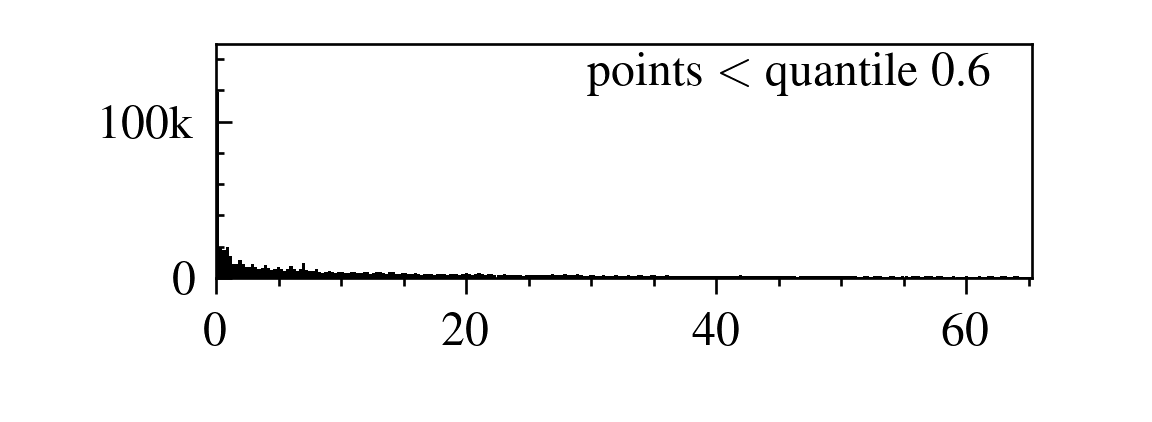}
  \caption{\textbf{Stack Overflow} (time between badge awards)}
  \label{fig:app_so_hist}
\end{subfigure}%

\begin{subfigure}[t]{5.5in}
  \centering
  \includegraphics[width=2.85in, trim=0      0.18in 0.3in 0, clip]{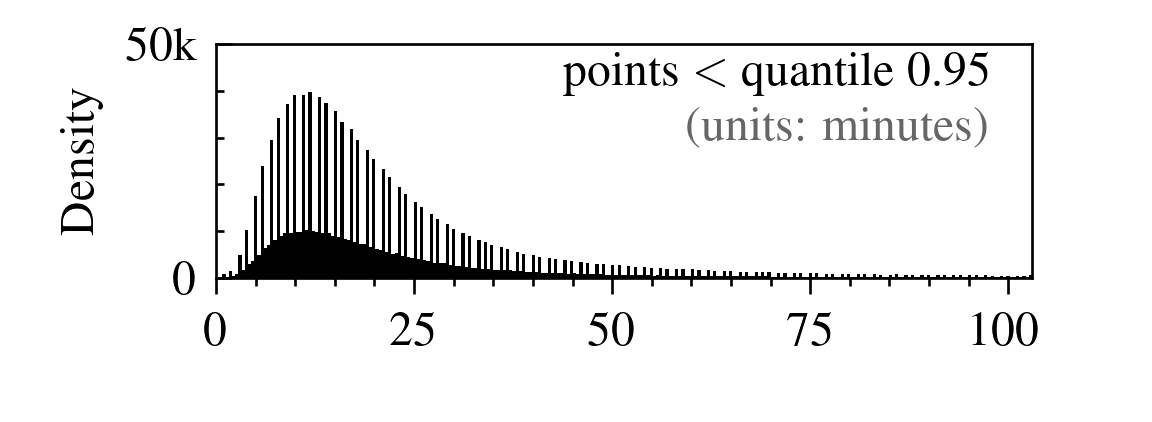}\hfill%
  \includegraphics[width=2.64in, trim=0.20in 0.18in 0.3in 0, clip]{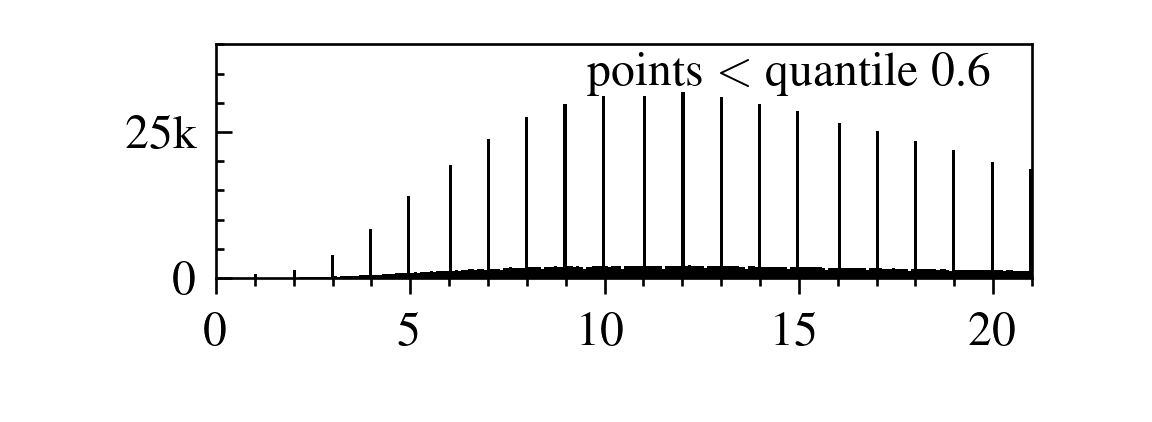}
\caption{\textbf{NYC taxi} (time between pickup times)}
  \label{fig:app_nyc_hist}
\end{subfigure}%

\caption{Histograms for 7 real-world datasets. Histograms extend until the 0.95 quantile (\textbf{left}) and 0.6 quantile (\textbf{right}). All histograms have 128 bins. The $y$-axis shows normalized counts.}
\label{fig:app_ds_histB}
\end{figure*}

The Gini coefficients for these datasets are listed in \cref{tbl:app_gini}. The Gini coefficient is a measure of distribution inequality, which is useful in this context for indicating distributions which \textit{might} have a mix of continuous and discrete parts. A distribution where probability is concentrated in a few isolated points in an otherwise large support will have a high Gini coefficient. The NYC Taxi dataset has such a distribution (see \cref{fig:app_nyc_hist}). As shown in \cref{fig:app_gini_vs_bins}, the performance improvement of adding intervals to the piecewise-constant density head scales with the Gini coefficient of the dataset.
This makes the Gini coefficient a useful indicator of the potential benefit of increasing the bin count of the piecewise-constant density head.
The Gini coefficient is calculated without binning and operates on the resolution of the data type, float32.

\subsection{Sources of the datasets} \label{app:dataset_sources}
The Last.fm, MOOC, Wikipedia and Yelp Toronto datasets used in this work were those made available alongside \citet{shchurIntensityFreeLearningTemporal2019}. The PUBG, Twitter, Yelp Airport, Yelp Mississauga and two Reddit datasets were those made available alongside \citet{ludkeAddThinDiffusion2023}. The Taobao and Amazon datasets are from \citet{xueEasyTPPOpenBenchmarking2024}. \citet{xueEasyTPPOpenBenchmarking2024} also provide a small subset of NYC Taxi pickup data and a small subset of Stack Overflow badge reward data; however, we instead reverted to the original sources for these two datasets to obtain larger datasets. For the Stack Overflow dataset, reverting to the original source was also necessary to adhere to the terms of use specified by Stack Overflow (no public re-sharing). While we do not distribute the data, it can be readily obtained from the Stack Overflow website directly. The taxi pickup data was obtained from \citet{whongNYCsTaxiTrip2014} and the Stack Overflow badge data was obtained from \citet{soDataDump2024}. From both of these sources, we draw enough sequences to reach \( 2^{21} \) inter-event times.

\subsection{Inter-event time units} \label{app:dataset_units}
Many of the datasets were published with non-standard units; for example, the Taobao dataset was published in units of 10,000 seconds (\textasciitilde 3 hours). In this work, we rescale event times of each dataset to a standard unit, such as seconds, minutes or days. The specific unit is chosen to be a unit that makes sense for each dataset (e.g. seconds or minutes are more suitable than milliseconds or days for taxi pickup intervals). Where there are multiple suitable units, we choose the unit that brings the median inter-event time closer to 10, geometrically. This latter criterion prevents the magnitude of MAE scores from differing by many orders of magnitude across datasets and makes the common magnitude easy to format. The scaling has no impact on model performance, as all models normalize their inputs and outputs. %

\subsection{Dataset splits}
For all datasets, each experiment repeat involved randomly sampling training, validation and test sets from the full dataset, in a 6:2:2 ratio. For the Taobao and Amazon datasets, which were published with predefined splits, the splits were first merged before resampling. As most of the datasets contain heavy tails, the dataset split can have a large effect on the performance. For this reason, we believe that it is important to include the variability of performance across multiple random splits, although we acknowledge that this prevents a direct comparison to previous works which report results targeting a single fixed split.

\begin{figure*}
\centering
\captionsetup[subfigure]{skip=2pt, belowskip=3pt, justification=centering, margin=0cm}
\begin{subfigure}[t]{5.5in}
  \centering
  \includegraphics[width=2.85in, trim=0      0.18in 0.3in 0, clip]{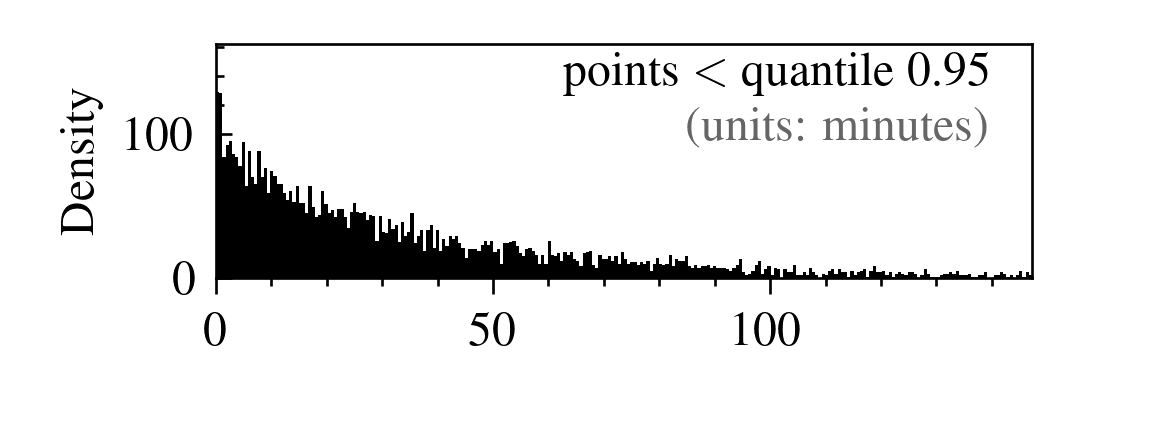}\hfill%
  \includegraphics[width=2.64in, trim=0.20in 0.18in 0.3in 0, clip]{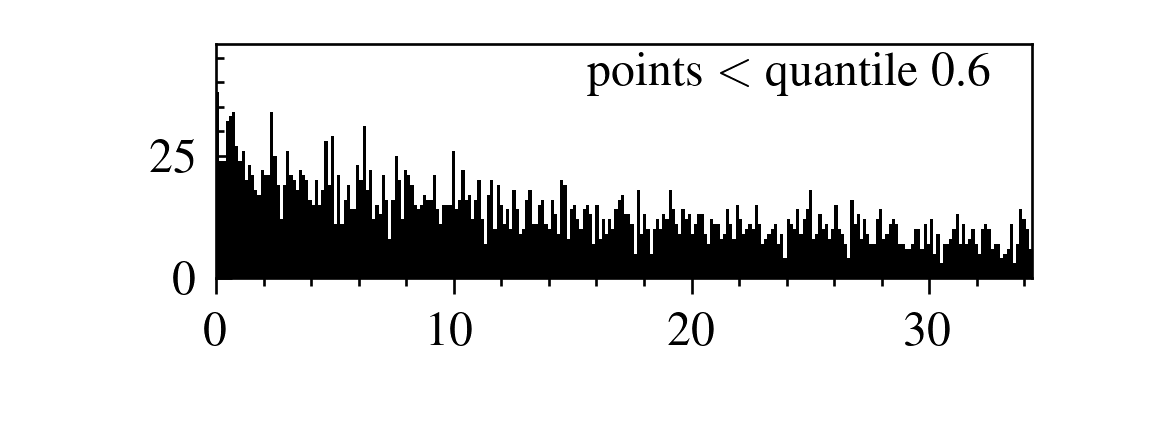}
  \caption{\textbf{Yelp airport} (time between check-ins)}
  \label{fig:app_yelp_airport_hist}
\end{subfigure}%

\begin{subfigure}[t]{5.5in}
  \centering
  \includegraphics[width=2.85in, trim=0      0.18in 0.3in 0, clip]{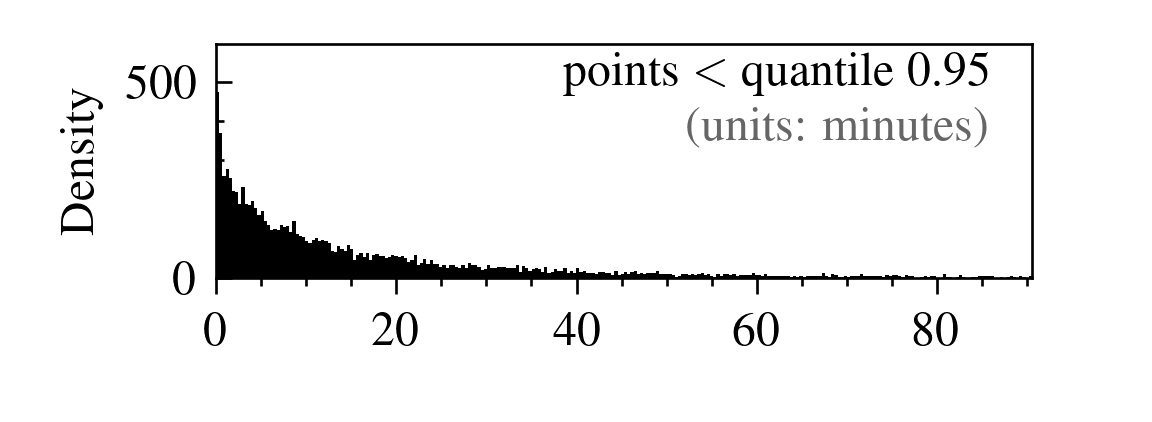}\hfill%
  \includegraphics[width=2.64in, trim=0.20in 0.18in 0.3in 0, clip]{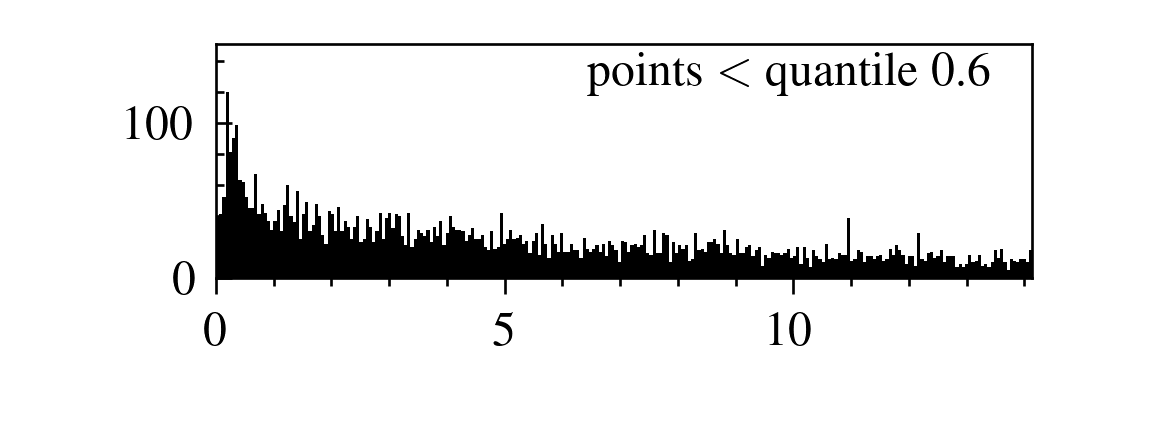}
\caption{\textbf{Yelp Mississauga} (time between check-ins)}
\end{subfigure}%

\begin{subfigure}[t]{5.5in}
  \centering
  \includegraphics[width=2.85in, trim=0      0.18in 0.3in 0, clip]{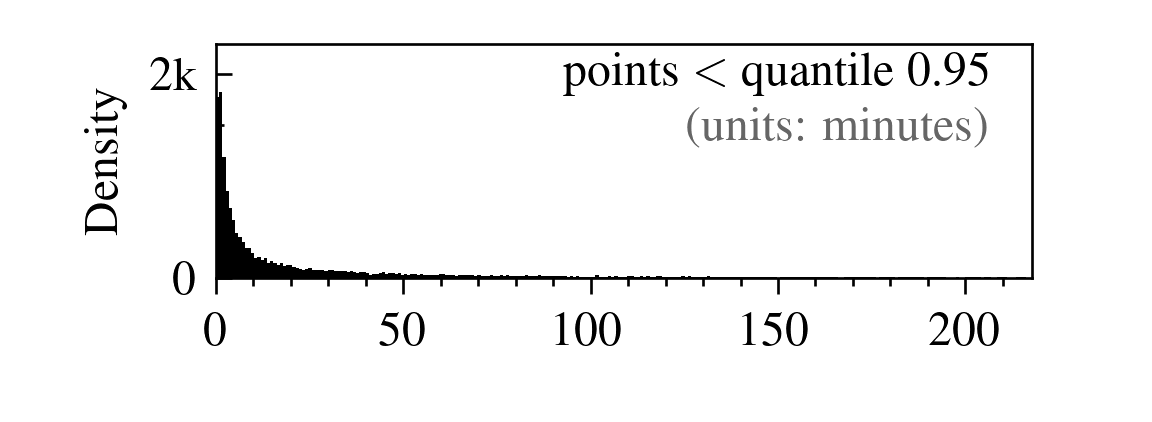}\hfill%
  \includegraphics[width=2.64in, trim=0.20in 0.18in 0.3in 0, clip]{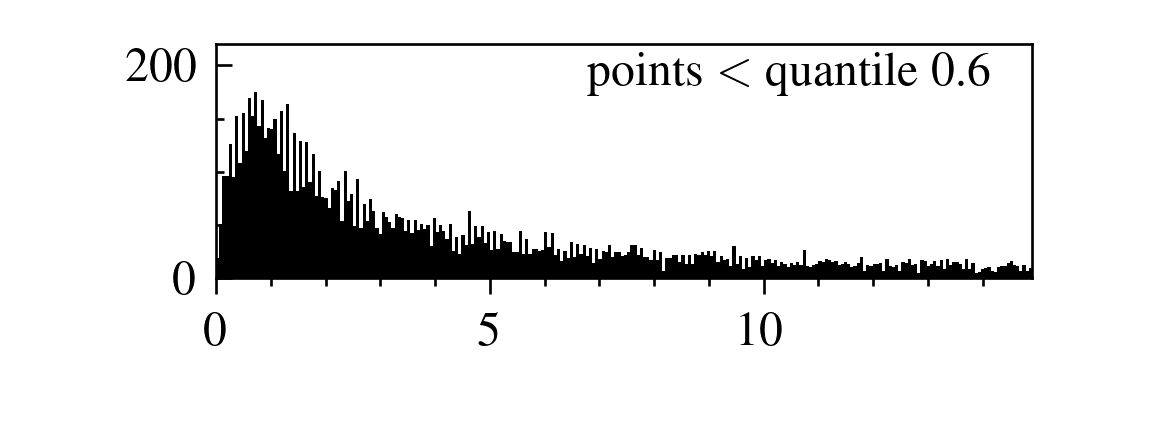}
  \caption{\textbf{Twitter} (time between tweets)}
\end{subfigure}%

\begin{subfigure}[t]{5.5in}
  \centering
  \includegraphics[width=2.85in, trim=0      0.18in 0.3in 0, clip]{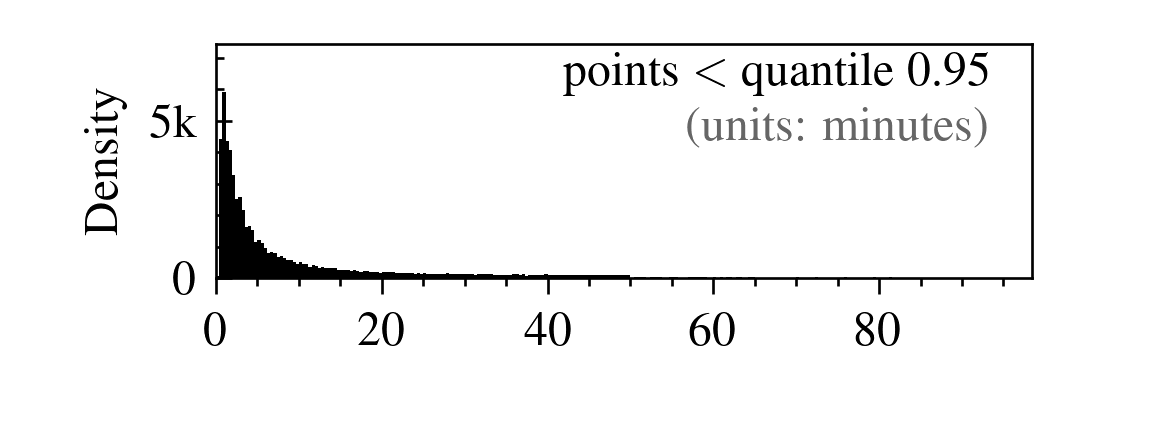}\hfill%
  \includegraphics[width=2.64in, trim=0.20in 0.18in 0.3in 0, clip]{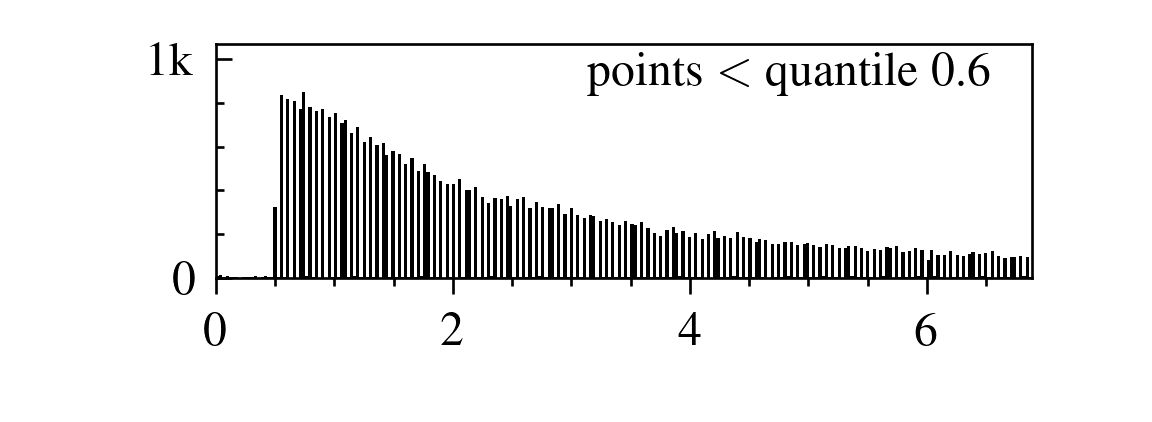}
  \caption{\textbf{Taobao} (time between user interactions)}
\end{subfigure}%

\begin{subfigure}[t]{5.5in}
  \centering
  \includegraphics[width=2.85in, trim=0      0.18in 0.3in 0, clip]{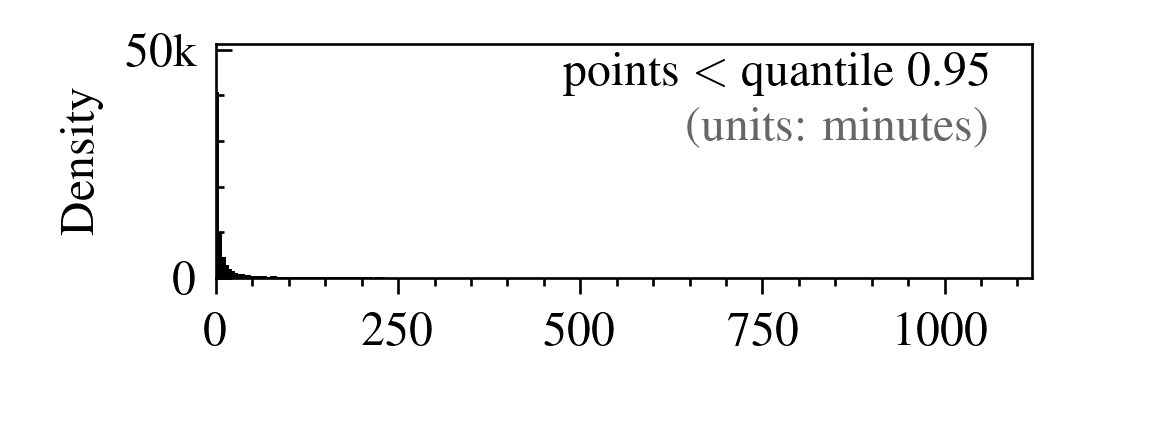}\hfill%
  \includegraphics[width=2.64in, trim=0.20in 0.18in 0.3in 0, clip]{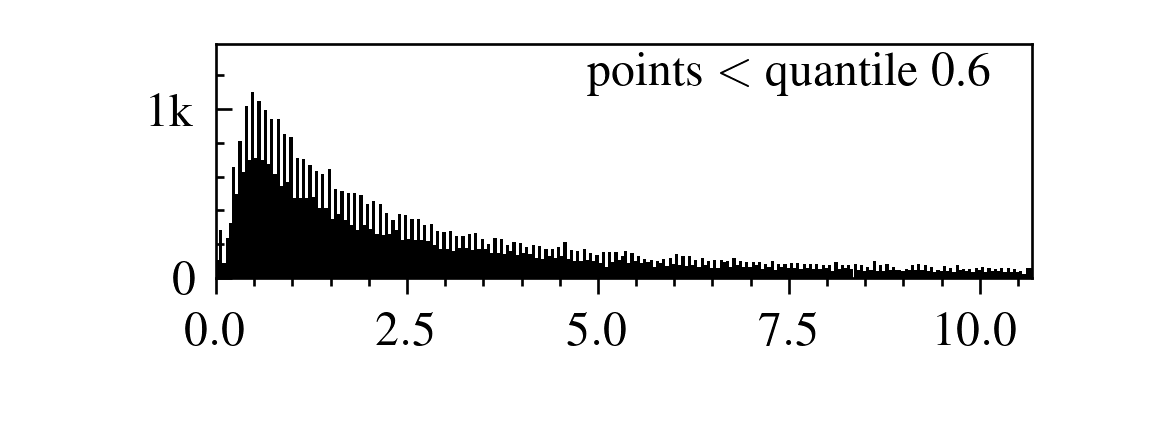}
  \caption{\textbf{Wikipedia} (time between page edits)}
\end{subfigure}%

\begin{subfigure}[t]{5.5in}
  \centering
  \includegraphics[width=2.85in, trim=0      0.18in 0.3in 0, clip]{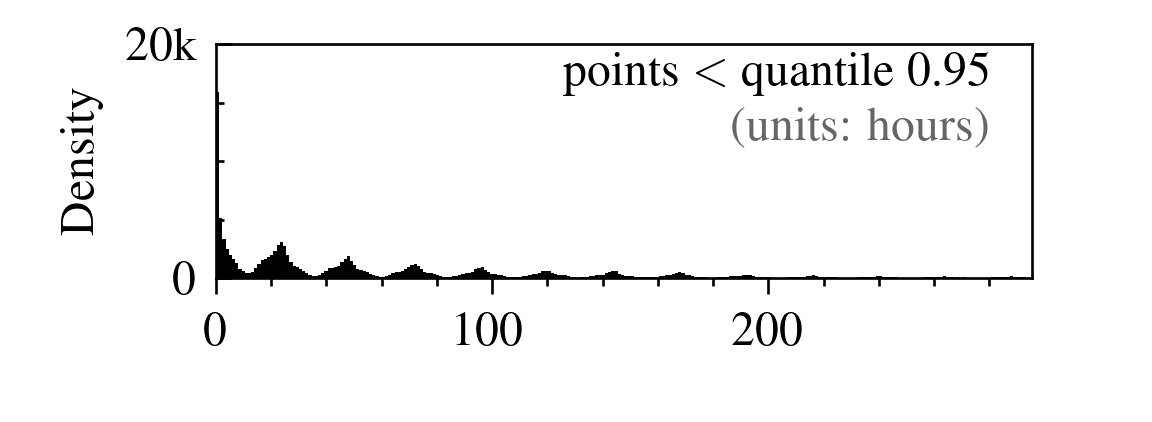}\hfill%
  \includegraphics[width=2.64in, trim=0.20in 0.18in 0.3in 0, clip]{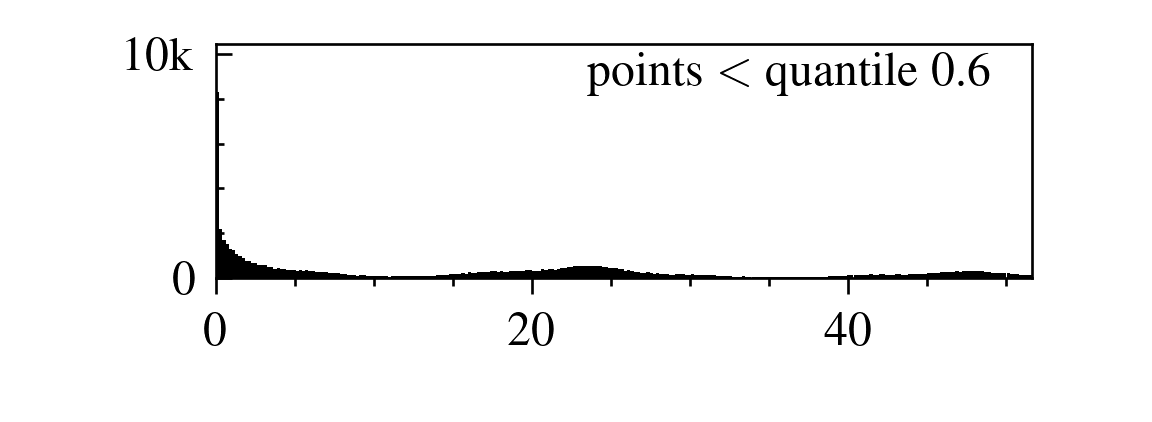}
  \caption{\textbf{Yelp Toronto} (time between check-ins)}
\end{subfigure}%

\begin{subfigure}[t]{5.5in}
  \centering
  \includegraphics[width=2.85in, trim=0      0.18in 0.3in 0, clip]{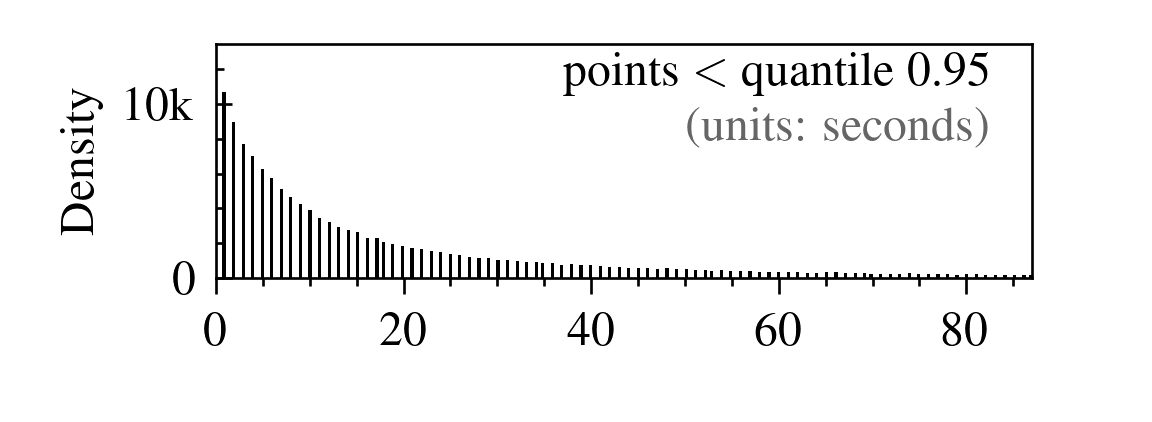}\hfill%
  \includegraphics[width=2.64in, trim=0.20in 0.18in 0.3in 0, clip]{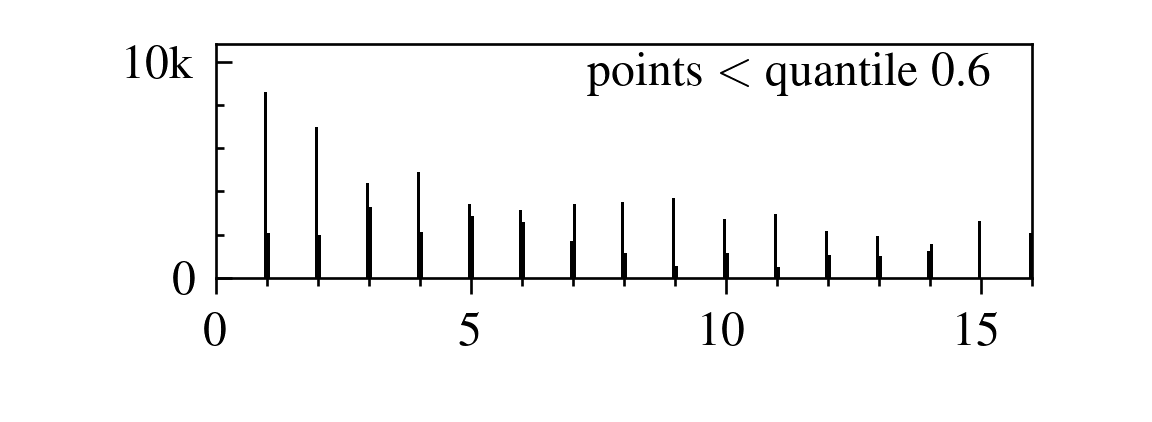}
  \caption{\textbf{PUBG} (time between player deaths)}
  \label{fig:app_pubg_hist}
\end{subfigure}%
\caption{Histograms for 7 real-world datasets. Histograms extend until the 0.95 quantile (\textbf{left}) and 0.6 quantile (\textbf{right}). All histograms have 128 bins. The $y$-axis shows normalized counts.}
\label{fig:app_ds_histA}
\end{figure*}

\begin{figure*}
  \centering
\includegraphics[width=0.9\linewidth]{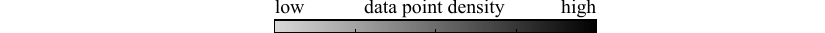}%
\par\vspace{0.10in}%
\captionsetup[subfigure]{skip=6pt, belowskip=4pt, justification=centering, margin=0cm}
\centering
\begin{subfigure}[t]{2.881in}
  \includegraphics[width=\linewidth, trim=20pt 0 0 0, clip]{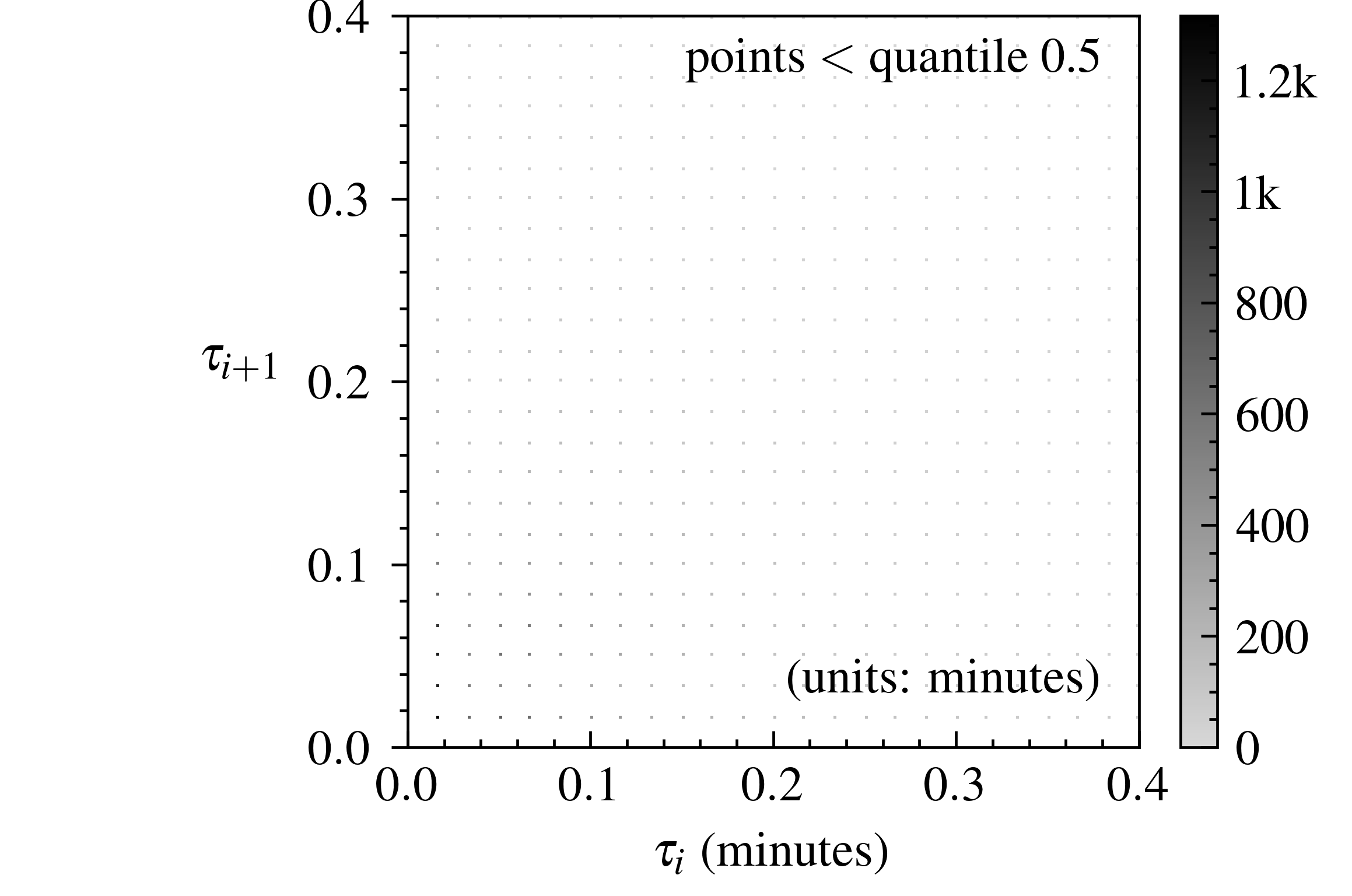}
  \caption{\textbf{MOOC}}
\end{subfigure}\hfill%
\begin{subfigure}[t]{2.619in}
  \includegraphics[width=\linewidth, trim=40pt 0 0 0, clip]{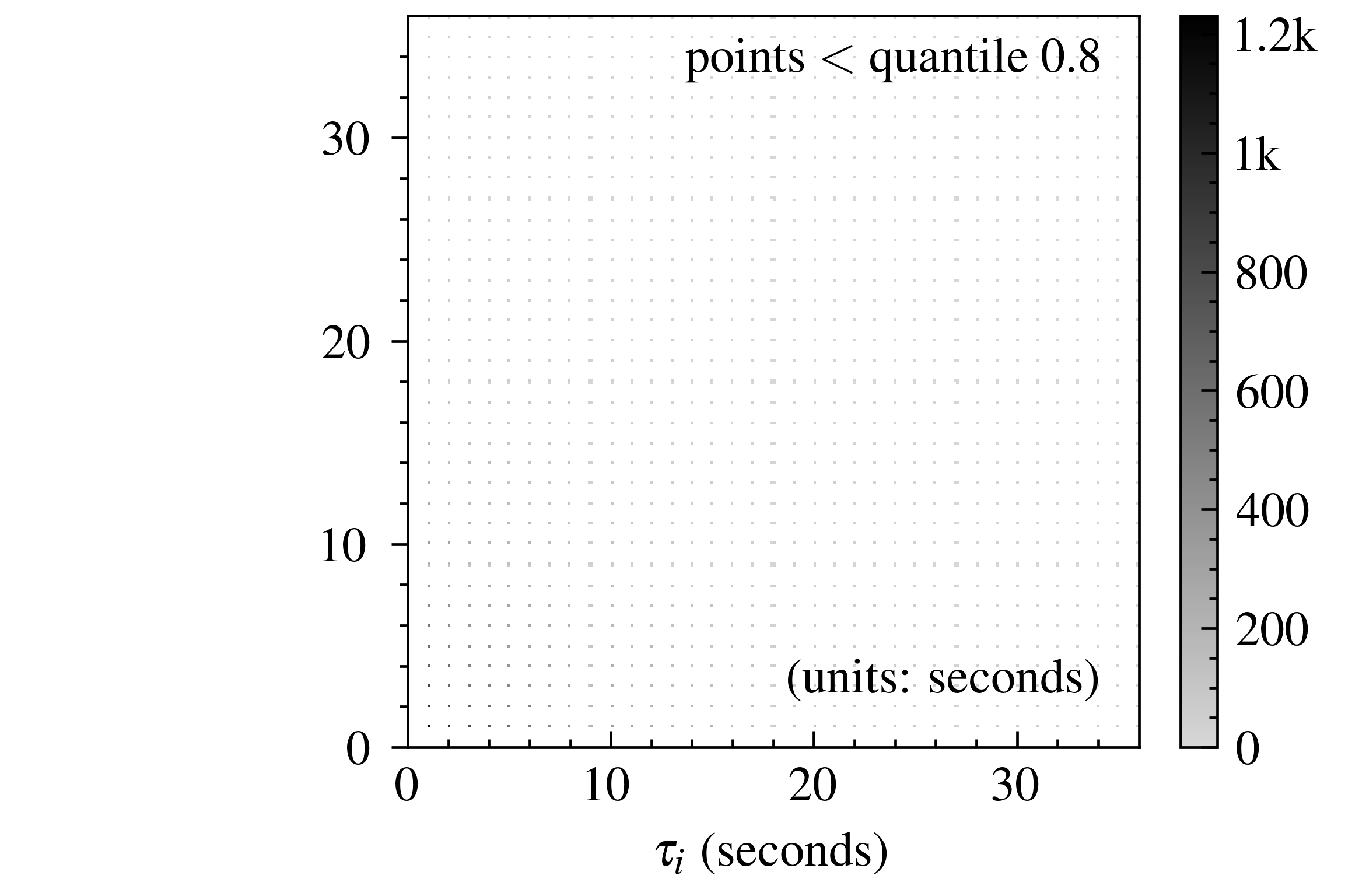}
  \caption{\textbf{PUBG}}
\end{subfigure}%

\begin{subfigure}[t]{2.881in}
    \includegraphics[width=\linewidth, trim=20pt 0 0 0, clip]{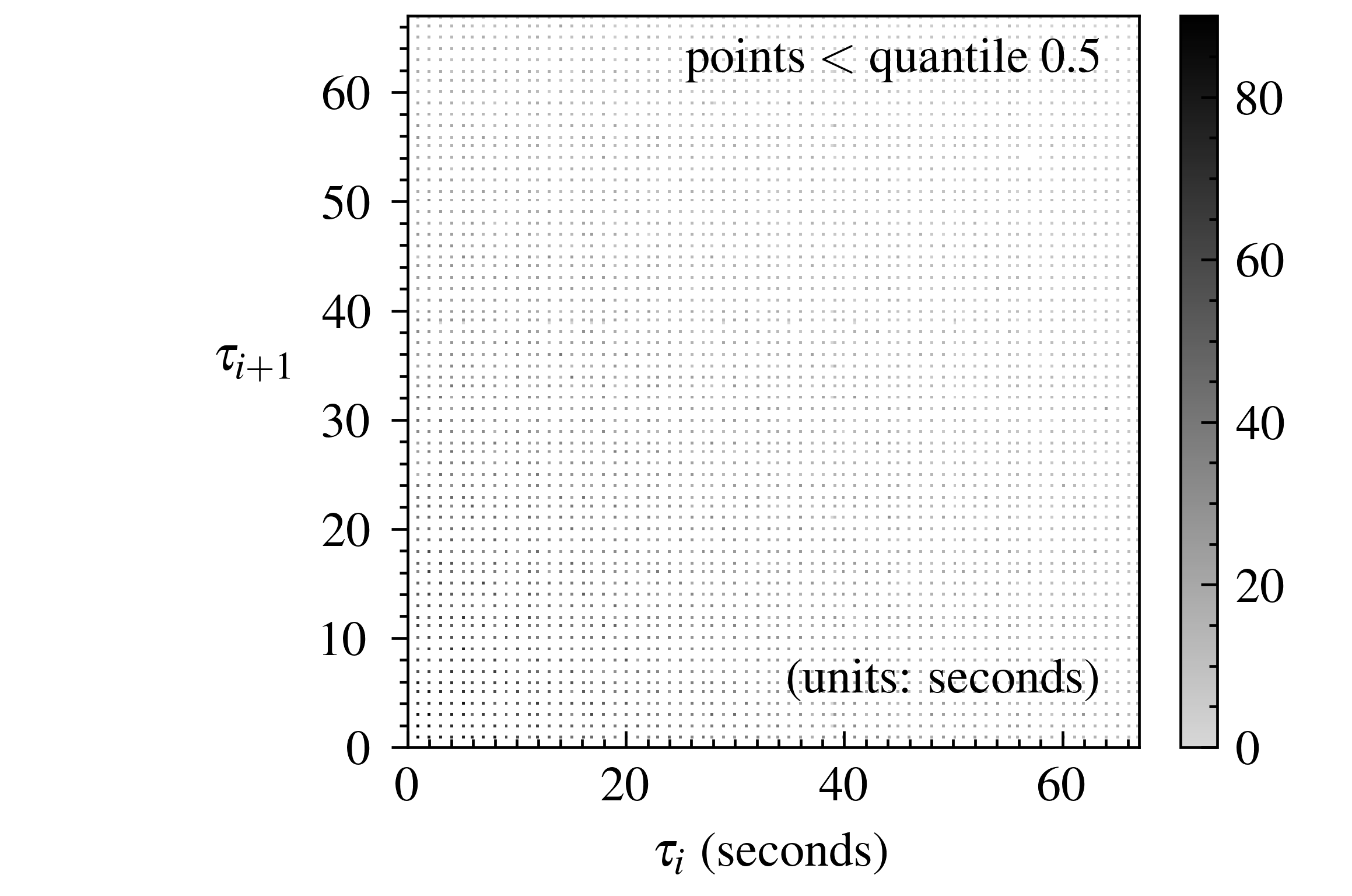}
    \caption{\textbf{Reddit AskScience}}
  \end{subfigure}\hfill%
\begin{subfigure}[t]{2.619in}
  \includegraphics[width=\linewidth, trim=40pt 0 0 0, clip]{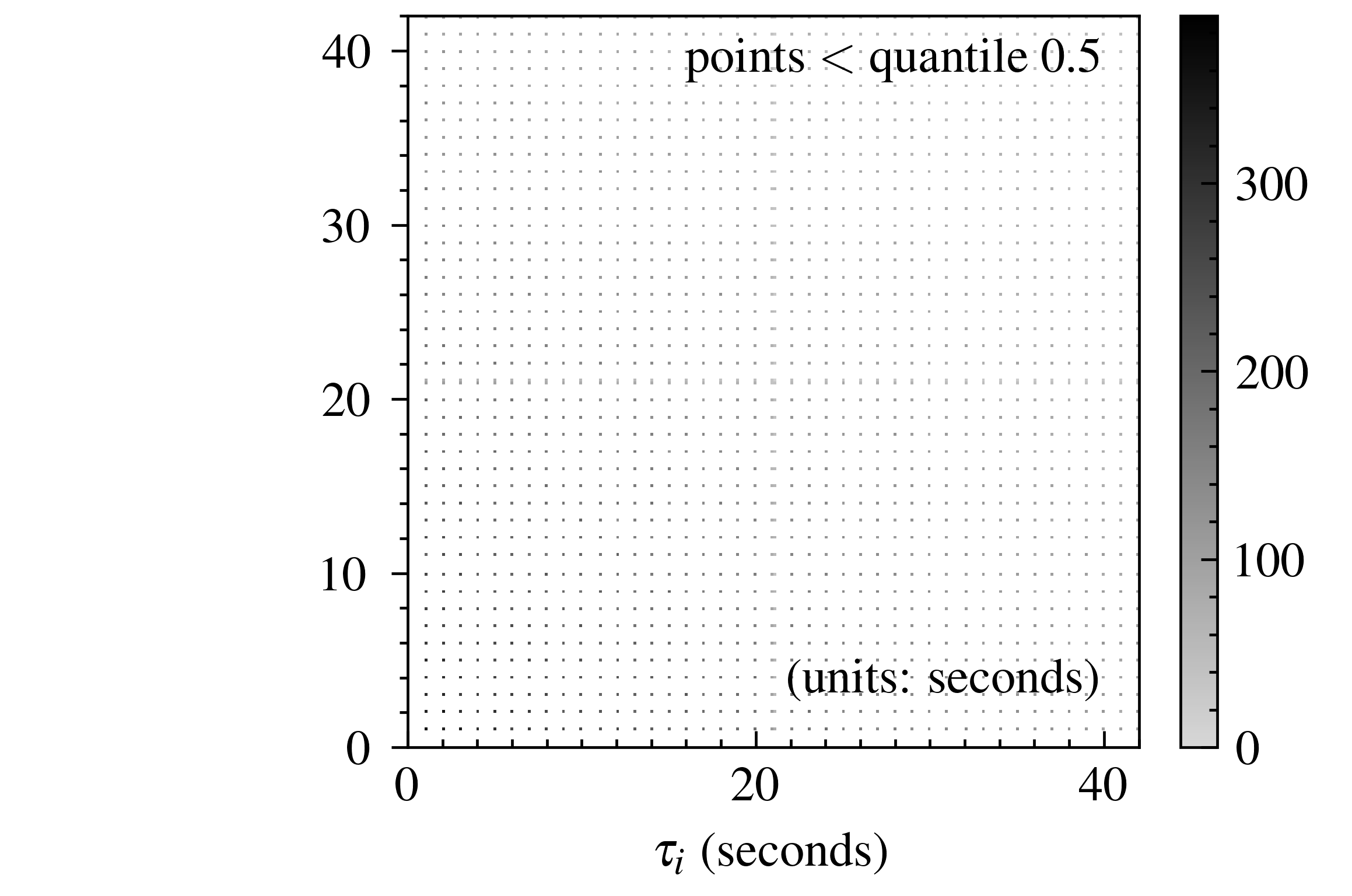}
  \caption{\textbf{Reddit Politics}}
\end{subfigure}%

\begin{subfigure}[t]{2.881in}
  \includegraphics[width=\linewidth, trim=20pt 0 0 0, clip]{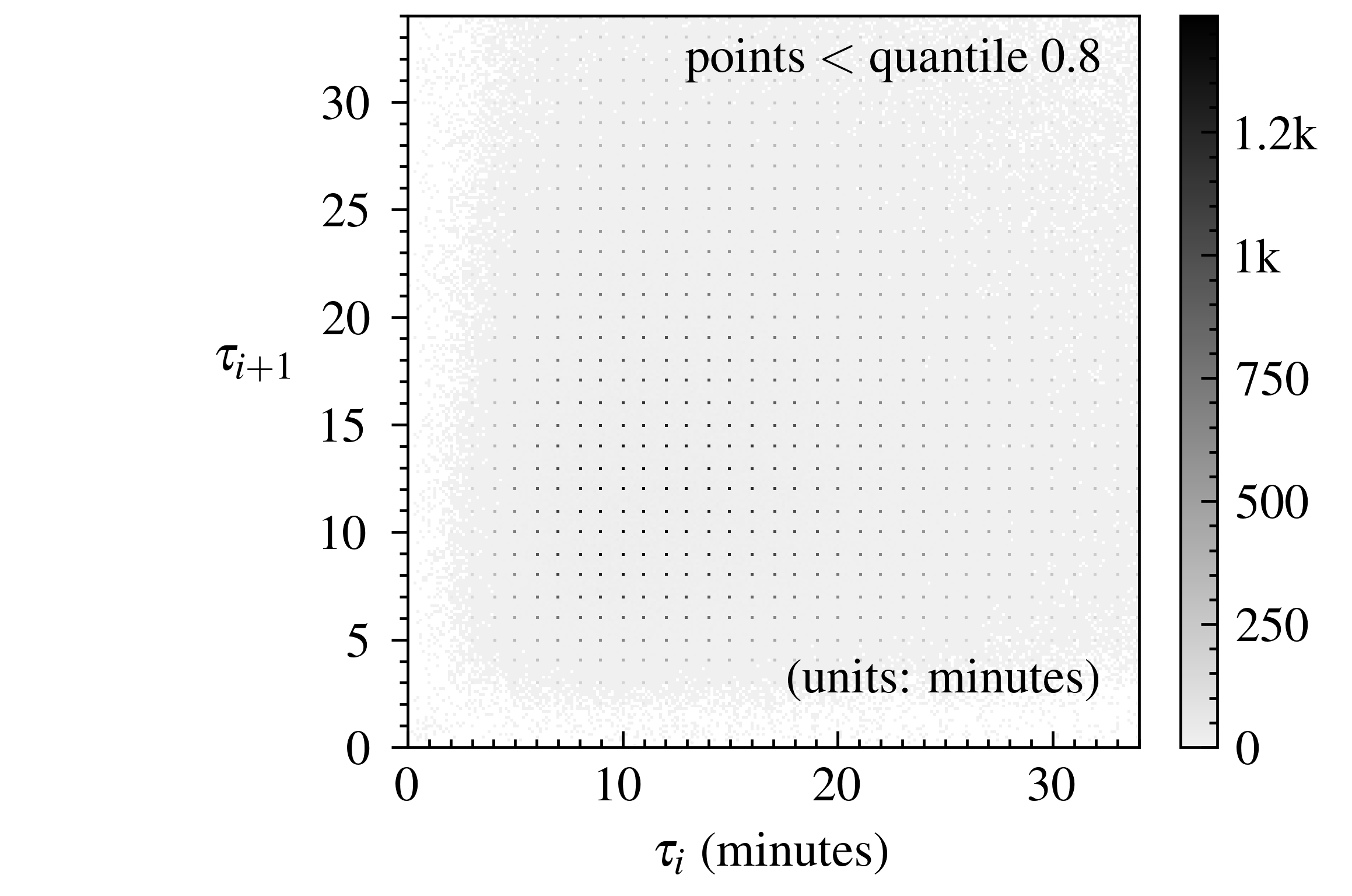}
  \caption{\textbf{NYC Taxi}} 
\end{subfigure}\hfill%
\begin{subfigure}[t]{2.619in}
  \includegraphics[width=\linewidth, trim=40pt 0 0 0, clip]{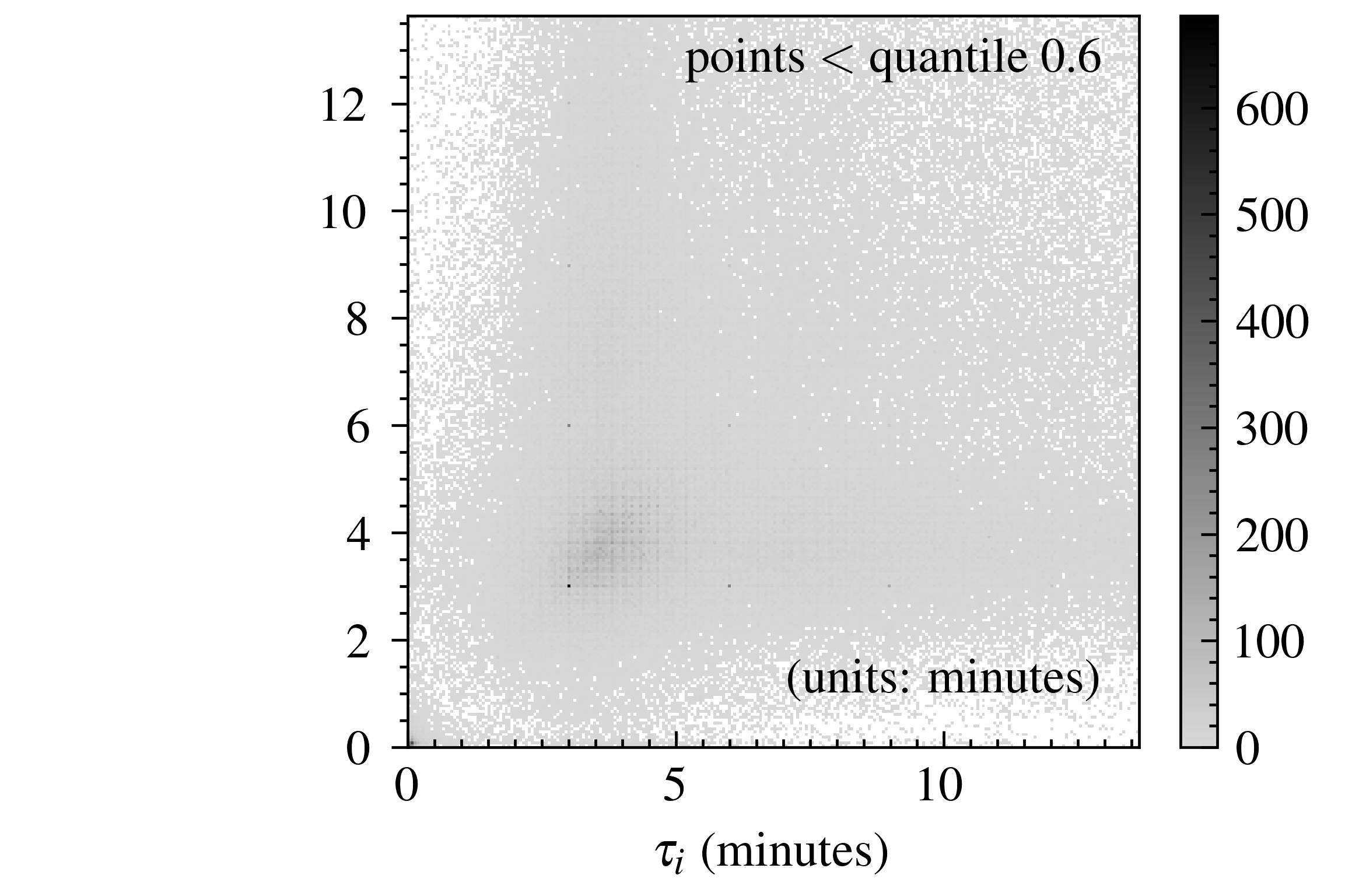}
  \caption{\textbf{Last.fm}} 
\end{subfigure}%
\caption{
Distribution of pairs of inter-event times, \( \tau_i \) and \( \tau_{i+1} \) for 6
real-world datasets. The first 4 datasets (\textbf{MOOC}–\textbf{Reddit Politics}) appear to have a grid structure on account of the resolution at which the data was recorded. The last 2 datasets (\textbf{NYC Taxi} and \textbf{Last.fm}) have a grid structure embedded within a smoother distribution. Each figure has a quantile threshold chosen so that the structure is visible. 
Each figure is a rasterization of event counts into a \( 256 \times 256 \) pixel grid.
}\label{fig:app_ds_hist2d_discrete}
\end{figure*}

\FloatBarrier

\clearpage

\section{Model details} \label{app:models}
This section describes the implementation of the \model{rnn}, \model{gpt-a} and \model{gpt-b} stems, and the implementation of the existing heads and models used in this work. 

\subsection{\model{rnn}, \model{gpt-a} and \model{gpt-b} stems} \label{app:gpt_arch}
The \model{rnn} stem is a single-layer gated recurrent unit (GRU) \citep{choLearningPhraseRepresentations2014} with 64 hidden units. The input is a 32-dimensional vector containing normalized inter-event times.

The \model{gpt-a} and \model{gpt-b} transformers follow the standard GPT-2 architecture \citep{radfordLanguageModelsAre2019} with the layer configurations (layers-heads-embedding size) set to 2-4-16 and 6-4-32 respectively. This gives \model{gpt-a} and \model{gpt-b} 108k and 1.20M trainable parameters, respectively. The standard GPT-2 small configuration defined by \citet{radfordLanguageModelsAre2019} uses a 12-12-768 configuration, highlighting that both \model{gpt-a} and \model{gpt-b} are very small in the context of GPTs used as large language models. Computational limits prevented the investigation of larger models. 

\subsection{Input encoding for \model{gpt} based models} \label{app:valueenc}
For the \model{gpt-a} and \model{gpt-b} stem architectures, the scalar values representing inter-event times are converted to a vector representation according to \cref{alg:encodeValue} before being input into the transformers. This approach is similar to that used by \citet{zuoTransformerHawkesProcess2020}. A point of difference is that we do not hard-code the sinusoidal frequencies but choose appropriate scales for each dataset so that both long and short intervals of each dataset are captured.

\begin{algorithm*}
\caption{Encode a scalar value \( v \) into a vector of even size \( n \). \(v_{\text{epsilon}}\) should be chosen near the smallest inter-event time expected to be encountered and \( v_{\text{max}} \) should be chosen to be near the largest time expected to be spanned by inputs to the model, and will depend on the length of a model's input. In this work, we choose \( v_{\text{epsilon}} \) as the smallest inter-event time observed in the training set, and \( v_{\text{max}} \) as the largest time between events \( t_i \) and \( t_{i+L} \) where \( L \) is the input length of the model.}
\label{alg:encodeValue}
\begin{algorithmic}[1]
  \STATE \textbf{Function} encodeValue$(v, v_{\text{epsilon}}, v_{\text{max}}, n)$
  \STATE \quad $\text{scale}_{\text{min}} \gets 2 \pi / v_{\text{max}}$ \quad \COMMENT{Or use $\frac{3}{2} \pi$ to add a little more range}
  \STATE \quad $\text{scale}_{\text{max}} \gets 2 \pi / v_{\text{epsilon}}$
  \STATE \quad $\text{scale} \gets \operatorname{linspace}(\log(\text{scale}_{\text{min}}), \log(\text{scale}_{\text{max}}), n/2)$ \quad \COMMENT{A vector of length \( n/2 \)}
  \STATE \quad $x_1 \gets \cos(\exp(\text{scale}) \cdot v)$
  \STATE \quad $x_2 \gets \sin(\exp(\text{scale}) \cdot v)$
  \STATE \quad $x  \gets \text{concat}(x_1, x_2)$
  \STATE \quad \textbf{return} $x$
\end{algorithmic}
\end{algorithm*}

\subsection{Implementation of existing models} \label{app:othermodels}
The \model{const}, \model{exp} and \model{nn} heads are implemented following the code accompanying \citep{omiFullyNeuralNetwork2019}. The \model{logmix} head is implemented following the code accompanying \citet{shchurIntensityFreeLearningTemporal2019}. The \model{thp} model is implemented partially following \citep{zuoTransformerHawkesProcess2020}, but deferring to the implementation accompanying \citep{xueEasyTPPOpenBenchmarking2024} for a number of bug fixes to the original model. The \model{dtpp} model is implemented following the code accompanying \citet{panosDecomposableTransformerPoint2024}.

One deviation from these implementations is in model initialization. In order to improve convergence speed and stability while training, all models were initialized so that their outputs have first and second moments similar to those of the training set's empirical distribution. 

\clearpage

\section{Limitations of existing output representations} \label{app:model_limitations}
The existing output structures considered in this work have limitations in terms of distribution flexibility and training stability, both of which contribute to preventing them from representing multimodal distributions with very narrow peaks.

The constant, exponential and softplus intensity heads have limited distribution flexibility. The \model{logmix} head has much greater capacity to represent complex distributions; however, it experiences training instability in the presence of narrow peaks in the target distribution. The \model{nn} head can also represent complex distributions; however, 
it appears to have limited ability to output different complex distributions for different inputs.

\subsection{Distribution flexibility} \label{app:distribution_flexibility}
The constant intensity, exponential intensity and the softplus intensity correspond to distributions over \( [0, \infty) \) with 1 or 2 parameters. With only 1 or 2 parameters, they do not have the flexibility to represent distributions where there are multiple separated peaks of high probability density, which is a property of several of the real-world datasets (see \cref{fig:app_ds_hist2d_discrete}).

\textbf{The constant intensity} output corresponds to the exponential distribution. The exponential distribution has a single scalar parameter with the distribution's mode fixed at 0. The distribution cannot concentrate probability at any non-zero point.

\textbf{The exponential intensity} output corresponds to the Gompertz distribution. The Gompertz distribution has two parameters, a scale and a shape parameter. It is a unimodal distribution so cannot concentrate probability at two or more separated points.

\textbf{The softplus intensity} function corresponds to a unimodal probability distribution with two shape parameters. Its unimodal shape is inferred from the monotonicity of the softplus function: a monotonic intensity implies a convex cumulative intensity, and both then imply a unimodal density function. Being unimodal, it is restricted in its ability to concentrate probability at multiple separated points. The distribution lacks a familiar name and does not have an elementary closed form (it can be expressed in terms of a polylogarithm). Rather than being motivated by a particular distributional family, the softplus function is used as it is strictly positive and smooth \citep{meiNeuralHawkesProcess2017, zuoTransformerHawkesProcess2020}.

For these three heads, a theoretical argument is sufficient to explain how they cannot represent distributions with many peaks. For the \model{nn} and \model{logmix} heads, which have much greater capacity to represent complex distributions, we present hypotheses for limitations along with supporting empirical evidence.

\subsection{Mixture of lognormals instability} \label{app:logmix_instability}
A mixture of lognormals can theoretically represent distributions with multiple sharp peaks. However, when training models with this output structure on datasets where there is a significant discrete component to the inter-event distributions, we do not observe the models concentrating probability in narrow peaks to the same extent as can be seen with the piecewise-constant model; see the densities in \cref{fig:head_densities}, for example. We argue that such models struggle to represent such distributions on account of training instability in the presence of singularities.  
A mixture of lognormals on \( t \in (0, \infty) \) is equivalent to a mixture of Gaussians on \(y \in (-\infty, \infty) \), where \( y = \log(t) \), and so models that output a mixture of lognormals can be thought of as outputting a mixture of Gaussians on a log scale. A mixture of Gaussians can produce any number of modes, up to the number of components in the mixture. 

The process of fitting these mixtures using the expectation-maximization (EM) algorithm demonstrates how this flexibility does not necessarily allow the mixture to represent distributions with sharp peaks. When fitting a mixture of Gaussians using the EM algorithm, a component's variance can collapse toward zero at a single point. To avoid this, strategies such as introducing priors on the parameters or resetting the EM optimization procedure are used \citep{bishopPRML2006}. 

\begin{figure*}[tb]
  \begin{subfigure}{2.625in}
\includegraphics[width=2.625in]{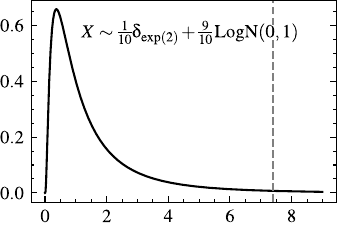}
\caption{The probability distribution of a dataset generated from a mixture of a point mass at \( t = \exp(2) \sim 7.4 \) and a lognormal distribution with parameters \( \mu = 0 \) and \( \sigma = 1.0 \).}
\label{fig:logmix_stability_data}
\end{subfigure}
\hfill
\begin{subfigure}{2.625in}
\includegraphics[width=2.625in]{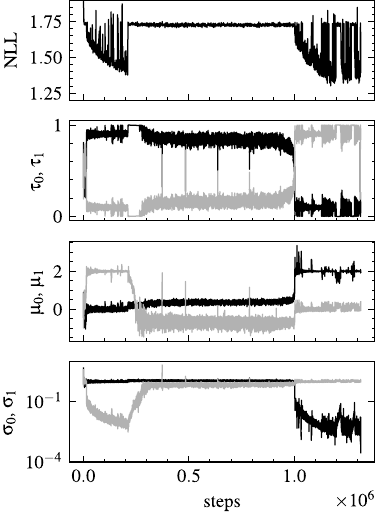}
\caption{How the NLL (evaluated on a held-out validation set) and the model's six parameters change while training.}
\end{subfigure}
\caption{The instability of a logmix output head when a dataset has a probability mass at a certain point. \textbf{(a):} shows the distribution of inter-event times. The dataset is formed from a mixture where 10\% of values come from a point source at \( t = \exp(2) \), and the remaining values are drawn from a lognormal distribution with parameters \( \mu = 0 \) and \( \sigma = 1.0 \). All events are independent. The model used is a \model{gpt-a} stem with a 2-component lognormal mixture output. The model is trained using stochastic gradient descent with momentum (0.9) and a learning rate of \(5\times 10^{-4} \). \textbf{(b):} 
shows the mixture parameters (the model output) and the NLL (calculated on an evaluation set) over the course of training. The six model parameters are the mixing parameters (\(\tau_0, \tau_1\)) and the means and variances ( \( \mu_0, \mu_1, \sigma_0, \sigma_1 \)) of the two underlying Gaussian distributions. 
Notably, neither \( \mu_0 \) nor \( \mu_1 \) approach 2.0 in a stable manner. Training stopped when evaluating the distribution median produced \texttt{NaN} values. This instability occurs reliably across runs.}
\label{fig:logmix_stability}
\end{figure*}

Although our models are trained with stochastic gradient descent rather than EM, the singularity in the log-likelihood function as \( \sigma \to 0 \) is a property of the objective function, not the optimizer, and we hypothesize that stochastic gradient descent encounters it. If this is the case, we would expect to see training instability in the form of sudden loss spikes and parameter perturbations near when a component's variance parameter approaches zero.
A simple case study (\cref{fig:logmix_stability}) shows \model{gpt-a}, outputting a 2-component lognormal mixture, trained on a dataset generated from a mixture of a lognormal distribution and a point mass. As one of the output components narrows around the single point, sudden spikes in gradient cause the parameters of the component to be perturbed and its mixing weight to suddenly decrease. "Real-world" datasets such as the NYC taxi dataset have such point masses, and training stability in the presence of these sharp distributions is a candidate explanation for why the \model{logmix} head is not observed to concentrate probability at such points in a stable manner. To prevent the instability, regularization could be introduced in the form of a prior on the variance of each component. Doing so amounts to specifying a resolution of interest for the task and would be evidence of the benefit of acknowledging the discrete nature of recorded event times.

\clearpage

\subsection{The \model{nn} head can easily translate but not modulate peaks} \label{app:nn_inflexibility}

The \model{nn} head from \citet{omiFullyNeuralNetwork2019} is, in principle, a flexible output structure. However, the results in the main body suggest there is some limitation that prevents it from achieving NLL scores similar to those of the \model{logmix} or \model{pc} heads. We find, empirically, that the \model{nn} head can successfully model sharp peaks for a stationary distribution, but struggles to switch these peaks on and off for different inputs. Inspecting the implementation of the \model{nn} head suggests a mechanism for this phenomenon. Next, we outline the implementation of the \model{nn} head. Then we start by presenting the theoretical argument along with empirical evidence. 

\subsubsection{\model{nn} head implementation}
The \model{nn} head has two inputs: \(\tau\), the inter-event time being evaluated, and \( \textbf{h} \), the hidden state vector emitted by the RNN stem. The output of the \model{nn} head is a scalar value representing the cumulative hazard function, \( \Lambda(\tau) \) evaluated at the input \( \tau \) and conditional on the history representation, \( \textbf{h} \). The hazard function, \( \lambda(\tau) \), is calculated from automatic differentiation of \( \Lambda(\tau) \) with respect to the input \( \tau \). As \( \Lambda \) must be a non-decreasing function of \( \tau \) in order to be a valid cumulative hazard function, the multiplicative weights in the \( \tau \) pathway of the \model{nn} head are constrained to be non-negative. 

The relationship between input and output can be described as follows. Let \( W_{\tau} \in \mathbb{R}^{d \times 1}\), \( W_h \in \mathbb{R}^{d \times d_{\text{RNN}}}\) and \( \mathbf{b_h} \in \mathbb{R}^d \) be the first weights in the network that map the input pair \( (\tau, \mathbf{h}) \) to the first embedding vector \( \mathbf{u} \in \mathbb{R}^d \). It will be useful, as shorthand, to define the function \( \phi(\mathbf{h}) = W_h \mathbf{h} + \mathbf{b_h} \). The remaining layers, of which there are 3, are linear layers with two tanh activations and a final softplus activation. The input-output relationship can be written as:

\begin{align}
  \phi(\mathbf{h}) &= W_h \mathbf{h} + \mathbf{b_h} \label{eq:phi_def} \\
  \mathbf{u} &\gets \tanh(W_{\tau} \tau + \phi(\mathbf{h})) \nonumber \\
  u_{\text{final}} &\gets W_{\text{out}} \tanh(W_2 \tanh(W_1 \mathbf{u} + \mathbf{b_1}) + \mathbf{b_2}) + b_{\text{out}} \nonumber \\
  \Lambda &\gets \text{softplus}(u_{\text{final}}) \nonumber
\end{align}

Where \( W_{\tau}, W_1, W_2, \) and \( W_{\text{out}} \) matrices have non-negative elements.

\begin{figure*}[t]
\centering
\begin{subfigure}[t]{1.80in}
    \centering
  \includegraphics[width=\linewidth]{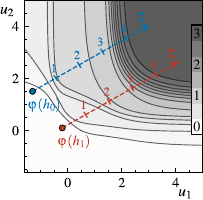}
  \caption{The cumulative intensity, \( \Lambda \), evaluated for a square of \(u \)-space. Values are rounded
    to the nearest integer for visualization purposes.}
  \label{fig:app_nn_demo_f1}
\end{subfigure}\hfill
\begin{subfigure}[t]{1.68in}
    \centering
  \includegraphics[width=\linewidth]{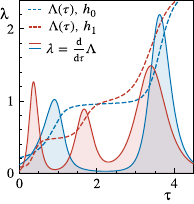}
  \caption{\( \Lambda \) and its derivative along the two \(\tau\)-paths from \textbf{(a)}, which are \(W_{\tau} \tau + \phi(h) \) for \( h_1 \) and \( h_2 \).}
  \label{fig:app_nn_demo_f2}
\end{subfigure}\hfill
\begin{subfigure}[t]{1.55in}
    \centering
  \includegraphics[width=\linewidth]{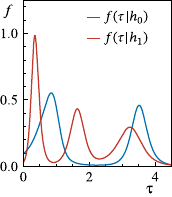}
  \caption{The probability densities corresponding to the two paths in \textbf{(a)}.}
  \label{fig:app_nn_demo_f3}
\end{subfigure}%
  \caption{
    An \code{nn} head where the first embedding vector \( \mathbf{u} \) is 2-dimensional. The matrix projecting \(\tau\) into \(\textbf{u}\)-space, \( W_{\tau} \in \mathbb{R}^{2\times 1}\), is the unit vector \( [0.86, 0.51]^{T} \).
    The second and third layers have 5 and 3 units respectively (\( W_1 \in \mathbb{R}^{5\times2}, W_2 \in \mathbb{R}^{5\times 3}, W_{\text{out}} \in \mathbb{R}^{3\times 1} \) ). 
  \textbf{(b)} shows the altitude change along the \(\tau\)-paths from \textbf{(a)}; this is the cumulative intensity \( \Lambda \). The derivative, \( \lambda \), is a solid curve. The corresponding probability densities are shown in \textbf{(c)}. All possible probability densities for this head correspond to a different starting position in the \(u\)-space.}
\label{fig:app_nn_demo}
\end{figure*}

\subsubsection{Parallel paths in a fixed, climbing, terraced landscape}
The \model{nn} head can be described as being only able to map time points to a line through a fixed landscape. A consequence of the \model{nn} head's structure is that for a given history representation, \( \mathbf{h} \), the inter-event time, \( \tau \), defines a straight path in the \( \mathbf{u} \)-space whose image is non-decreasing. \cref{fig:app_nn_demo} illustrates a toy \model{nn} head where the \(\textbf{u}\)-space is 2-dimensional; the following layers have 5 and 3 units respectively. The first sub-figure, \cref{fig:app_nn_demo_f1}, is a contour plot showing the image through the network of a square of \( \textbf{u} \)-space. The parameters that map the pair \( (\tau, \mathbf{h} ) \) to \(\textbf{u}\)-space are fixed once learned (\( W_{\tau}, W_{h} \) and \( \textbf{b}_h \)), and the only way to control the path taken through \( \textbf{u} \)-space is through the vector \( \textbf{h} \) emitted from the model stem. As \( W_{\tau} \) is fixed, all possible paths are parallel lines. The middle figure, \cref{fig:app_nn_demo_f2}, plots the altitudes of the landscape in \cref{fig:app_nn_demo_f1}, which is the cumulative hazard \( \Lambda(\tau) \), along two example paths, starting at \( \phi(\mathbf{h_1}) \) and \( \phi(\mathbf{h_2}) \). The resulting probability density functions are shown in \cref{fig:app_nn_demo_f3}. The real \model{nn} head has a 64-dimensional \( \mathbf{u} \)-space, rather than a 2-dimensional one; however, the same limitation that, once trained, all possible probability distributions correspond to parallel paths in \( \mathbf{u} \)-space, still applies. It is clear how moving \( \phi(\mathbf{h}) \) along the \( W_{\tau} \) direction can translate the output distribution. Moving \( \phi(\mathbf{h}) \) in a different direction slices through a different region of the landscape and can change the distribution's shape; however, the change is constrained by whatever shape variations the fixed landscape contains, and so many distribution shapes are not reachable from any \( \phi(\mathbf{h}) \).

We presented the toy model as a way to give an intuitive picture of the \model{nn} head's possible limitations—the toy model does not act as conclusive evidence for any specific limitation.  The next section presents empirical evidence which (partially) supports the hypothesis that the \model{nn} head can easily translate but not modulate peaks in a distribution.

\subsubsection{Empirical evidence}
We take a closer look at the behaviour of the \model{nn} head by looking at the output of the \model{rnn-nn} model trained on four datasets (1 real, 3 synthetic). We also include the outputs of \model{rnn-pc} and \model{rnn-logmix} for comparison.

\begin{wrapfigure}[21]{R}{0.50\textwidth}
\vspace{-0.35cm}
  \includegraphics[width=2.70in]{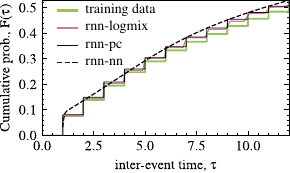}
  \caption{Empirical and predicted cumulative distribution function (CDF) for the PUBG dataset. The \textcolor[HTML]{4b6e1a}{green curve} is the empirical CDF of the training set's inter-event times. The other curves are mean CDFs calculated from \code{rnn-nn}, \code{rnn-logmix} and \code{rnn-pc} outputs, meaned over 4096 input histories drawn from the training set.} 
  \label{fig:app_nn_cdf_pubg}
\end{wrapfigure}

\textbf{Narrow densities in a real-world dataset}. First, we look at how well the \model{rnn-nn} model's output lines up with the empirical distribution for the PUBG dataset, a dataset with multiple narrow peaks in the distribution of inter-event times (see \cref{fig:app_pubg_hist}). 10 models (1 per run) were trained for the results of the main body (\cref{sec:results}), and we take 1 of these and inspect its output. We present just 1 model here, but we have checked all 10 and they show the same pattern that we present below. In \cref{fig:app_nn_cdf_pubg}, the \acp{CDF} corresponding to the output emitted by the \model{rnn-nn}, \model{rnn-pc} and \model{rnn-logmix} models trained on the PUBG dataset are compared to the training set's empirical CDF. For visual comparison, the CDF rather than the PDF is used as the PDFs of the models have a disparate range of values. The model CDFs are calculated by averaging over 4096 input histories from the training set. What is apparent from \cref{fig:app_nn_cdf_pubg} is that jumps in the CDF corresponding to narrow peaks in the PUBG distribution are matched, albeit roughly, by the \model{rnn-logmix} and \model{rnn-pc} models, but not by the \model{rnn-nn} model whose CDF remains smooth. This is an example where the \model{rnn-nn} model fails to reflect narrow peaks present in the target distribution. The next case study shows that the \model{rnn-nn} model \textit{does} have the capacity to represent narrow peaks, if the target distribution is stationary.

\begin{figure*}[t]
\centering
 \begin{subfigure}[t]{2.00in}
     \centering
 \includegraphics[width=\linewidth]{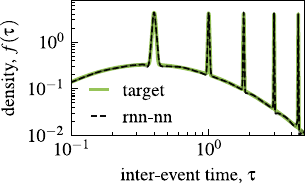}
  \caption{rnn-nn}
\end{subfigure}\hfill
\begin{subfigure}[t]{1.65in}
    \centering
  \includegraphics[width=\linewidth]{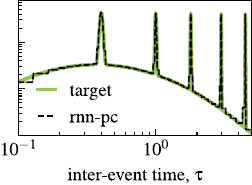}
  \caption{rnn-pc}
\end{subfigure}\hfill
\begin{subfigure}[t]{1.65in}
    \centering
  \includegraphics[width=\linewidth]{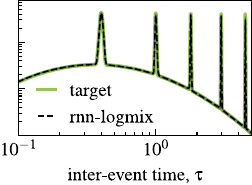}
  \caption{rnn-logmix}
\end{subfigure}%
\caption{
  Probability distributions emitted by \model{rnn-nn}, \model{rnn-pc} and \model{rnn-logmix} models trained on a synthetic sequence where inter-event times are drawn from a stationary mixture (lognormal and five narrow Gaussian peaks). The true distribution of the inter-event times is shown in \textcolor[HTML]{4b6e1a}{green}. The \model{rnn-nn}'s distribution contains five peaks at the correct locations. A black dashed line is the mean of five training runs. The individual runs are shown as semi-transparent gray curves; however, all are close to the mean and difficult to distinguish. 
} 
\label{fig:app_nn_peaks_multi}
\end{figure*}

\textbf{Fixed peaks}. We use a synthetic dataset to show a case where the \model{rnn-nn} model can be trained to fit a distribution with multiple sharp peaks. Results are shown in \cref{fig:app_nn_peaks_multi}. A synthetic sequence's inter-event times are independently drawn from a six-component mixture: a lognormal distribution and five narrow Gaussian distributions. The inter-event times do not depend on the event history. This case study is evidence that the \model{nn} head can represent distributions with multiple narrow peaks. Interestingly, the \model{rnn-logmix} model struggles to fit all five peaks.

\begin{figure*}[t]
\captionsetup[subfigure]{skip=3pt, belowskip=7pt, justification=centering, margin=0cm}
\centering
\begin{subfigure}[t]{2.09in}
    \centering
    \includegraphics[width=\linewidth]{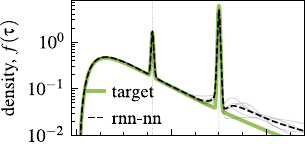}
  \captionsetup{justification=raggedleft, singlelinecheck=false, margin=0.2in}
  \caption{rnn-nn, target has peak}
\end{subfigure}\hfill
\begin{subfigure}[t]{1.60in}
    \centering
  \includegraphics[width=\linewidth, trim=0 -5pt 0 0, clip]{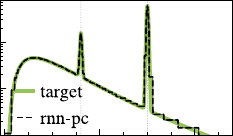}
  \caption{rnn-pc, target has peak}
\end{subfigure}\hfill
\begin{subfigure}[t]{1.60in}
    \centering
  \includegraphics[width=\linewidth, trim=0 -5pt 0 0, clip]{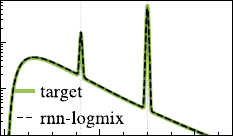}
  \caption{rnn-logmix, target has peak}
\end{subfigure} \\
\begin{subfigure}[t]{2.09in}
    \centering
  \includegraphics[width=\linewidth]{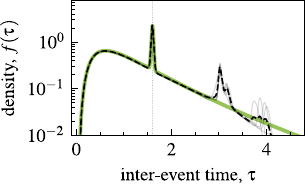}
  \captionsetup{justification=raggedleft, singlelinecheck=false, margin=0.1in}
  \caption{rnn-nn, target without peak}
\end{subfigure}\hfill
\begin{subfigure}[t]{1.60in}
    \centering
  \includegraphics[width=\linewidth]{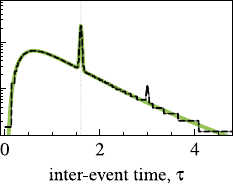}
  \caption{rnn-pc, target without peak}
\end{subfigure}\hfill
\begin{subfigure}[t]{1.60in}
    \centering
  \includegraphics[width=\linewidth]{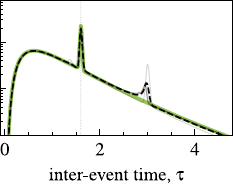}
  \caption{rnn-logmix, target without peak}
\end{subfigure}%
\caption{Two different outputs for three models trained on a synthetic sequence where the distribution of the next inter-event time switches between having and not having a narrow peak. The distribution of \( \tau_{i+1} \) depends only on \( \tau_{i} \): if \( \tau_{i} \ge 1 \), then \( p(\tau_{i+1}| \tau_i) \sim \operatorname{lognormal}(0, 1) \), else \( p(\tau_{i+1} | \tau_i) \) is a mixture of the same lognormal distribution and a narrow Gaussian (\( \sigma = 0.02 \)) centered at \( \tau_{i+1} = 2 \). The mixture is \( 0.7 \cdot \operatorname{LogNormal}(0,1) + 0.3 \cdot \mathcal{N}(2,0.02^2) \).
Both \model{rnn-pc} and \model{rnn-logmix} can (somewhat) use the history input to modulate the probability density at the peak location. We do not observe the \model{rnn-nn} model carrying out this modulation. This experiment was repeated five times with different random seeds, and the same pattern was observed each time (the figures above show just one run).} 
\label{fig:app_nn_peaks_binary}
\end{figure*}

\textbf{Present/absent peaks}. Modulating peaks is not as well achieved. Now we use a synthetic sequence with a smooth inter-event time distribution with a single peak; but this time, the peak is either present or absent depending on the input. If the most recent inter-event time, \( \tau_{i} \) is less than 1, then the peak is present in the distribution, otherwise it is absent. \cref{fig:app_nn_peaks_binary} shows the results of training \model{rnn-nn} on this dataset, along with \model{rnn-pc} and \model{rnn-logmix}. Both \model{rnn-pc} and \model{rnn-logmix} effectively modulate the density at the location of the peak, whereas \model{rnn-nn}'s predictions underestimate when the peak is present and overestimate when the peak is absent. Peak/non-peak density is \( \sim 25\times \) for the \model{rnn-pc} and \model{rnn-logmix} models and only \( \sim 2 \times \) for the \model{rnn-nn} model. This case study is evidence (though not conclusive) that the \model{nn} head struggles to modulate sharp peaks in a distribution, even though it can represent both in the case of a stationary distribution. It is tempting to connect this failure to the idea put forward by the toy model of \cref{fig:app_nn_demo}: that both distributional shapes may not be present in the \( \mathbf{u} \)-space along the fixed \( W_{\tau} \) direction, or that they are present but the model struggles to learn to move \( \phi(\mathbf{h}) \) between them. We do not have enough evidence to support this connection and instead simply state it as a hypothesis.

\begin{figure*}[t]
\captionsetup[subfigure]{skip=3pt, belowskip=7pt, justification=centering, margin=0cm}
\centering
\begin{subfigure}[t]{2.09in}
    \centering
  \includegraphics[width=\linewidth]{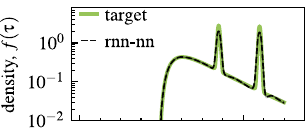}
  \captionsetup{justification=raggedleft, singlelinecheck=false, margin=0.2in}
  \caption{rnn-nn, peak at \( \tau = 3.8 \)}
\end{subfigure}\hfill
\begin{subfigure}[t]{1.60in}
  \centering
  \includegraphics[width=\linewidth, trim=0 -5pt 0 0, clip]{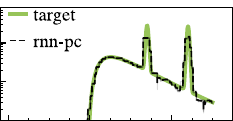}
  \caption{rnn-pc, peak at \( \tau = 3.8 \)}
\end{subfigure}\hfill
\begin{subfigure}[t]{1.60in}
    \centering
  \includegraphics[width=\linewidth, trim=0 -5pt 0 0, clip]{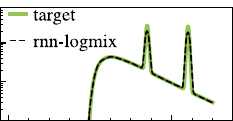}
  \caption{rnn-logmix, peak at \( \tau = 3.8 \)}
\end{subfigure} \\
\begin{subfigure}[t]{2.09in}
    \centering
  \includegraphics[width=\linewidth]{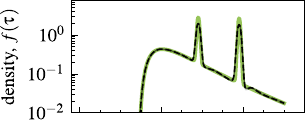}
  \captionsetup{justification=raggedleft, singlelinecheck=false, margin=0.2in}
  \caption{rnn-nn, peak at \( \tau = 2.8 \)}
\end{subfigure}\hfill
\begin{subfigure}[t]{1.60in}
    \centering
  \includegraphics[width=\linewidth, trim=0 -5pt 0 0, clip]{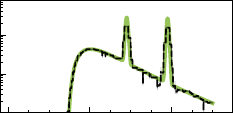}
  \caption{rnn-pc, peak at \( \tau = 2.8 \)}
\end{subfigure}\hfill
\begin{subfigure}[t]{1.60in}
    \centering
  \includegraphics[width=\linewidth, trim=0 -5pt 0 0, clip]{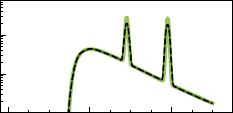}
  \captionsetup{justification=raggedright, singlelinecheck=false, margin=-0.06in, skip=-1pt}
  \caption{rnn-logmix, peak at \( \tau = 2.8\)}
\end{subfigure} \\
\begin{subfigure}[t]{2.09in}
    \centering
  \includegraphics[width=\linewidth]{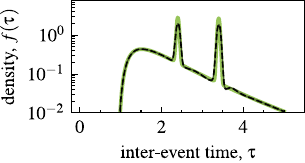}
  \captionsetup{justification=raggedleft, singlelinecheck=false, margin=0.2in}
  \caption{rnn-nn, peak at \( \tau = 1.8 \)}
\end{subfigure}\hfill
\begin{subfigure}[t]{1.60in}
    \centering
  \includegraphics[width=\linewidth]{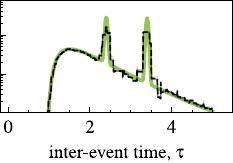}
  \caption{rnn-pc, peak at \( \tau = 1.8 \)}
\end{subfigure}\hfill
\begin{subfigure}[t]{1.60in}
    \centering
  \includegraphics[width=\linewidth]{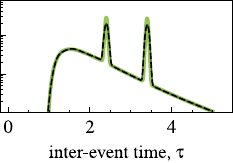}
  \caption{rnn-logmix, peak at \( \tau = 1.8 \)}
\end{subfigure}%
\caption{Three models trained on a synthetic sequence in which the distribution of the next inter-event time has a narrow peak whose center is equal to the previous inter-event duration. The distribution of \( \tau_{i+1} \) is a mixture of a background lognormal and a Gaussian (\( \sigma = 0.03\tau_i \)) centered at \( \tau_i \). Specifically,  
  \( p(\tau_{i+1} \mid \tau_i) = 0.6 \cdot \operatorname{LogNormal}(0, 1) + 0.4 \cdot \mathcal{N}\bigl(\tau_i,\,(0.03\,\tau_i)^2\bigr) \). All three models can use the history input to translate the location of a peak.}
\label{fig:app_nn_peaks_moving}
\end{figure*}

\textbf{Translated peaks}. A third case study, similar to the second, considers the \textit{translation} of a single peak in an otherwise smooth distribution, where the amount of translation depends on the input. The output distribution of the next inter-event time \( \tau_{i+1} \) for this synthetic event sequence is a lognormal and Gaussian mixture, where the narrow Gaussian is centered at a copy of the previous inter-event time, \( \tau_{i} \). 
The training set contains a range of \( \tau_{i} \) values between 0 and 10. \cref{fig:app_nn_peaks_moving} shows three example outputs for \model{rnn-nn}, \model{rnn-pc} and \model{rnn-logmix}. For this dataset, the peak is not narrow enough to cause training instability for the \model{logmix} head. In this case, \model{rnn-nn} and \model{rnn-logmix} perform similarly, with the output containing a peak that translates with input. It is the \model{rnn-pc} model that struggles the most in this case. This case study is evidence that the \model{nn} head \textit{can} translate peaks in the output distribution conditioned on input history. If we were to explain this result using the toy model of \cref{fig:app_nn_demo}, we could argue that a translated distribution must be present in the \( \mathbf{u} \)-space, and that a movement of \( \phi(\mathbf{h}) \) in the \( W_{\tau} \) direction would enable this translation. We do not have enough evidence to support this connection, and present it only as a candidate explanation.

\begin{figure*}[t]
\centering
 \begin{subfigure}[t]{1.875in}
     \centering
   \includegraphics[width=\linewidth]{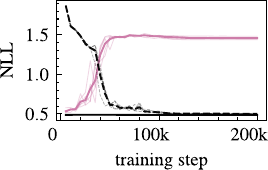}
  \caption{fixed}
  \label{fig:app_nn_train_curves_fixed}
\end{subfigure}\hfill
\begin{subfigure}[t]{1.69in}
    \centering
  \includegraphics[width=\linewidth]{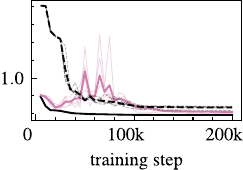}
  \caption{present/absent}
\end{subfigure}\hfill
\begin{subfigure}[t]{1.69in}
    \centering
    \includegraphics[width=\linewidth]{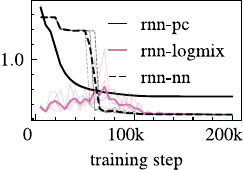}
  \caption{translated}
\end{subfigure}%
\caption{
  NLL evaluated every 5k steps throughout 200k-step training runs for the three models (\model{rnn-nn}, \model{rnn-pc} and \model{rnn-logmix}) on the three synthetic datasets (fixed peaks, binary peaks, and moving peaks). The NLL is calculated on \(2^{16} \) samples from a held-out set. There are 5 individual runs displayed as semi-transparent curves, and the mean across them is opaque.} 
\label{fig:app_nn_train_curves}
\end{figure*}

\begin{figure*}[t]
\captionsetup[subfigure]{skip=6pt, belowskip=6pt, justification=centering, margin=0cm}
\centering
 \begin{subfigure}[t]{2.63in}
     \centering
   \includegraphics[width=\linewidth]{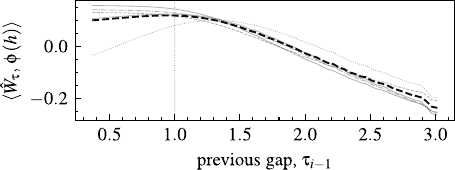}
   \caption{\( \hat{W}_{\tau} \) for the present/absent dataset}
\end{subfigure}\hfill
\begin{subfigure}[t]{2.62in}
     \centering
   \includegraphics[width=\linewidth]{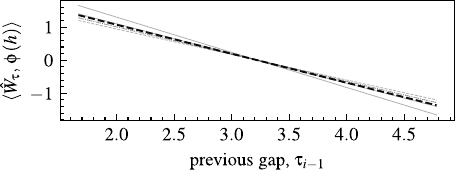}
   \caption{\( \hat{W}_{\tau} \) for the moving dataset}
\end{subfigure} \\ 
\begin{subfigure}[t]{2.54in}
    \centering
    \includegraphics[width=\linewidth]{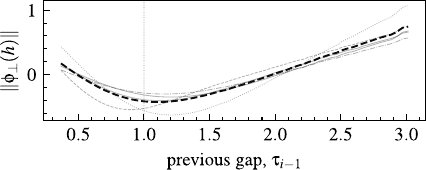}
    \caption{\( W_{\tau}^{\perp} \) for the present/absent dataset}
\end{subfigure}\hfill
\begin{subfigure}[t]{2.71in}
    \centering
    \includegraphics[width=\linewidth]{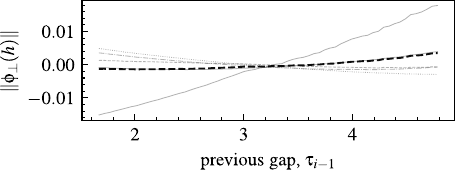}
    \caption{\( W_{\tau}^{\perp} \) for the moving dataset}
\end{subfigure}%
\caption{Decomposition of \( \phi(h) \in \mathbb{R}^{64} \) (from \cref{eq:phi_def}) along the unit direction \( \hat{W}_{\tau} = W_{\tau} / \|W_{\tau}\| \), and the (63-dimensional) orthogonal complement, \( W_{\tau}^{\perp} \). This is done for two datasets, the \textit{present/absent peaks} dataset (\textbf{left}) and the \textit{translated peaks} dataset (\textbf{right}). 
  The component of \( \phi(h) \) parallel to \( \hat{W}_{\tau} \) is \( \langle \hat{W}_{\tau}, \, \phi(h) \rangle \) (\textbf{top}) and the norm of the 63-d orthogonal component is \( \|\phi(h) - \langle \hat{W}_{\tau},\, \phi(h) \rangle \hat{W}_{\tau}\| \), which we denote as \( \|\phi_{\perp}(h)\| \) (\textbf{bottom}). For each dataset we draw \( 2^{17} \) histories from a held-out set and evaluate \( \phi(h) \), making note of \( \tau_{i-1} \). This is done for 5 runs (grey), and the mean across runs is plotted in black. All runs have their individual mean subtracted to bring the curves into the same range. The per-run mean operates on values binned by the x-axis, \( \tau_{i-1} \) into 64 bins from the 5th until the 95th quantile of \( \tau_{i-1} \). The parallel component is strongly correlated with \( \tau_{i-1} \) for the \textit{translated peaks} dataset, and not for the \textit{present/absent peaks} dataset.}
\label{fig:app_nn_W_proj}
\end{figure*}

\textbf{Training details}. For the three case studies above, the three models, \model{rnn-nn}, \model{rnn-pc} and \model{rnn-logmix}, were trained for 200k steps with a batch size of 64, with all other training settings the same as the main body (see \cref{app:training_settings}). The checkpoints with the best loss were used for evaluation. Each experiment was repeated five times to ensure the results were consistent. We present just one run for each experiment to keep the figures clear. Training curves for all runs are shown in \cref{fig:app_nn_train_curves}.

\subsubsection{Discussion}
The test cases and the toy model suggest (but do not conclusively prove) that the \model{nn} head can easily translate peaks in a distribution, but is less capable of making point changes to a distribution such as adding and removing a narrow peak. We suggest that this may explain why the \model{nn} head is outperformed by the \model{logmix} and \model{pc} heads on many of the datasets considered in this work. We have presented only limited evidence to support this implication, and so we present it as a theory rather than a conclusion. 

This section focused on the \model{nn} head; however, some of the results are relevant to the \model{logmix} head as well. The \model{rnn-logmix} training loss curve in \cref{fig:app_nn_train_curves_fixed} shows how the model's loss increases and doesn't recover. Inspecting the output distribution of a model checkpoint after this spike shows that the model loses the ability to represent all of the peaks in the distribution. This result supports the previous section's argument that the \model{logmix} head can be perturbed while training if there are narrow peaks in the target distribution. The training curves for the two other datasets (present/absent and translated peaks) also suggest instability in the training process.

\subsubsection{Model ablation: repeat for \model{gpt-a} stem}
The synthetic experiments in the above section used the RNN stem for all models. To rule out the possibility that the results are peculiar to the RNN stem, the same experiments were repeated with the \model{gpt-a} stem, and those results appear below.

\FloatBarrier

\begin{figure*}[h]
\centering
 \begin{subfigure}[t]{2.00in}
     \centering
 \includegraphics[width=\linewidth]{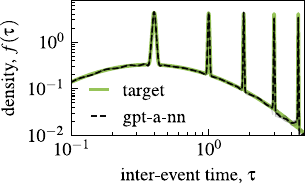}
  \caption{gpt-a-nn}
\end{subfigure}\hfill
\begin{subfigure}[t]{1.65in}
    \centering
  \includegraphics[width=\linewidth]{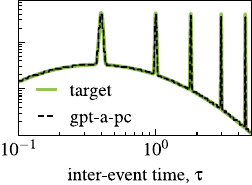}
  \caption{gpt-a-pc}
\end{subfigure}\hfill
\begin{subfigure}[t]{1.65in}
    \centering
  \includegraphics[width=\linewidth]{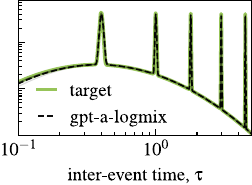}
  \caption{gpt-a-logmix}
\end{subfigure}%
\caption{
  Probability distributions emitted by \model{gpt-a-nn}, \model{gpt-a-pc} and \model{gpt-a-logmix} models trained on a synthetic sequence where inter-event times are drawn from a stationary mixture (lognormal and five narrow Gaussian peaks). The true distribution of the inter-event times is shown in \textcolor[HTML]{4b6e1a}{green}. The \model{gpt-a-nn}'s distribution contains five peaks at the correct locations. A black dashed line is the mean of five training runs. The individual runs are shown as semi-transparent gray curves; however, all are close to the mean and difficult to distinguish. 
} 
\label{fig:app_nn_peaks_multi_gpt}
\end{figure*}

\begin{figure*}[h]
\captionsetup[subfigure]{skip=3pt, belowskip=7pt, justification=centering, margin=0cm}
\centering
\begin{subfigure}[t]{2.09in}
    \centering
  \includegraphics[width=\linewidth]{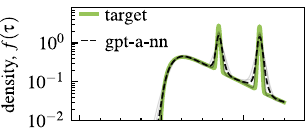}
  \captionsetup{justification=raggedleft, singlelinecheck=false, margin=0.2in}
  \caption{gpt-a-nn, peak at \( \tau = 3.8 \)}
\end{subfigure}\hfill
\begin{subfigure}[t]{1.60in}
    \centering
  \includegraphics[width=\linewidth, trim=0 -5pt 0 0, clip]{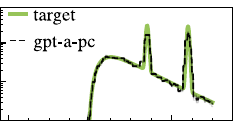}
  \caption{gpt-a-pc, peak at \( \tau = 3.8 \)}
\end{subfigure}\hfill
\begin{subfigure}[t]{1.60in}
    \centering
  \includegraphics[width=\linewidth, trim=0 -5pt 0 0, clip]{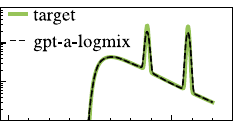}
  \caption{gpt-a-logmix, peak at \( \tau = 3.8 \)}
\end{subfigure} \\
\begin{subfigure}[t]{2.09in}
    \centering
  \includegraphics[width=\linewidth]{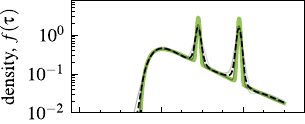}
  \captionsetup{justification=raggedleft, singlelinecheck=false, margin=0.2in}
  \caption{gpt-a-nn, peak at \( \tau = 2.8 \)}
\end{subfigure}\hfill
\begin{subfigure}[t]{1.60in}
    \centering
  \includegraphics[width=\linewidth, trim=0 -5pt 0 0, clip]{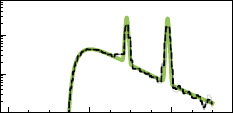}
  \caption{gpt-a-pc, peak at \( \tau = 2.8 \)}
\end{subfigure}\hfill
\begin{subfigure}[t]{1.60in}
    \centering
  \includegraphics[width=\linewidth, trim=0 -5pt 0 0, clip]{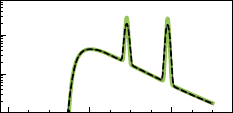}
  \captionsetup{justification=raggedright, singlelinecheck=false, margin=-0.06in, skip=-1pt}
  \caption{gpt-a-logmix, peak at \( \tau = 2.8 \)}
\end{subfigure} \\
\begin{subfigure}[t]{2.09in}
    \centering
  \includegraphics[width=\linewidth]{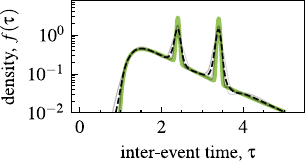}
  \captionsetup{justification=raggedleft, singlelinecheck=false, margin=0.2in}
  \caption{gpt-a-nn, peak at \( \tau = 1.8 \)}
\end{subfigure}\hfill
\begin{subfigure}[t]{1.60in}
    \centering
  \includegraphics[width=\linewidth]{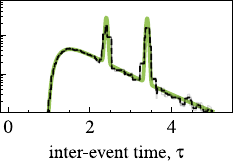}
  \caption{gpt-a-pc, peak at \( \tau = 1.8 \)}
\end{subfigure}\hfill
\begin{subfigure}[t]{1.60in}
    \centering
  \includegraphics[width=\linewidth]{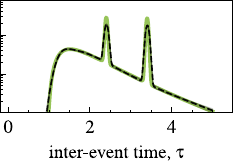}
  \caption{gpt-a-logmix, peak at \( \tau = 1.8 \)}
\end{subfigure}%
\caption{Three models trained on a synthetic sequence in which the distribution of the next inter-event time has a narrow peak whose center is equal to the previous inter-event duration. The distribution of \( \tau_{i+1} \) is a mixture of a background lognormal and a Gaussian (\( \sigma = 0.03\tau_i \)) centered at \( \tau_i \). Specifically,  
  \( p(\tau_{i+1} \mid \tau_i) = 0.6 \cdot \operatorname{LogNormal}(0, 1) + 0.4 \cdot \mathcal{N}\bigl(\tau_i,\,(0.03\,\tau_i)^2\bigr) \). All three models can use the history input to translate the location of a peak.}
\label{fig:app_nn_peaks_moving_gpt}
\end{figure*}

\begin{figure*}[h]
\centering
 \begin{subfigure}[t]{1.875in}
     \centering
   \includegraphics[width=\linewidth]{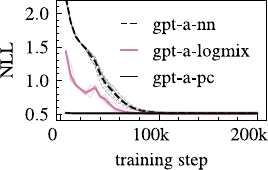}
  \caption{fixed}
\end{subfigure}\hfill
\begin{subfigure}[t]{1.69in}
    \centering
  \includegraphics[width=\linewidth]{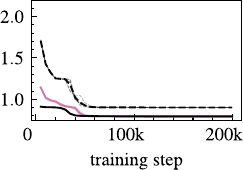}
  \caption{present/absent}
\end{subfigure}\hfill
\begin{subfigure}[t]{1.69in}
    \centering
    \includegraphics[width=\linewidth]{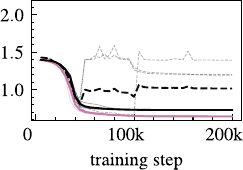}
  \caption{translated}
\end{subfigure}%
\caption{
  NLL evaluated every 5k steps throughout 200k-step training runs for the three models (\model{gpt-a-nn}, \model{gpt-a-pc} and \model{gpt-a-logmix}) on the three synthetic datasets (fixed peaks, binary peaks, and moving peaks). The NLL is calculated on \(2^{16} \) samples from a held-out set. There are 5 individual runs displayed as semi-transparent curves, and the mean across them is opaque.} 
\label{fig:app_nn_train_curves_gpt}
\end{figure*}

\begin{figure*}[h]
\captionsetup[subfigure]{skip=6pt, belowskip=6pt, justification=centering, margin=0cm}
\centering
 \begin{subfigure}[t]{2.6in}
     \centering
   \includegraphics[width=\linewidth]{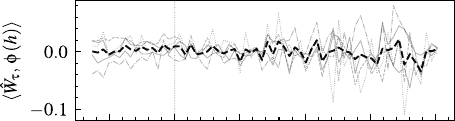}
   \caption{\( \hat{W}_{\tau} \) for the present/absent dataset}
\end{subfigure}\hfill
\begin{subfigure}[t]{2.5in}
     \centering
   \includegraphics[width=\linewidth]{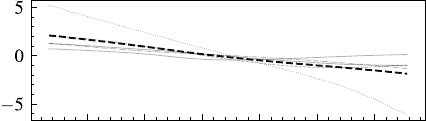}
   \caption{\( \hat{W}_{\tau} \) for the moving dataset}
\end{subfigure} \\ 
 \begin{subfigure}[t]{2.6in}
    \centering
    \includegraphics[width=\linewidth]{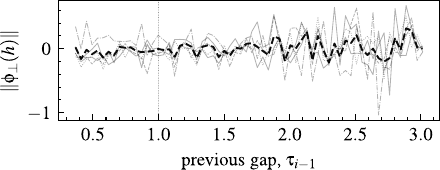}
    \caption{\( W_{\tau}^{\perp} \) for the present/absent dataset}
\end{subfigure}\hfill
\begin{subfigure}[t]{2.5in}
    \centering
    \includegraphics[width=\linewidth]{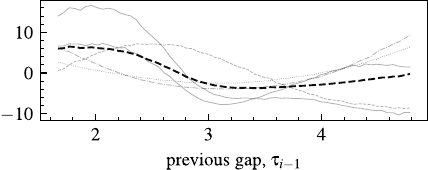}
    \caption{\( W_{\tau}^{\perp} \) for the moving dataset}
\end{subfigure}%
\caption{Decomposition of \( \phi(h) \in \mathbb{R}^{64} \) (from \cref{eq:phi_def}) along the unit direction \( \hat{W}_{\tau} = W_{\tau} / \|W_{\tau}\| \), and the (63-dimensional) orthogonal complement, \( W_{\tau}^{\perp} \). This is done for two datasets, the \textit{present/absent peaks} dataset (\textbf{left}) and the \textit{translated peaks} dataset (\textbf{right}). 
  The component of \( \phi(h) \) parallel to \( \hat{W}_{\tau} \) is \( \langle \hat{W}_{\tau}, \, \phi(h) \rangle \) (\textbf{top}) and the norm of the 63-d orthogonal component is \( \|\phi(h) - \langle \hat{W}_{\tau},\, \phi(h) \rangle \hat{W}_{\tau}\| \), which we denote as \( \|\phi_{\perp}(h)\| \) (\textbf{bottom}). For each dataset we draw \( 2^{17} \) histories from a held-out set and evaluate \( \phi(h) \), making note of \( \tau_{i-1} \). This is done for 5 runs (grey), and the mean across runs is plotted in black. All runs have their individual mean subtracted to bring the curves into the same range. The per-run mean operates on values binned by the x-axis, \( \tau_{i-1} \) into 64 bins from the 5th until the 95th quantile of \( \tau_{i-1} \). The parallel component is strongly correlated with \( \tau_{i-1} \) for the \textit{translated peaks} dataset, and not for the \textit{present/absent peaks} dataset.}
\label{fig:app_nn_W_proj_gpt}
\end{figure*}

\clearpage

\section{Training and evaluation details} \label{app:training_settings}
This appendix describes hyperparameters and other settings associated with training and evaluation. It also describes the compute resources used. 

\textbf{Software}. PyTorch 2.6.0 with CUDA 13.2 was used for training and evaluation.

\textbf{Optimizer}. All training runs use the AdamW optimizer described by \citet{loshchilovDecoupledWeightDecay2018}. AdamW parameters (\( \beta_1 \), \( \beta_2 \), eps, weight decay) were set to (0.9, 0.99, \( 1 \times 10^{-5}\), 0.02)—chosen roughly in line with heuristics described by \citet{howardDeepLearningCoders2020}. Weight decay was applied to all parameters except bias and embedding parameters.

\textbf{Precision}. PyTorch's automatic mixed precision and gradient scaling were used during training. In cases where greater numerical precision is needed, such as when calculating probabilities from hazard outputs, 32-bit floating-point precision was used.

\textbf{Loss function}. All models, except one, were trained to minimize negative log-likelihood. The exception is \model{thp} which uses a loss term that is the sum of a negative log-likelihood term and a mean-squared error term \citep{zuoTransformerHawkesProcess2020}.

\textbf{Model choice}. The final models are selected from the checkpoints taken during training with the lowest validation loss. 

\textbf{Learning rate scheduler}. The 1-cycle learning rate policy with 3 phases described by \citet{smithSuperconvergenceVeryFast2019} was used.

\textbf{Maximum learning rate}. The maximum learning rate, which parameterizes the learning rate scheduler, was chosen by carrying out learning rate range tests as described by \citet{smithCyclicalLearningRates2017a}. To alleviate the sensitivity to noise that this approach has, we used 8 repeats and Kalman smoothing to make the selection more robust. All learning rate curves and chosen learning rates were also manually inspected, and a small number of automatically chosen learning rates were manually overridden where the selection algorithm had failed to correctly interpret the learning rate sweep curve.

\textbf{Batch and epochs}. A batch size of 128 was used. The number of epochs was set to 512; however, this can be less if early stopping is triggered (but always greater than 256). Evaluations were run at the end of every epoch.

\textbf{Early stopping}. An early-stopping strategy was used to reduce wasted compute. After at least 1/2 of the total training steps have been completed (when the learning rate is decreasing), early stopping is triggered if the smoothed validation loss (exponential moving average with smoothing factor 0.2) has not improved over 12 evaluations. Post-training, training and validation loss curves were inspected to ensure that early stopping was not triggered prematurely.

\subsection{Edge selection algorithm runtime} \label{app:dp_runtime}
\begin{figure}
  \centering
  \includegraphics[width=4.5in]{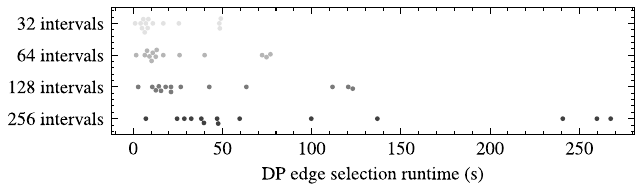}
  \caption{Seconds taken to calculate edges (\cref{alg:edge_opt}) for the 14 real-world datasets across 4 interval counts. Variability is due to differences in the dataset sizes and the number of unique inter-event times within each dataset. All runs used a candidate edge set of size either 15k or 10\% of the training set, whichever was less.}
\label{fig:app_dp_runtime}
\end{figure}

Calculating the edges via the dynamic programming procedure (\cref{alg:edge_opt}) for the main results, with 128 intervals, takes on the order of tens of seconds for the 14 real-world datasets. \cref{fig:app_dp_runtime} shows the runtime for the 14 datasets across 4 bin counts, 32, 64, 128, 256.

\subsection{Compute resources} \label{app:compute_resources}
Model training was carried out on 10 Nvidia A40 GPUs from an internal cluster and took \textasciitilde 320 GPU-days. Model evaluation was run on a workstation with GPU, CPU and RAM specifications: Nvidia RTX 4090 GPU, AMD Ryzen Threadripper PRO 5975WX and 256 GiB RAM. Evaluation took \textasciitilde 35 hours.

\end{document}